\documentclass{article}

\usepackage{arxiv}

\usepackage[utf8]{inputenc} 
\usepackage[T1]{fontenc}    
\usepackage{hyperref}       
\usepackage{url}            
\usepackage{booktabs}       
\usepackage{amsfonts}       
\usepackage{nicefrac}       
\usepackage{microtype}      
\usepackage{lipsum}		
\usepackage{graphicx}
\usepackage{natbib}
\usepackage{doi}
 \usepackage{float}
 \usepackage{multirow}
\usepackage{caption}
 \usepackage{subcaption}
 \usepackage[dvipsnames]{xcolor}

\title{Impact of enriched meaning representations for language generation in dialogue tasks: A comprehensive exploration of the relevance of tasks, corpora and metrics.}

\author{ \href{https://orcid.org/0000-0002-3108-5874}{\includegraphics[scale=0.06]{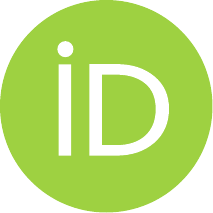}\hspace{1mm}Alain Vázquez} \\
	Speech Interactive Group\\
	University of the Basque Country (UPV/EHU)\\
	Sarriena s/n, 48940 Leioa,  Spain \\
	\texttt{alain.vazquez@ehu.eus} \\
	\And
	\href{https://orcid.org/0000-0002-1773-3214}{\includegraphics[scale=0.06]{orcid.pdf}\hspace{1mm}Maria Inés Torres} \\
	Speech Interactive Group\\
	University of the Basque Country (UPV/EHU)\\
	Sarriena s/n, 48940 Leioa,  Spain \\
	\texttt{manes.torres@ehu.eus} \\
}

\renewcommand{\shorttitle}{Impact of enriched meaning representations for language generation in dialogue tasks.}

\hypersetup{
pdftitle={Impact of prompt-based meaning representations for language generation in dialogue tasks.},
pdfsubject={q-bio.NC, q-bio.QM},
pdfauthor={A. Vazquez, M. Ines Torres},
pdfkeywords={Natural Language Generation, Dialogue Systems, Dialogue Acts, Prompt-
based Learning, Evaluation of NLG},
}

\begin{document}
\maketitle

\begin{abstract}
Conversational systems should generate diverse language forms to interact fluently and accurately with a variety of users. In this context, Natural Language Generation (NLG) engines convert Meaning Representations (MRs) into sentences, directly influencing user perception. These MRs usually encode the communicative function (e.g., inform, request, confirm) via DAs and enumerate the semantic content with slot–value pairs. In this work, our objective is to analyse whether providing a task demonstrator to the generator enhances the generations of a fine-tuned model. This demonstrator is an MR–sentence pair extracted from the original dataset that enriches the input at training and inference time. The analysis involves five metrics (BLEU, BLEURT, LaBSE, Slot Accuracy and Dialogue Act Accuracy) that focus on different linguistic aspects, and four datasets  (E2E, ViGGO, MultiWOZ, and EMPATHIC) that differ in multiple features, such as domain, size, lexicon, MR variability, and acquisition process. To the best of our knowledge, this is the first study on dialogue NLG, implementing a comparative analysis of the impact of MRs on the generation quality across domains, corpora characteristics and the metrics used to evaluate these generations. Our key insight is that the proposed enriched inputs are effective for complex tasks and small datasets with high variability in MRs and sentences. They are also beneficial in zero-shot settings for any domain. Moreover, the analysis of the metrics shows that the semantic metrics capture the generation quality more accurately than lexical metrics. In addition, among these semantic metrics, those trained with human ratings can detect omissions and other subtle semantic issues that embedding-based metrics often miss. Finally, the evolution of the metric scores and the excellent results for Slot Accuracy and Dialogue Act Accuracy demonstrate that the generative models present fast adaptability to different tasks and robustness at semantic and communicative intention levels.
\end{abstract}


\section{Introduction}

The impressive development of Transformers, along with the rapid increase in computing capabilities and the accessibility of vast amounts of data, has revolutionised natural language processing (NLP) in recent years. Large language models (LLMs) have excelled in tasks such as transcription, translation or summarisation, among others. However, LLMs present well-known challenges, including the opacity of training data and the lack of clarity in their behaviour. In conversational systems, a major concern is their tendency to hallucinate, generate misinformation and lack reasoning. Furthermore, while LLMs could recognise lexical and structural patterns, they struggle with semantic inferences and extralinguistic reasoning essential for real-world knowledge tasks \citep{lappin2024assessing}.

Conversational systems should be able to generate diverse language forms to interact fluently and accurately with a variety of users. In this context, natural language generation (NLG) plays a crucial role in dialogue, directly influencing user perception. In open-domain dialogues, the NLG must be able to express opinions and emotions, convey facts, provide information, ask contextual questions (\textit{who, what, where, when, why}), confirm user inputs or give instructions. Conversely, closed-domain dialogues must align with the specific task, objectives, and conversational context. In both cases, generated sentences should be semantically controlled, accurate and free of hallucinations.

Although end-to-end architectures have advanced in the design of dialogue systems \citep{ji2024wavchat,zorrilla2022multilingual}, they still have difficulty achieving dialogue coherence and interpretability. In this context, the concepts of dialogue acts (DA), which represent a sentence's intentionality or purpose, and meaning representation (MR), which adds specific attributes and entities in the shape of slot-value pairs, remain essential for managing conversational flow. Specifically, NLG generates MR-based sentences, i.e., it takes the meaning representation as input and produces sentences as output. When combined with LLMs, both closed- and open-domain dialogue systems \citep{wang2024towards} benefit from improved adaptability to complex tasks \citep{galland2024generating}, as well as enhanced coherence and response accuracy \citep{griol2024combining,vazquez2024knowledge,hu2022dialogue}.

In this work, we aim to explore different ways of representing meaning for sentence generation. Our objective is to analyse whether providing the generator model with additional information, in the form of a task demonstrator, enhances generation quality. This task demonstrator is a sample from the original dataset consisting of an MR–sentence pair and enriches the input both at training and inference time. Specifically, we investigate whether these one-shot representations improve generation compared to inputs containing only the MR.

The comparison requires evaluation metrics. In NLG, classic metrics, some originally from translation or language model evaluation, do not always capture relevant differences in generated sentences. Nevertheless, each metric assesses a different generation aspect, making their comparison essential for an in-depth study like ours. In this work, we use reference-based metrics such as BLEU \citep{papineni2002bleu}, BLEURT \citep{sellam2020bleurt}, and LABSE \citep{feng2022language}, as well as reference-free metrics like Slot Accuracy \citep{li-etal-2020-slot}, complemented by the more recent Dialogue Act Accuracy (DAC) \citep{ramirez2023controllable}. Our goal is to analyse each metric’s ability to identify differences in sentences generated from the proposed MR formats.

Four corpora have been selected for this study: E2E \citep{novikova2017e2e}, VIGGO \citep{juraska2019viggo}, MULTIWOZ \citep{budzianowski2018multiwoz}, and EMPATHIC \citep{vazquez2023dialogue}. All four are dialogue-oriented NLG corpora in which each sample consists of a DA-based MR and its corresponding example sentence. These datasets differ significantly in domain, size, lexicon, input variability, and acquisition process, among other features. We will analyse the suitability of each MR and assess the discriminative power of each metric based on the characteristics of each corpus.

In summary, this work aims to study the impact of meaning representations on language generation in dialogue tasks. To this end, we thoroughly analyse the relevance of the tasks, the characteristics of the corpora, and the aspects each metric evaluates in the experimental results. For this comparative study, we selected GPT-2 \citep{radford2019language}, which is computationally more accessible and manageable than recent state-of-the-art LLMs such as GPT-4 \citep{achiam2023gpt}, Gemini \citep{team2023gemini}, and LLaMA 3 \citep{dubey2024llama}. In addition, recent work has shown that enriched inputs with task demonstrators can substantially enhance the generation capabilities of medium-sized models \citep{liu2023large,vazquez2023dialogue}.

The main contributions can be outlined as follows:
\begin{itemize}
\item Analysis of the impact of prompt forms on the quality of generated sentences, as well as their zero-shot capabilities.
\item Comparative evaluation of each metric’s ability to assess different quality aspects of generated sentences.
\item Assessment of the suitability of classifying generated sentences into dialogue acts as an evaluation metric for generation quality.
\item Analysis of the influence of corpus design and characteristics, such as the number of references and attributes, on the quality of sentence generation.
\item Cross-analysis of the distribution of the resulting quality scores in terms of representations, corpora and metrics.
\item Human qualitative analysis of the generated sentences.
\end{itemize}

To the best of our knowledge, this is the first in-depth study on NLG in the context of dialogue, implementing a comparative analysis of the impact of meaning representations on the generation quality. This thorough analysis involves multiple domains, tasks, and corpora characteristics, as well as the metrics used to evaluate these generations.

Overall, our experiments yield several practical insights. At the representation level, our proposed inputs are particularly helpful in complex domains and small datasets with high variability in terms of MRs, as well as the lexicon and semantics of their sentences. They also show zero-shot capabilities across domains. In terms of metrics, we observe that semantic metrics capture the quality of the generations more accurately than lexical ones based on n-gram overlap. In addition, when comparing semantic metrics, those trained with human ratings can detect omissions and other subtle semantic issues that embedding-based metrics often miss. Finally, the evolution of metric scores during fine-tuning, together with strong results on Slot Accuracy and DAC, indicates that the models adapt quickly to the task and are robust at the semantic and communicative-intent levels.

The remainder of the paper is structured as follows. Section \ref{sec:related_work} explores the related work. Section \ref{sec:corpora} introduces the corpora used in our experiments and highlights their key differences. Section \ref{sec:representations} explains the explored meaning representations, including the novel enriched representations and the methodology for selecting task demonstrators. Section \ref{sec:metrics} defines the evaluation metrics. Section \ref{sec:results} details the experimental results and discusses their key findings. Section \ref{sec:conclusion} summarises the work with the main conclusions.

\section{Related work}
\label{sec:related_work}

\subsection{Natural Language Generation for dialogue tasks}

Task-oriented dialogue systems commonly adopt modular pipeline architectures due to the higher understandability and controllability of the process than the end-to-end systems \citep{algherairy2024review}. Among these modules, NLG is crucial, transforming structured inputs—MRs—into coherent and fluent natural language outputs. These MRs typically consist of a DA indicating the communicative intent accompanied by slot-value pairs that encode the semantic content to be expressed. Advanced NLG engines that leverage diverse DA-based MRs contribute to generating more dynamic and engaging dialogues, making NLG an interesting area of research \citep{vazquez2025prompt,vazquez2024knowledge,ramirez2023controllable,reed2022jurassic}.

This field experienced a revolution driven by Transformer-based architectures and LLMs. The Transformers \citep{vaswani2017attention} significantly improved the handling of linguistic long-range dependencies, i.e., relationships between tokens or discourse elements that are far apart in the input sequence, and allowed for efficient parallelisation. These architectures laid the foundation for the first LLMs, such as GPT-2 \citep{radford2019language}, T5 \citep{raffel2020exploring}, and BERT \citep{devlin2019bert}, which have proven highly effective for NLG tasks. Models like DialoGPT \citep{zhang2020dialogpt} and SC-GPT \citep{peng2020few} demonstrate the adaptability of GPT-2 for dialogue tasks. On the one hand, DialoGPT, a tunable response generation model trained with conversations, can be integrated into open-domain or domain-specific systems with small-scale fine-tuning. On the other hand, SC-GPT is pre-trained on a large annotated NLG corpus, acquiring the capability to controllably transform MRs into natural language outputs. Finally, this model with a small fine-tuning has also shown capabilities to generate adequate responses for spoken dialogue systems in noisy environments \citep{mousavi2024llms}.

The new era of LLMs---GPT-4 \citep{achiam2023gpt}, Gemini \citep{team2023gemini}, and LLaMA 3 \citep{dubey2024llama}---much larger than their precursors,  have further advanced the state of the art by showcasing capabilities to perform various NLG tasks with minimal or no fine-tuning. ChatGPT\footnote{\url{https://chatgpt.com/}} represents this evolution, demonstrating the ability to handle multi-turn dialogues with coherent, fluent and logical responses. Additionally, \citet{ramirez2023controllable} have shown that these models can produce semantically almost perfect generations without fine-tuning through prompt learning techniques that enrich the input with demonstrators of the task to perform. Similarly, \citet{vazquez2024knowledge} have built a high-quality synthetic dialogue dataset across multiple domains that presents fewer hallucinations and omissions than corpora written by humans.

However, the computational requirements of these new LLMs pose significant challenges, such as hardware requirements, memory constraints or real-time performance. Consequently, some authors still prefer using smaller models, such as GPT-2 \citep{audio2023zorrilla}. \citet{liu2023large,vazquez2023dialogue} have demonstrated that enriched inputs with task demonstrators enhance the generation capabilities of these models. Inspired by them, we perform a detailed study of the impact of enriched MRs in the performance of GPT-2 models, with the innovation of fine-tuning with these representations—a relatively underexplored approach.

\subsection{Meaning representations for NLG}

NLG inputs are wide and varied, as are their tasks. In the bibliography, we find works about description generation of the information given in a table, summarisation of an article from the news or sentence generation of the meaning encapsulated in a graph \citep{gehrmann2021gem, parikh2020totto, moryossef2019step}, among others. However, DA-based MRs remain the predominant input type used in most dialogue tasks \citep{wu2023diacttod, wang2022task, du2020schema}.

These representations became more important with the rise of neural models and Transformer architectures, which resulted in an increment of the number of annotated datasets \citep{rastogi2020towards,budzianowski2018multiwoz,novikova2017e2e}. For instance, the well-known E2E dataset, annotated in terms of DA, was created as part of the E2E NLG challenge \citep{novikova2017e2e}. This challenge demonstrated the capability of diverse models and architectures to generate fluent and varied outputs using DA-based MRs. \citet{peng2020few} also showed that models capable of transforming these inputs to sentences in one domain can adapt to new domains with a small fine-tuning. Furthermore, these MRs have been employed with relevant contributions in works that employed methodologies like continual learning \citep{mi2020continual} or reinforcement learning \citep{wang2021modelling}. Finally, MRs are a practical way to encapsulate information for the NLG module in dialogue system architectures \citep{vazquez2026role,vazquez2023dialogue,wang2022task}.

Efforts to improve traditional DA-based MRs have led to significant advancements. For instance, \citet{du2020schema} enriched inputs with descriptions about the nature of DAs and slots and \citet{vazquez2023how} transformed DAs into their descriptive forms. These two strategies enhanced semantic accuracy, robustness, and diversity. In a different approach, Abstract Meaning Representations (AMRs) emerged as a novel way to capture the meaning of a sentence by abstracting its syntax, focusing on relationships between concepts and their roles \citep{hryhoryeva2025data,yang2024improving,banarescu2013abstract}. They have also produced inspiring advances in tasks like machine comprehension, machine translation, and summarisation \citep{jin2024analyzing, baptista2024lexicalized,tohidi2022short}. In dialogue, although AMRs are predominantly used in natural language understanding modules \citep{bonial2023abstract, brutti2022abstract}, some studies highlight their potential in improving NLG as well \citep{hryhoryeva2025data,yang2024improving}.

LLMs have revolutionised the field by enabling most NLP tasks to be tackled effectively through prompt-based learning techniques \citep{gu2022ppt,gao2021making, brown2020language}. These methodologies involve augmenting inputs with task-specific instructions, explanations, or task demonstrators to leverage LLMs’ zero-shot and few-shot learning capabilities. Specifically, in this study, we explore new enriched representations that include a task demonstrator in the input. Building on prior work that demonstrated their benefits for a specific domain \citep{vazquez2023how}, we present a novel comparison that explores their impact on multiple domains and datasets.

\subsection{NLG datasets for dialogue}

High-quality datasets have facilitated progress in the research on NLG for dialogue systems. They provide a basic tool to train and evaluate models. Therefore, over the years, researchers have created a variety of datasets tailored to different dialogue tasks, ranging from single-domain scenarios to multi-domain and knowledge-grounded interactions.

\citet{wen2015semantically} and \citet{novikova2017e2e} were pioneers in creating annotated datasets designed to generate task-specific sentences. \citet{wen2015semantically} define a series of DAs and slots to develop dialogue systems that provide information about restaurants and hotels. Similarly, the E2E dataset \citep{novikova2017e2e}, focused on restaurant recommendations, was created to address the problems of classic NLG approaches.

Following these early works, multi-domain datasets were developed to address the growing need for systems capable of handling more diverse and complex dialogues. The Schema-Guided Dialogue dataset \citep{rastogi2020towards} introduced a unified schema for intents and slots that enhances scalability and generalisation across domains, even unseen ones. Similarly, MultiWOZ \citep{budzianowski2018multiwoz} is a large-scale dataset that offers richly annotated dialogues in domains like restaurants, hotels, and transportation. This dataset is an evaluation benchmark for tasks such as MR-to-text generation or dialogue state tracking.  

Recently, the need to create systems that generate knowledge-grounded outputs has resulted in a new trend of datasets. These datasets incorporate information from external sources in their MRs. For instance, ViGGO \citep{juraska2019viggo}, a dataset in the domain of video games, includes the information extracted from WikiData in its MRs as the content of their attribute-value pairs. In an extension of this work, \citet{vazquez2024knowledge} develop a systematic methodology that combines the ViGGO DAs with WikiData knowledge graphs to create a similar dataset, WikiDialog, with seven new domains. Similarly, OpenDialKG, which obtains external information from the database Freebase, generates knowledge-rich dialogues across domains like music, movies, and sports \citep{moon2019opendialkg}.

In this study, we compare datasets spanning single-domain (E2E), multi-domain (MultiWOZ), and knowledge-grounded dialogues (ViGGO). Additionally,  we have incorporated the EMPATHIC corpus \citep{torres2019empathic,montenegro2019dialogue,olaso2021empathic,vazquez2023dialogue,vazquez2026role}. This corpus is extracted from health coaching sessions where a virtual coach helps users change their unhealthy habits in the domain of nutrition. Therefore, we add this scarcely explored and particularly challenging task, coaching, to our already novel and thorough comparison of the impact of an alternative enriched representation across a diverse range of datasets.

\subsection{NLG metrics}

The evaluation metrics for NLG have advanced significantly in recent years \citep{sai2022survey}. However, metrics originally developed for machine translation or summarisation tasks keep being the most widely used for NLG \citep{schmidtova2024automatic, sai2022survey, celikyilmaz2020evaluation}. Metrics such as ROUGE \citep{lin2004rouge}, METEOR \citep{banerjee2005meteor}, NIST \citep{doddington2002automatic}, and mainly BLEU \citep{papineni2002bleu}—which rely on n-gram overlap—are almost universally present in evaluations of NLG systems \citep{vazquez2024knowledge, ramirez2023controllable, wu2023diacttod} and included in the most prominent NLG benchmarks and challenges \citep{gehrmann2021gem, novikova2017e2e, colin2016webnlg}. Consequently, we selected BLEU for our study as a standard metric in NLG and to include a widely used lexical perspective in our evaluation, which complements the other metrics and evaluation approaches considered in this work.

Metrics that evaluate semantic accuracy are also widely used in the NLG community \citep{schmidtova2024automatic, ramirez2023controllable, question2023ruiz, reed2022jurassic}, as they tend to align more closely with human ratings. Among these, BLEURT \citep{sellam2020bleurt} is particularly relevant due to its strong correlation with human evaluations, as it has been explicitly trained with human ratings. In this work, along with BLEURT, we introduce a new dimension to the analysis using LaBSE \citep{feng2022language}. LaBSE, a language-agnostic embedding model, computes semantic similarity as the cosine similarity of sentence embeddings. While conceptually distinct, BLEURT and LaBSE have not yet been compared in detail regarding their performance in prior research.

NLG metrics typically evaluate outputs based on ground-truth references, but metrics that compare the outputs with their corresponding input have gained importance in recent years \citep{schmidtova2024automatic}. In dialogue systems, where MRs containing slots are prevalent, slot error rate or Slot Accuracy is often used to evaluate a generator's robustness \citep{wang2023dspm, du2020schema, li-etal-2020-slot}. This metric determines how many values, usually entities, in the input are present in the output. Recently, some studies have adopted an alternative approach to analyse DA-based input-output alignment: the Dialogue Act Accuracy (DAC) \citep{vazquez2025prompt,vazquez2024knowledge, ramirez2023controllable, vazquez2023dialogue, reed2022jurassic}. In these works, DA classifiers are proposed to assess output quality, where a generated sentence is considered acceptable if the classifier can predict its source DA. Based on the findings of these works, we incorporated the underexplored DAC along with Slot Accuracy to broaden the scope of our analysis.

In summary, our study evaluates NLG engines and representation approaches using a diverse range of automatic metrics, including underexplored dimensions. Additionally, unlike previous works that primarily validate models and methodologies through metrics, we explore metrics' focus and provide a detailed analysis of how these metrics interact with features of representations, datasets and tasks to shape outcomes.

\section{Corpora}
\label{sec:corpora}

The datasets define the domains/tasks in which we validate our proposed representations. As we aim to analyse the suitability of different representations across multiple domains, languages, and corpus characteristics, we selected four datasets that differ in domain, size, lexicon, input variability, and acquisition process, among other features. In particular, we have included two single-domain datasets obtained through crowdsourcing, the E2E dataset with only informative samples \citep{novikova2017e2e} and the ViGGO corpus with more DA variability \citep{juraska2019viggo}, one multi-domain dataset that contains complete simulated and annotated dialogues, the MultiWOZ corpus \citep{budzianowski2018multiwoz}, and a corpus extracted from simulated human-machine health coaching sessions labelled with ad-hoc MRs, the EMPATHIC corpus \citep{torres2019empathic,montenegro2019dialogue,olaso2021empathic,vazquez2023dialogue,vazquez2026role}. These datasets are dialogue-related corpora whose samples consist of MRs and their corresponding example sentences. These MRs, despite differences in corpus characteristics, are all DA-based MRs, where the DA primarily represents the communicative intent of the sentence, while the slot\footnote{The slots can also be referred to as attributes.}–value pairs encapsulate the semantic content. This representation serves as the baseline of our experiments, as we explain in Section~\ref{sec:representations}.  Tables~\ref{tab:e2e-examples}--\ref{tab:empathic-examples} show that this representation structure is shared across all datasets.

\subsubsection*{\textbf{E2E dataset}\footnote{Available free on \url{https://github.com/tuetschek/e2e-dataset}}}

E2E dataset \citep{novikova2017e2e} is a well-known corpus in the domain of restaurants obtained through crowdsourcing. Although it was created for dialogue tasks, it was collected through crowdsourcing by writing sentences for predefined MRs without providing any contextual dialogue. Table \ref{tab:e2e-examples} shows that all the MRs of this dataset use the same DA, \texttt{inform}, with multiple combinations of attributes and values. The two samples in the table inform about two different restaurants, sharing one slot-value pair (\texttt{$eatType=pub$}) and multiple attributes such as \texttt{priceRange} or \texttt{near}, but also presenting attributes (\texttt{customerRating} and \texttt{food}) and values ("The Vault" vs "The Cambridge Bleu" or  "more than £30" vs "cheap") that are different.

\begin{table}[tbh]
\caption{Samples of the E2E dataset.}
\resizebox{\textwidth}{!}{
\begin{tabular}{@{\extracolsep{\fill}}l@{\hskip 0.5in}l}
\toprule
\textbf{MR} & \textbf{Sentence}\\
\midrule
\begin{tabular}[l]{@{}l@{}}inform (name = The Vaults ; eatType = pub ; priceRange = more than £30 ;\\ customerRating = 5 out of 5 ; near = Café Adriatic)\end{tabular} & \begin{tabular}[l]{@{}l@{}}The Vaults pub near Café Adriatic has a 5 star rating. Prices\\start  at £30. \end{tabular} \\ \addlinespace
\begin{tabular}[l]{@{}l@{}}inform ( name = The Cambridge Blue ; eatType = pub ; food = English ; \\ priceRange = cheap ; near = Café Brazil )\end{tabular} & \begin{tabular}[l]{@{}l@{}}Close to Café Brazil, The Cambridge Blue pub serves delicious \\ Tuscan Beef for the cheap price of £10.50. Delicious Pub food. \end{tabular} \\ \bottomrule
\end{tabular}
}
\label{tab:e2e-examples}
\end{table}

\subsubsection*{\textbf{ViGGO corpus}\footnote{Available free on \url{https://nlds.soe.ucsc.edu/viggo} }}

ViGGO corpus \citep{juraska2019viggo} is a dataset in the domain of video games collected with the same protocol as E2E. However, ViGGO contains more variability in terms of DA. Table \ref{tab:ViGGO-examples} presents samples of three of the nine different communicative intentions: inform, give an opinion or ask about a preference for a specific video game characteristic. The attributes and values appear in multiple slot-value pair combinations in the MRs. The attributes are domain-specific, such as \texttt{release\_year} or \texttt{esrb} in the samples of Table \ref{tab:ViGGO-examples}.

\begin{table}[tbh]
\caption{Samples of the ViGGO corpus}
\resizebox{\textwidth}{!}{
\begin{tabular}{@{\extracolsep{\fill}}l@{\hskip 0.5in}l}
\toprule
\textbf{MR} & \textbf{Sentence}\\
\midrule
\begin{tabular}[l]{@{}l@{}}inform ( name = The Forest of Doom ; release\_year = 2014 ; \\ genres = role-playing, text adventure ; has\_multiplayer = no )\end{tabular} & \begin{tabular}[l]{@{}l@{}}The Forest of Doom is a role-playing text adventure game \\ released  in 2014. It is a single-player only game. \end{tabular} \\ \addlinespace
\begin{tabular}[l]{@{}l@{}}give\_opinion ( name = Undertale ; esrb = E (for Everyone) ;\\rating = excellent )\end{tabular} & \begin{tabular}[l]{@{}l@{}}Undertale is a downright excellent game, and it's rated E \\ so everyone can enjoy it. \end{tabular} \\ \addlinespace
\begin{tabular}[l]{@{}l@{}}request\_attribute ( has\_multiplayer = ?  )\end{tabular} & \begin{tabular}[l]{@{}l@{}}Do you prefer to game with others, I mean, multiplayer? \end{tabular} \\ \bottomrule
\end{tabular}
}
\label{tab:ViGGO-examples}
\end{table}

\subsection*{\textbf{MultiWOZ corpus}\footnote{Available free on \url{https://github.com/budzianowski/multiwoz}}}

MultiWOZ corpus \citep{budzianowski2018multiwoz} is a large-scale dataset that contains dialogues covering seven domains: restaurants, hotels, attractions, taxis, trains, hospitals, and police. It includes annotated complete dialogues between a user and an assistant. In these dialogues, the user asks for information in one of these domains, and the assistant, thought to be replaced by an automatic system, provides this information. They are simulated conversations since they were obtained by employing crowdsourcing in a  Wizard-of-Oz (WOZ) framework, where the crowd workers wrote the next turn for a given dialogue context previously created by other workers. As the complete dialogues were annotated, Table \ref{tab:multiwoz-examples} presents turns of users (\texttt{USER} in DAs) and assistants (\texttt{SYSTEM}). Unlike the other datasets, MultiWOZ contains MRs that combine simple DAs, like the first example of Table \ref{tab:multiwoz-examples}. In these cases, the DA is the union of simple DAs, so in this example of the table, the DA is the union of \texttt{SYSTEM\_Booking\_Book} and \texttt{SYSTEM\_Restaurant\_Inform}. This DA is domain-specific, but MultiWOZ also presents open-domain DAs, such as the DA of the second example of the table: \texttt{USER\_General\_Thank}. Finally, the attributes are domain-specific, like \texttt{bookday} or \texttt{address}, and are the entities that appear in the dialogues.

\begin{table}[tbh]
\caption{Samples of the MultiWOZ corpus}
\resizebox{\textwidth}{!}{
\begin{tabular}{@{\extracolsep{\fill}}l@{\hskip 0.5in}l}
\toprule
\textbf{MR} & \textbf{Sentence}\\
\midrule
\begin{tabular}[l]{@{}l@{}}SYSTEM\_Booking\_Book ( bookday = Wednesday ; bookpeople = five ;\\ booktime = 11 am ; name = this restaurant ; ref = DU8IWQZ2 ) \&  \\SYSTEM\_Restaurant\_Inform ( address = 2 Rose Crescent City Centre ; \\ name = The Gardenia ) \end{tabular} & \begin{tabular}[l]{@{}l@{}} The Gardenia is located at 2 Rose Crescent City Centre.  I have \\ booked your table at this restaurant for five people, Wednesday\\ at 11 am. Your reference number is DU8IWQZ2. \end{tabular} \\ \addlinespace
\begin{tabular}[l]{@{}l@{}}USER\_general\_thank (  )\end{tabular} & \begin{tabular}[l]{@{}l@{}} No that will be all, thank you. \end{tabular} \\ \bottomrule
\end{tabular}
}
\label{tab:multiwoz-examples}
\end{table}

\subsection*{\textbf{EMPATHIC corpus\footnote{Available at low cost on \url{https://catalog.elra.info/en-us/repository/browse/ELRA-S0414/}}}}

The EMPATHIC corpus was gathered during the EMPATHIC project \citep{torres2019empathic,montenegro2019dialogue,olaso2021empathic,vazquez2023dialogue, vazquez2026role}. This project aims to change users' unhealthy habits through questions and consequent reflection. For this end, a dialogue system was built to carry out health coaching sessions following a behaviour model, the GROW model \citep{leach2020behavioural}.

\begin{table}[tbh]
\caption{Samples of the EMPATHIC corpus}
\centering
\resizebox{0.7\linewidth}{!}{
\begin{tabular}{@{\extracolsep{\fill}}l@{\hskip 0.5in}l}
\toprule
\textbf{MR} & \textbf{Sentence}\\
\midrule
 Gen\_Hello ( user\_name = Elisabeth ) & Hello Elisabeth! \\ \addlinespace
 Int\_know\_coaching (  ) & \begin{tabular}[c]{@{}l@{}}Do you know what coaching is?\end{tabular}  \\ \addlinespace
 GSQ\_what\_obj ( action = change ) & \begin{tabular}[c]{@{}l@{}}Would you like to change anything?\end{tabular} \\ \addlinespace
 RQ\_curr\_sit ( food = water ) & \begin{tabular}[c]{@{}l@{}}Do you think you're drinking enough water?\end{tabular} \\ \bottomrule
\end{tabular}
}
\label{tab:empathic-examples}
\end{table}

During this project, the EMPATHIC corpus was collected through WoZ experiments. These experiments were simulated coaching sessions where a real user spoke with a wizard who played the role of a virtual coach. From these spoken and complete interactions, only the coach turns were annotated in terms of DAs. Table \ref{tab:empathic-examples} shows four samples of the EMPATHIC corpus, where the ad-hoc MRs designed for this corpus can be observed. This corpus presents two similarities between this dataset and MultiWOZ. First, it contains open-domain DAs such as \texttt{Gen\_Hello} and close-domain DAs such as \texttt{GSQ} or \texttt{RQ}. This second group is highly bound to the GROW behavioural model based on questions, so \texttt{GSQ} and \texttt{RQ} stand for Goal Set Question and Reality Question, respectively. Second, the attributes, which are close-domain, were the entities found in the previously acquired conversations.

\subsection*{\textbf{Corpora features}}

Table \ref{tab:corpus-characteristic} summarises the characteristics of the four datasets and shows how different they are. The first two columns have already been mentioned in the definition of the datasets. So, in terms of domains, 
only E2E and MultiWOZ share a domain, restaurants. Meanwhile, only E2E and ViGGO employed the same protocol for the acquisition processes. 

\begin{table}[htb]
\caption{Characteristics of the four selected corpora. \textit{Domains} stands for the topics the datasets cover. \textit{Acquisition process} means the methodology to obtain the datasets. \textit{Numbers of DAs, attributes, and MRs} quantify the diversity of inputs. \textit{Corpus size} stands for the number of MR-sentence pairs. \textit{Running words} include all the words in the corpus, whereas \textit{Vocabulary} is the number of different words. In the \textit{Number of DAs} for MultiWOZ, 798 represents the total number of DAs that appear as a combination of the 52 simple DAs defined for the corpus.}
\resizebox{\textwidth}{!}{
\begin{tabular}{@{\extracolsep{\fill}}cccccccccc}
\toprule
\textbf{Dataset} & \textbf{Domains} & \textbf{\begin{tabular}[c]{@{}c@{}}Acquisition \\ process\end{tabular}} & \textbf{\begin{tabular}[c]{@{}c@{}}Number of \\ DAs\end{tabular}} & \textbf{\begin{tabular}[c]{@{}c@{}}Number of \\ attributes\end{tabular}} & \textbf{\begin{tabular}[c]{@{}c@{}}Number of \\ MRs\end{tabular}}   & \textbf{\begin{tabular}[c]{@{}c@{}}Corpus \\ size\end{tabular}} & \textbf{\begin{tabular}[c]{@{}c@{}}Running\\ words \end{tabular}} & \textbf{Vocabulary} \\
\midrule
\textbf{E2E} & Restaurants & Crowd & 1 & 8 & 6.0k & 51k & 1.0M & 5.1k \\ \addlinespace
\textbf{ViGGO} & Video games & Crowd & 9 & 14 & 2.3k & 6.9k & 150k & 4.4k \\ \addlinespace
\textbf{MultiWOZ} & 7 domains & WOZ + Crowd & \multicolumn{1}{c}{798 (52)} & 25 & 54k & 140k & 1.9M & 30k \\ \addlinespace
\textbf{\begin{tabular}[c]{@{}c@{}}EMPATHIC \end{tabular}} & Coaching & WOZ & 78 & 21 & 1.2k & 2.3k & 13k & 1.7k \\ \bottomrule
\end{tabular}
}
\label{tab:corpus-characteristic}
\end{table}

The number of DAs and attributes in relation to the number of MRs also makes a difference between the corpora. While E2E only employ one DA and eight attributes for 6k MRs, EMPATHIC employ many more DAs (78) and attributes (21) for a reduced quantity of MRs (1.2k). These numbers represent a big difference in the variability of the MRs.  Only the large-scale MultiWoZ contains higher variability than EMPATHIC in the representations, with 52 simple DAs combined for a total of 798 DAs, due to its special MRs.  

Another important difference among the corpora is their size, i.e, the number of samples. In this regard, MultiWOZ and E2E are much bigger than the other two datasets. 

The datasets also differ in the number of references. The references are the sentences associated with the same MR in the corpus. This feature does not appear in the table but is highly bound with the \textit{Corpus size} and the \textit{Number of MRs}. The first feature is higher than the second for all the datasets, i.e., all the datasets present multiple sentences for the same MR. In Section \ref{sec:distr_effect_results}, we analyse the importance of this dataset characteristic.

Finally, while \textit{Running words} are the total words in the corpus, \textit{Vocabulary} stands for the number of different words. In this regard, E2E adds a low variability of the lexicon og their sentences to its low MR variability with 1M of running words but only 5.1k different words. By contrast, the rest of the datasets show a notable lexicon variability, where, for instance, ViGGO's vocabulary is very close to E2E's with one order less in \textit{Corpus size }and \textit{Running words}.

\section{Meaning representations}
\label{sec:representations}

In this section, we propose different ways to represent the input to feed NLG systems. First, we chose the original MRs of the corpora (Table \ref{tab:e2e-examples}--\ref{tab:empathic-examples}) as \textit{Baseline} representations. As we described in Section~\ref{sec:corpora}, these MRs consist of a DA and a set of slot-value pairs. Next, building on the promising research in \citet{vazquez2023dialogue}, we propose including a task demonstrator in the input connected to the Baseline MR. This Baseline representation is also referred to as \textit{input MR} because it is the MR for the generation in all these representations. Meanwhile, the task demonstrator is a sample extracted from the corpora, consisting of an MR (\textit{demonstrator MR}) and its corresponding sentence (\textit{demonstrator sentence}). These enriched representations change both the input at inference time and the training data.

\begin{figure}[tbh]
    \centering
    \includegraphics[width=\linewidth]{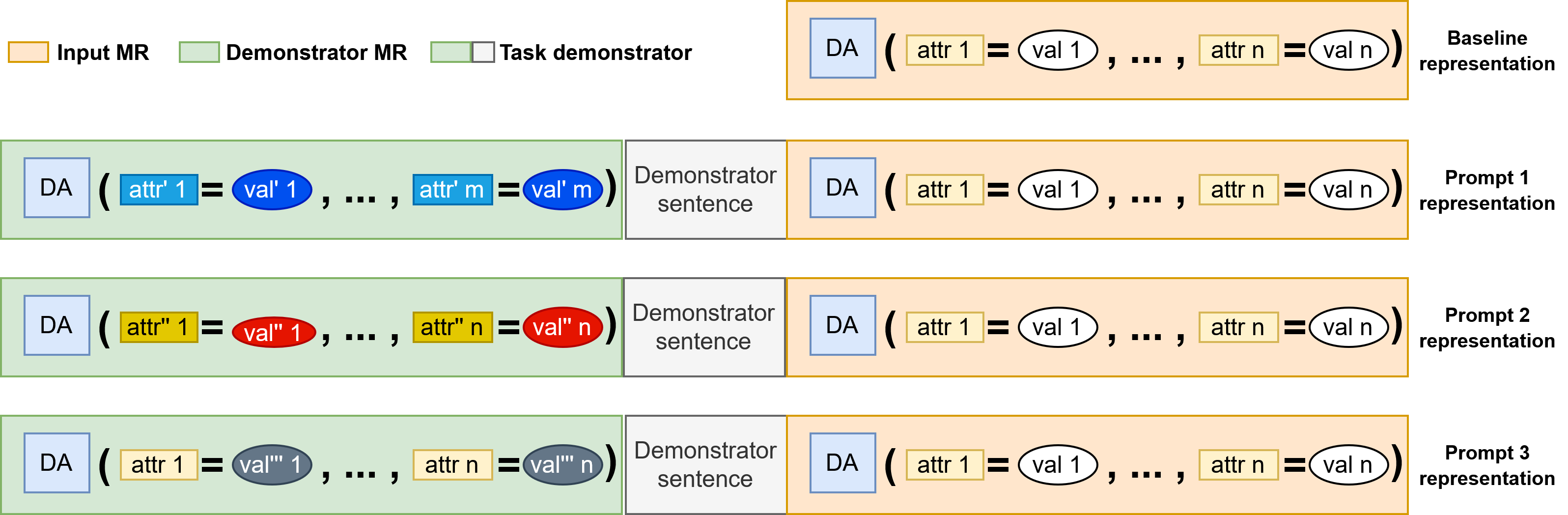}
    \caption{Meaning representations overview. One Baseline representation and three Prompt representations. The Baseline representation, which is the MR found in the original datasets, is the \textit{input MR} (orange boxes) for the generation in all the representations. Prompt representations are created by connecting a task demonstrator with the \textit{input MR}. This task demonstrator includes an MR, the \textit{demonstator MR} (green), and its corresponding example sentence, the \textit{demonstrator sentence} (grey). The colours of the boxes, the letters n and m for the number of the attributes and the use (or not) of apostrophes inside Input and Demonstrator MR boxes are used to indicate which elements are necessarily identical (or not) among the representations.}
    \label{fig:repr_overview}
\end{figure}

Figure \ref{fig:repr_overview} shows a schema of the four representations of the meaning studied in this work. The figure shows that \textit{Prompt 1}, \textit{Prompt 2}, and \textit{Prompt 3} follow the same structure, but they differ in how similar \textit{demonstrator MR}  is to the \textit{input MR}. In this regard, the conditions to select the demonstrator for each Prompt representation are as follows: 

\begin{itemize}
    \item \textbf{Prompt 1:} The DA of the \textit{demonstrator MR} must be the same as the \textit{input MR}, as it is indicated with the light blue boxes in Figure~\ref{fig:repr_overview}.
    \item \textbf{Prompt 2:}  \textit{Demonstrator MR} must include the same DA (light blue boxes) and number of attributes (n) as the \textit{input MR}
    \item \textbf{Prompt 3:} The \textit{Demonstrator MR} must contain the same DA (light blue boxes) and attributes (light yellow boxes) as the \textit{input MR}. 
\end{itemize}

\begin{table}[tbh]
\caption{Examples of task demonstrators linked to an \textit{input MR} for each dataset. DAs and attributes are in bold and coloured to highlight identical elements between the \textit{demonstrator MRs} and the \textit{input MR}. Note that brown colour is employed for those attributes that do not appear in the \textit{input MR}, but they do in any \textit{demonstrator MR}. Prompt 1 was discarded for E2E because the selection methodology would assign the same task demonstrator for all the E2E MRs.}
\centering
\resizebox{\textwidth}{!}{%
\begin{tabular}{@{\extracolsep{\fill}}ccl}
\toprule
\textbf{Dataset} & \textbf{\begin{tabular}[c]{@{}c@{}}Example\\ type\end{tabular}} & \multicolumn{1}{c}{\textbf{Examples}} \\ \midrule
\multirow{6}{*}{\textbf{E2E}} & \textbf{Input MR}  & \begin{tabular}[c]{@{}l@{}}\textbf{\textcolor{Blue}{inform}} ( \textbf{\textcolor{red}{name}} = Zizzi ; \textbf{\textcolor{orange}{eatType}} = coffee shop ; \textbf{\textcolor{magenta}{customer rating}} = high ; \textbf{\textcolor{violet}{near}} = Burger King )\end{tabular}  \\ \cmidrule{2-3}
 & \textbf{\begin{tabular}[c]{@{}c@{}}Prompt 2 \\ demonstrator\end{tabular}}  & \begin{tabular}[c]{@{}l@{}}\textbf{\textcolor{Blue}{inform}} ( \textbf{\textcolor{red}{name}} = Loch Fyne ; \textbf{\textcolor{orange}{eatType}} = restaurant ; \textbf{\textcolor{brown}{familyFriendly}} = yes ; \textbf{\textcolor{brown}{food}} = English ) + A \\ child friendly restaurant that has English food is Loch Fyne.\end{tabular} \\ \cmidrule{2-3} 
 & \textbf{\begin{tabular}[c]{@{}c@{}}Prompt 3\\ demonstrator\end{tabular}}  & \begin{tabular}[c]{@{}l@{}}\textbf{\textcolor{Blue}{inform}} ( \textbf{\textcolor{red}{name}} = Wildwood ; \textbf{\textcolor{violet}{near}} = Café Rouge ; \textbf{\textcolor{magenta}{customer rating}} = low ; \textbf{\textcolor{orange}{eatType}} = restaurant  ) + \\ Wildwood, a restaurant with a low customer rating, is near Café Rouge.\end{tabular} \\ \midrule
\multirow{10}{*}{\textbf{ViGOO}} & \textbf{Input MR}  & \begin{tabular}[c]{@{}l@{}}\textbf{\textcolor{Blue}{give\_opinion}} ( \textbf{\textcolor{red}{name}} = Super Mario World ; \textbf{\textcolor{orange}{player\_perspective}} = side view ; \textbf{\textcolor{magenta}{rating}} = excellent ;\\ \textbf{\textcolor{violet}{has\_multiplayer}} = yes ) \end{tabular}  \\ \cmidrule{2-3} 
 & \textbf{\begin{tabular}[c]{@{}c@{}}Prompt 1\\ demonstrator\end{tabular}}  & \begin{tabular}[c]{@{}l@{}}\textbf{\textcolor{Blue}{give\_opinion}} ( \textbf{\textcolor{red}{name}} = Guitar Hero: Smash Hits ; \textbf{\textcolor{magenta}{rating}} = poor ; \textbf{\textcolor{brown}{developer}} = Beenox ) + Guitar \\ Hero: Smash Hits is just not my kind of game. I find Beenox games are often not very fun.\end{tabular}  \\ \cmidrule{2-3} 
 & \textbf{\begin{tabular}[c]{@{}c@{}}Prompt 2 \\ demonstrator\end{tabular}}  & \begin{tabular}[c]{@{}l@{}}\textbf{\textcolor{Blue}{give\_opinion}} ( \textbf{\textcolor{red}{name}} = Worms: Reloaded ; \textbf{\textcolor{orange}{player\_perspective}} = side view ; \textbf{\textcolor{magenta}{rating}} = average ; \\ \textbf{\textcolor{brown}{available\_on\_Steam}} = yes ) +  Worms: Reloaded is an average quality side view game you can \\ at least find on Steam.\end{tabular}  \\ \cmidrule{2-3} 
 & \textbf{\begin{tabular}[c]{@{}c@{}}Prompt 3\\ demonstrator\end{tabular}}  & \begin{tabular}[c]{@{}l@{}}\textbf{\textcolor{Blue}{give\_opinion}} ( \textbf{\textcolor{red}{name}} = Portal 2 ; \textbf{\textcolor{orange}{player\_perspective}} = first person; \textbf{\textcolor{violet}{has\_multiplayer}} = yes ; \\ \textbf{\textcolor{magenta}{rating}} = excellent ) + I've been having an absolute blast playing Portal 2. First person games \\ are always my favorite, but the multiplayer makes it even better.\end{tabular}  \\ \midrule
\multirow{11}{*}{\textbf{MultiWOZ}} & \textbf{Input MR}  & \begin{tabular}[c]{@{}l@{}}\textbf{\textcolor{Blue}{SYSTEM\_Booking\_Book}} ( \textbf{\textcolor{red}{bookstay}} = 2 ; \textbf{\textcolor{orange}{ref}} = VH33JKKF ) \textbf{\textcolor{Blue}{SYSTEM\_general\_reqmore}} (  )\end{tabular}  \\ \cmidrule{2-3} 
 & \textbf{\begin{tabular}[c]{@{}c@{}}Prompt 1 \\  prompt\end{tabular}}  & \begin{tabular}[c]{@{}l@{}}\textbf{\textcolor{Blue}{SYSTEM\_Booking\_Book}} ( \textbf{\textcolor{orange}{ref}} = GS2CO4IV ) \textbf{\textcolor{Blue}{SYSTEM\_general\_reqmore}} (  ) + Your table is \\ booked and will be reserved for 15 minutes. Your reference number is GS2CO4IV. May I help \\ you with anything else today?\end{tabular}  \\ \cmidrule{2-3} 
 & \textbf{\begin{tabular}[c]{@{}c@{}}Prompt 2 \\ demonstrator\end{tabular}}  & \begin{tabular}[c]{@{}l@{}}\textbf{\textcolor{Blue}{SYSTEM\_Booking\_Book}} ( \textbf{\textcolor{brown}{name}} = Cambridge Belfry ; \textbf{\textcolor{orange}{ref}} = LJOEY6H2 ) \\  \textbf{\textcolor{Blue}{SYSTEM\_general\_reqmore}} (  ) + Okay. Your booking at the Cambridge Belfry was successful. \\ The reference number is LJOEY6H2 .  Can I help you with anything else today?\end{tabular} \\ \cmidrule{2-3} 
 & \textbf{\begin{tabular}[c]{@{}c@{}}Prompt 3\\ demonstrator\end{tabular}}  & \begin{tabular}[c]{@{}l@{}}\textbf{\textcolor{Blue}{SYSTEM\_Booking\_Book}} ( \textbf{\textcolor{red}{bookstay}} = 4 ; \textbf{\textcolor{orange}{ref}} = 3VUMBFZ0 )  \textbf{\textcolor{Blue}{SYSTEM\_general\_reqmore}} (  ) \\+ Great, you're booked for 4 nights with reference number 3VUMBFZ0 . May I help you with \\ anything else? \end{tabular} \\ \midrule
\multirow{8}{*}{\textbf{EMPATHIC}} & \textbf{Input MR}  & \textbf{\textcolor{Blue}{RQ\_curr\_sit}} ( \textbf{\textcolor{red}{action}} = cook ) \\ \cmidrule{2-3}
 & \textbf{\begin{tabular}[c]{@{}c@{}}Prompt 1\\ demonstrator\end{tabular}}  & \textbf{\textcolor{Blue}{RQ\_curr\_sit}}  ( \textbf{\textcolor{red}{action}} = tell ; \textbf{\textcolor{brown}{freq}} = daily ) + Can you tell me about your daily eating habits? \\ \cmidrule{2-3}
 & \textbf{\begin{tabular}[c]{@{}c@{}}Prompt 2 \\ demonstrator\end{tabular}}  & \textbf{\textcolor{Blue}{RQ\_curr\_sit}} ( \textbf{\textcolor{brown}{food}} = water ) + And do you think you drink enough water? \\ \cmidrule{2-3} 
 & \textbf{\begin{tabular}[c]{@{}c@{}}Prompt 3\\ demonstrator\end{tabular}}  & \textbf{\textcolor{Blue}{RQ\_curr\_sit}} ( \textbf{\textcolor{red}{action}} = change ) + Would you like to change your eating habits? \\ \bottomrule
\end{tabular}%
}
\label{tab:prompt_examples}
\end{table}

These conditions for selecting a demonstrator make task demonstrators from \textit{Prompt 3} more specific to the \textit{Baseline} than those from \textit{Prompt 2}, and in turn, those from \textit{Prompt 2} more specific than those from \textit{Prompt 1}.
Table \ref{tab:prompt_examples} presents real examples, one for each dataset, of \textit{input MRs} (first row of each dataset) with their corresponding demonstrator for the three Prompt representations (next rows). They illustrate that the conditions stated above are satisfied and show the specificity of the demonstrators for each proposed representation. In MultiWOZ, for instance, the DA for the Baseline representation and for the \textit{Demonstrator MR} of the three enriched representations is \texttt{SYSTEM\_Booking\_Book \& SYSTEM\_general\_reqmore}. Next, the number of attributes of the Input MR in this dataset is two, which matches the number of them in the corresponding \textit{Demonstrator MR} of Prompt 2 and Prompt 3. Finally, \texttt{bookstay} and \texttt{ref} (highlighted in red and orange in all the rows of this dataset) are the attributes for the Baseline representation and the Demonstrator MR of Prompt 3. By contrast, Prompt 1 and Prompt 2 present attributes that do not appear in the original representation (highlighted in brown). The same restrictions are followed in the other datasets. Prompt 1 was discarded for E2E because the selection methodology, which we explain next, would assign the same task demonstrator for all the E2E MRs.

\begin{table}[tbh]
\caption{Number of different demonstrators in each dataset for the three Prompt representations.}
\centering
\resizebox{0.5\textwidth}{!}{
\begin{tabular}{@{\extracolsep{\fill}}cccc}
\toprule
\textbf{Dataset} & \textbf{\begin{tabular}[c]{@{}c@{}}Prompt 1\end{tabular}} & \textbf{\begin{tabular}[c]{@{}c@{}}Prompt 2\end{tabular}} & \textbf{\begin{tabular}[c]{@{}c@{}}Prompt 3\end{tabular}} \\
\midrule
\textbf{E2E} & - \footnotemark & 6 & 120 \\ \addlinespace
\textbf{ViGOO} & 9 & 21 & 473 \\ \addlinespace
\textbf{MultiWOZ} & 798 & 2281 & 11644 \\ \addlinespace
\textbf{\begin{tabular}[c]{@{}c@{}}EMPATHIC\end{tabular}} & 78 & 197 & 537 \\ \bottomrule
\end{tabular}%
}
\label{tab:prompt-size}
\end{table}

In this proposal, we also aim to reduce the number of different demonstrators in training, making our approach extensible even to datasets with reduced size. Thus, our selection methodology of demonstrators assigns the same demonstrators to all MRs that satisfy identical conditions. In other words, all the MRs from the original dataset with the same DA are connected to the same demonstrator for \textit{Prompt 1}, similarly, those with the same DA and number of attributes for \textit{Prompt 2}, and the same DA and exact attributes for \textit{Prompt 3}. Therefore, the number of different demonstrators in the training set matches the number of DAs for \textit{Prompt 1}, the combination of DAs and the number of attributes for \textit{Prompt 2}, and the combinations of DAs and attributes for \textit{Prompt 3}.

Table \ref{tab:prompt-size} shows the number of demonstrators for each dataset in each Prompt representation. This table confirms that the number of demonstrators for \textit{Prompt 1} matches the number of DAs (Table~\ref{tab:corpus-characteristic}). Consequently, the number of different demonstrators for \textit{Prompt 1} in E2E would be one. So, we discarded this representation for this dataset since all \textit{input MRs} would have been linked to the same demonstrator. In addition, as expected, the more specific the demonstrators, the higher the number of them, i.e., Prompt 3 always contains more than Prompt 2, which, in turn, contains more than Prompt 1. Lastly, the corpora with more MR variability, such as MultiWOZ and EMPATHIC, employ more demonstrators than, for instance, E2E, with a very low variability in this regard.

\footnotetext{There is no Prompt 1 representation for E2E, because E2E only has one DA, so all the MRs from the original corpus would be connected to the same demonstrator for this representation.}

\section{Metrics}
\label{sec:metrics}

The metrics play an important role in this work. We aim to explore what each metric can detect and understand how the features of tasks, corpora and input representations affect the model evaluation with each metric. To this end, we have selected a range of metrics that allow us to evaluate different aspects of the outputs: lexical, semantics or coherence with the input, among others. They can be split into reference-driven and reference-less metrics and are as follows:

\begin{itemize}
    \item \textbf{Reference-driven metrics:} Metrics whose scores are obtained by comparing each generated sentence with the multiple references found in the corpus. We have selected three reference-driven metrics that cover lexical and semantic aspects as well as, to some extent, human criteria. 
    \begin{itemize}
        \item \textbf{BLEU \citep{papineni2002bleu}}: Metric universally employed in the NLG-related works, although it was originally created for the translation task. It is a good indicator of lexical accuracy as its evaluation is based on n-gram overlapping. In this work, we employ BLEU-4 (with n-grams from uni-grams to 4-grams) with a smoothing function \citep{chen2014systematic}.
        \item \textbf{BLEURT \citep{sellam2020bleurt}}: Machine-learned metric that aims to replicate human evaluation to some extent, thanks to its training process. This process employed a BERT model trained with a public collection of 430k sentence pairs and ratings based on human perception for each pair. As a result,  the model returns a similarity value for each pair of sentences. 
        \item \textbf{LaBSE \citep{feng2022language}}: Metric based on sentence embeddings. They are obtained with a BERT dual encoder characterized by being a language-agnostic model that supports more than 100 languages. The metric scores are the cosine similarity between the sentence embeddings of generation and references. 
    \end{itemize}
    \item \textbf{Referenceless metrics:} They can also be named source-based metrics because they compare the generated sentence (the output) with the MR (the input). In this case, we have selected two metrics that evaluate the coherence of the generation in terms of semantic content and communicative intention. 
    \begin{itemize}
        \item \textbf{Slot Accuracy \citep{li-etal-2020-slot}}: Metric that focuses on the values associated with the attributes in the MRs. It evaluates whether the model includes them in the generated sentences. So, the score is the percentage of values in the input that appear in the output, giving an indicator of the robustness of the NLG in terms of how much input information is communicated in the output. Note that this metric does not include generated sentences obtained from MR without attributes. Therefore, the evaluation set could be smaller than the one used in other metrics. 
        \item \textbf{Dialogue Act Accuracy \citep{ramirez2023controllable}:} The DAC is an underexplored metric that quantifies the accuracy of classifiers to predict the source DA of a generated sentence. The scores are the results of classification problems where 
        the inputs are the generated sentences, and the outputs are the DAs. In this work, four corpora were selected that differ in the number and typology of DAs. So, different types of DA classifiers were trained. There was no classifier for E2E because it only has one DA. Then, ViGGO and EMPATHIC classifiers are multiclass, as only one DA can be assigned to each generated sentence. By contrast, the MultiWOZ classifier is multilabel because its MR accept DAs that are a combination of simple DAs. For the classifiers, we employed DistilBERT models \citep{sanh2019distilbert} because they have demonstrated
        great training capabilities despite their reduced size. In addition, we split the original datasets in a proportion of 95/5 for training and validation, where the validation set gave us reference values to determine when a generation is reasonably good in this regard. 
    \end{itemize}
\end{itemize}

\section{Results \& discussion}
\label{sec:results}

This section presents a thorough comparative study that defines the objective of our work. This comparison allows us to analyse three different aspects: a) which representation of the meaning best leverages the capabilities of the generative models, b) the impact of these meaning representations on the generation quality across datasets, and c) the influence of dataset and task features in the model evaluation with each metric. Therefore, we will have a comprehensive view of how these three elements — representations, datasets, and metrics — interact and affect the final results.

In general terms, the study has shown that enriched representations perform particularly well on small datasets involving complex tasks and high variability in MRs, as well as the lexicon and semantics of the sentences. However, they can act as distractors if the baseline MR is already long and contains a lot of semantic information. These enriched representations are also effective in zero-shot\footnote{Strictly speaking, when using representations that include one demonstrator, these experiments correspond to a one-shot setting rather than a pure zero-shot one. Here, we use the term ``zero-shot'' to emphasise that no additional task-specific training or fine-tuning is performed.} settings, obtaining better results than the Baseline before training. The analysis of the metrics reveals that those that evaluate semantic aspects capture output quality more accurately than those that focus on the lexicon. Finally, the generative models stand out for their capacity to adapt to different tasks and for their robustness in handling the input at both semantic and communicative intention levels.

To explain this detailed study with the highest clarity, we structure the section as follows. First,  we present the experimental framework for training the models and generating the sentences in Section \ref{sec:exp_framework}. Next, we show and discuss the results obtained in the evaluation of the generated sentences. The analysis of the results is split into results with reference-driven (Section \ref{sec:results_with_references}) and referenceless metrics (Section \ref{sec:results_without_references}). Next, Section \ref{sec:distr_effect_results} goes deeper into evaluating the importance of dataset features in the results, exploring specific corpus features such as the number of references and attributes per MR. Finally, Section \ref{sec:generation_analysis} describes a human study of the generations that allows us to determine the role of metrics and representations in the results.

\subsection{Experimental framework}
\label{sec:exp_framework}

We selected one of the pre-trained GPT-2 models \citep{radford2019language} due to their acceptable computational requirements and their demonstrated suitability for fine-tuning and prompt-based learning approaches \citep{peng2020few,liu2023large,vazquez2023dialogue}. In particular, we perform experiments with the GPT-2 Medium model to validate the proposed MRs and compare their impact on the generations across domains.

These experiments involve fine-tuning procedures with 5-fold cross-validation, ensuring that the models' performances are rigorously evaluated across different subsets of the data. For consistency with the pre-training setup, we utilise the same byte-pair tokeniser \citep{sennrich2016neural} during the fine-tuning process. In training, we perform over five epochs using a learning rate scheduler with a linear warm-up that starts 5e-5, a batch size of 8 and an Adam optimiser with weight decay \citep{kingma2014adam}. For the generation phase, we generate five distinct outputs per MR. Each output is constrained to a maximum length of 80 tokens, whereas we set the temperature to 1.0, stimulating output variability.

\subsection{Analysis of the results with reference-driven metrics}
\label{sec:results_with_references}

To assess how well the different representations support generation quality across domains, we begin by analysing the scores obtained for the reference-driven metrics, namely BLEU, BLEURT, and LaBSE. These metrics calculate their scores by comparing the generated outputs, five per MR, with the references related to such MRs in the original corpora. This analysis offers insight into the behaviour of enriched and baseline representations across different datasets, using as evaluation criteria metrics that rely on such predefined outputs.

\begin{figure}[htb]
    \makebox[\linewidth][c]{
     \centering
     \begin{subfigure}[b]{0.29\textwidth}
         \centering
         \includegraphics[width=\textwidth]{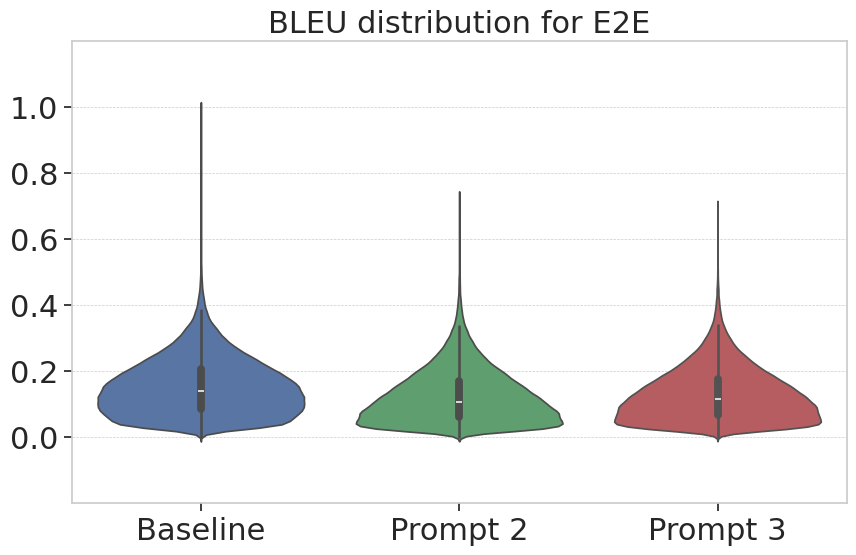}
         \caption{E2E (\textbf{BLEU})}
         \label{fig:bleu_distribution_e2e}
     \end{subfigure}
    \hfill
    \begin{subfigure}[b]{0.29\textwidth}
         \centering
         \includegraphics[width=\textwidth]{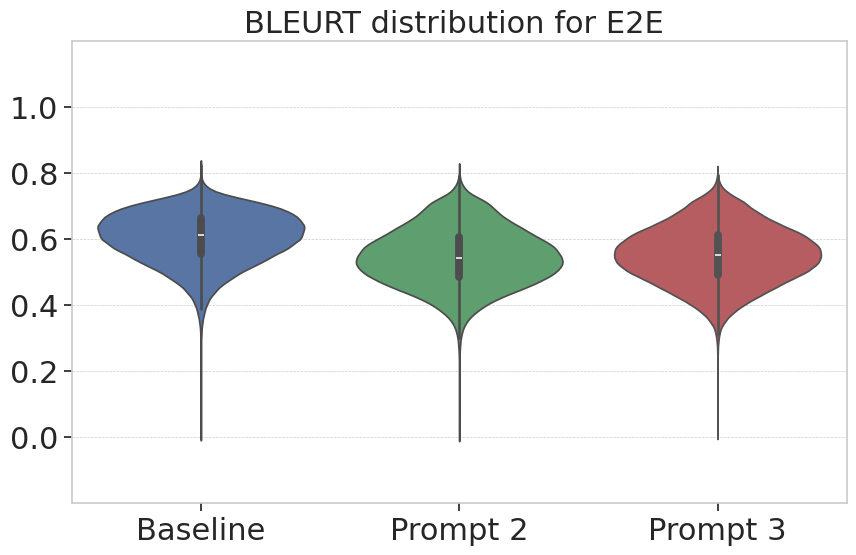}
         \caption{E2E (\textbf{BLEURT})}
         \label{fig:bleurt_distribution_e2e}
    \end{subfigure}
    \hfill
    \begin{subfigure}[b]{0.29\textwidth}
         \centering
         \includegraphics[width=\textwidth]{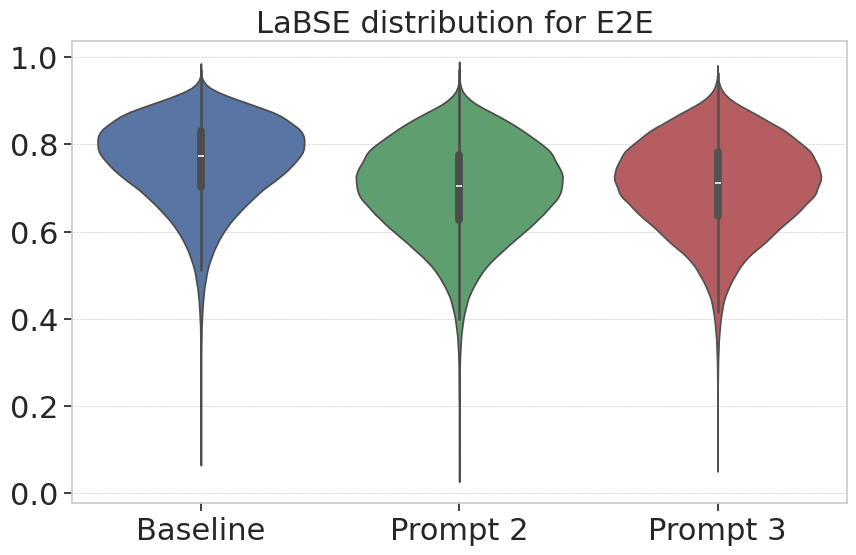}
         \caption{E2E (\textbf{LaBSE})}
         \label{fig:labse_distribution_e2e}
    \end{subfigure}
    }
    \makebox[\linewidth][c]{
     \centering
     \begin{subfigure}[b]{0.29\textwidth}
         \centering
         \includegraphics[width=\textwidth]{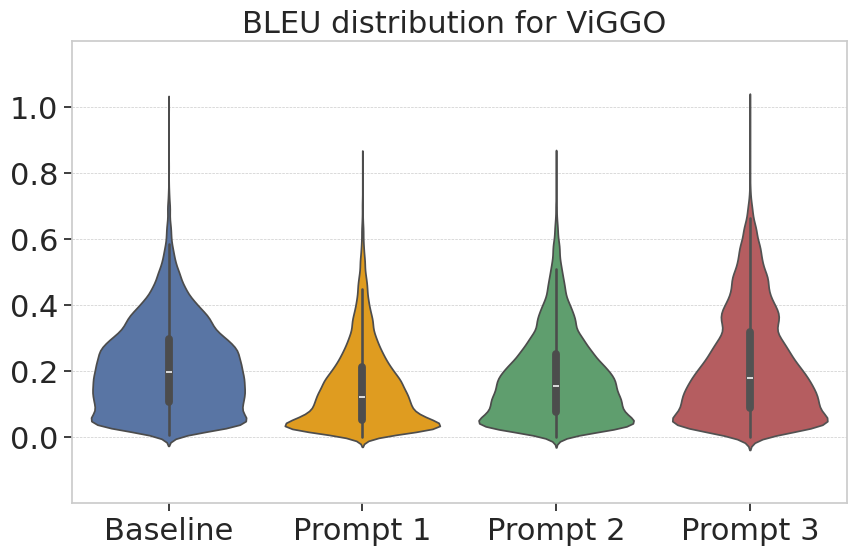}
         \caption{ViGGO (\textbf{BLEU})}
         \label{fig:bleu_distribution_viggo}
    \end{subfigure}
    \hfill
    \begin{subfigure}[b]{0.29\textwidth}
         \centering
         \includegraphics[width=\textwidth]{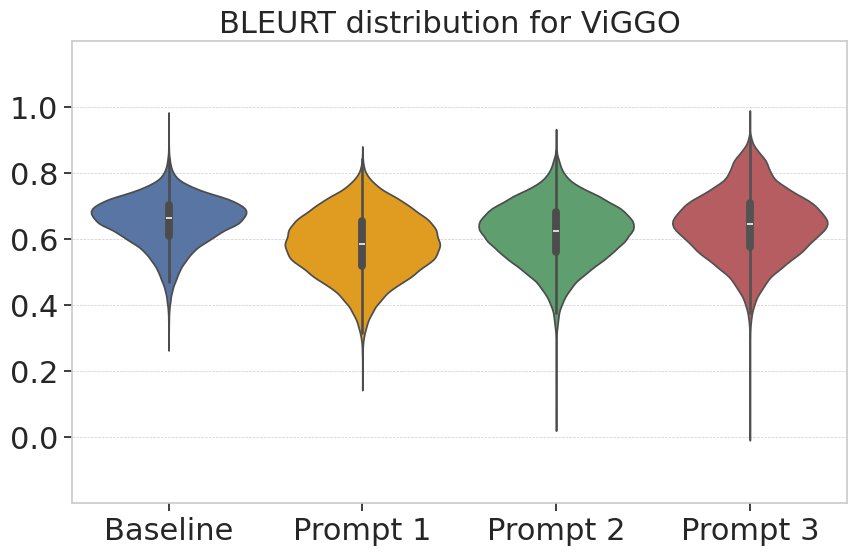}
         \caption{ViGGO (\textbf{BLEURT})}
         \label{fig:bleurt_distribution_viggo}
    \end{subfigure}
    \hfill
    \begin{subfigure}[b]{0.29\textwidth}
         \centering
         \includegraphics[width=\textwidth]{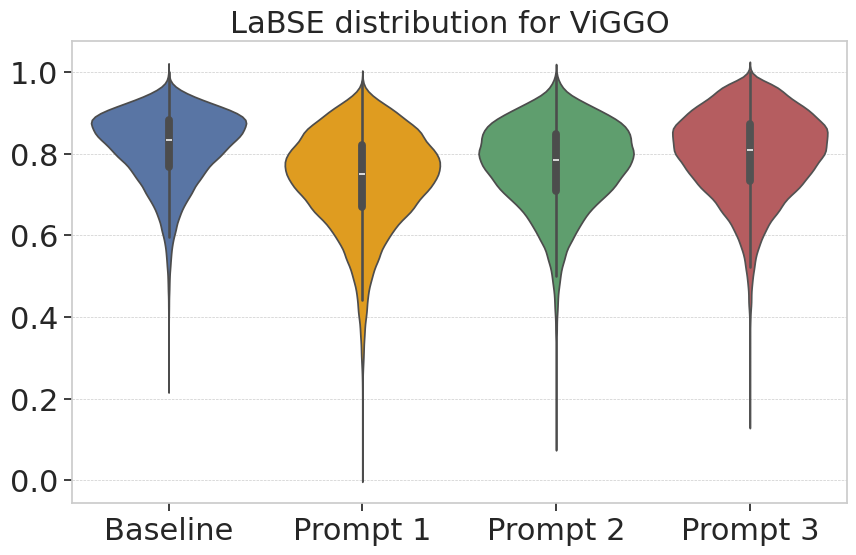}
         \caption{ViGGO (\textbf{LaBSE})}
         \label{fig:labse_distribution_viggo}
    \end{subfigure}
    }
    \makebox[\linewidth][c]{
     \centering
    \begin{subfigure}[b]{0.29\textwidth}
         \centering
         \includegraphics[width=\textwidth]{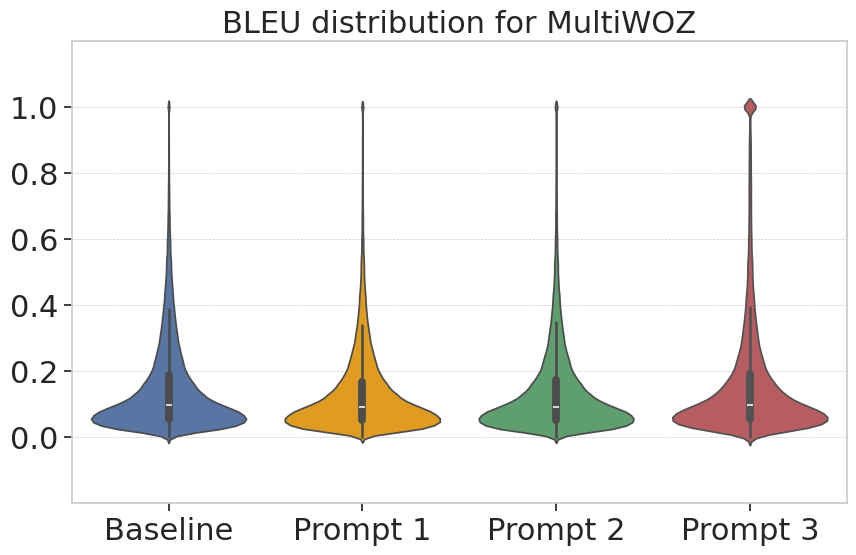}
         \caption{MultiWOZ (\textbf{BLEU})}
         \label{fig:bleu_distribution_multiwoz}
    \end{subfigure}
    \hfill
    \begin{subfigure}[b]{0.29\textwidth}
         \centering
         \includegraphics[width=\textwidth]{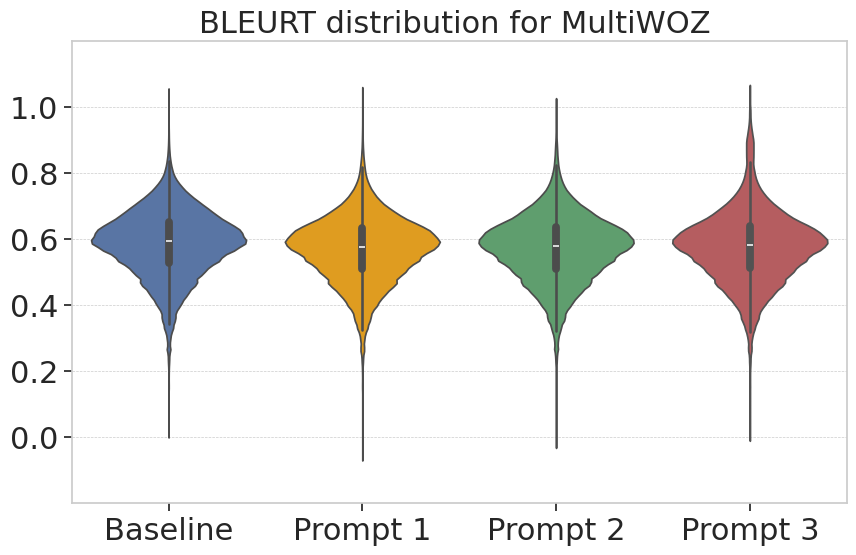}
         \caption{MultiWOZ (\textbf{BLEURT})}
         \label{fig:bleurt_distribution_multiwoz}
    \end{subfigure}
    \hfill
    \begin{subfigure}[b]{0.29\textwidth}
         \centering
         \includegraphics[width=\textwidth]{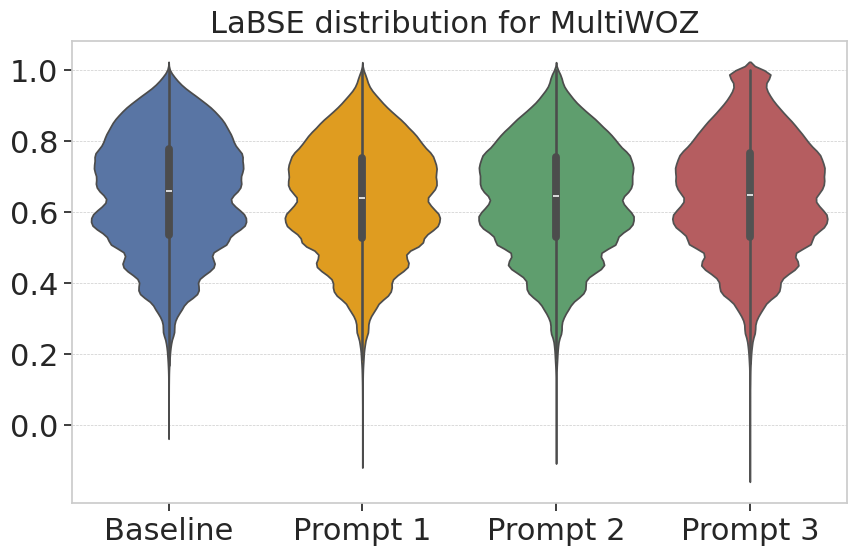}
         \caption{MultiWOZ (\textbf{LaBSE})}
         \label{fig:labse_distribution_multiwoz}
    \end{subfigure}
    }
    \makebox[\linewidth][c]{
    \centering
    \begin{subfigure}[b]{0.29\textwidth}
         \centering
         \includegraphics[width=\textwidth]{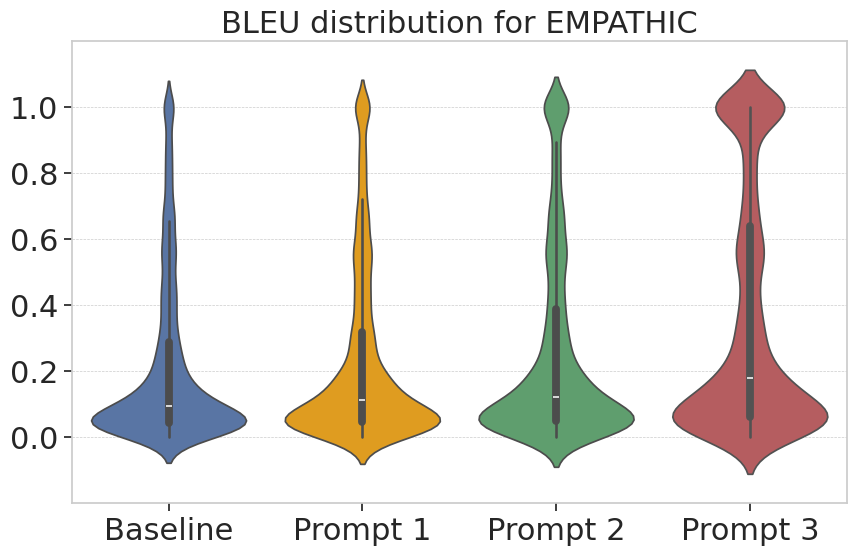}
         \caption{EMPATHIC (\textbf{BLEU})}
         \label{fig:bleu_distribution_empathic}
    \end{subfigure}
    \hfill
    \begin{subfigure}[b]{0.29\textwidth}
         \centering
         \includegraphics[width=\textwidth]{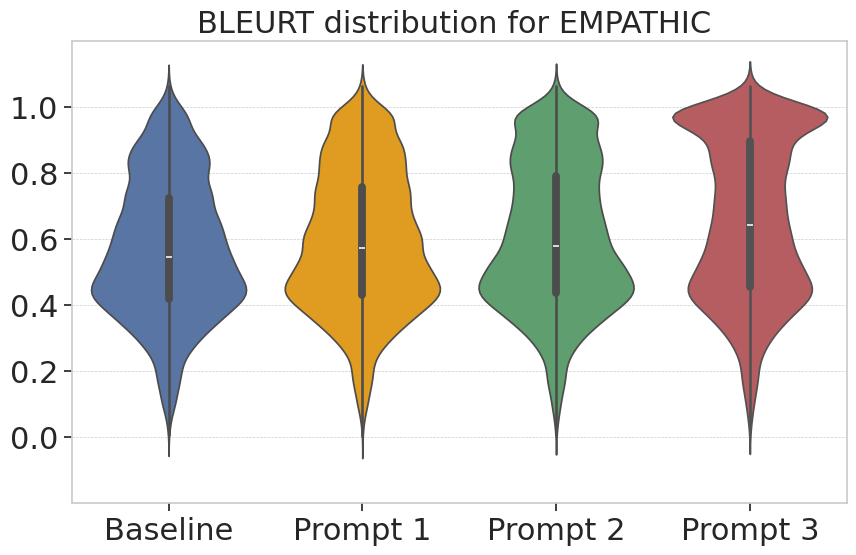}
         \caption{EMPATHIC (\textbf{BLEURT})}
         \label{fig:bleurt_distribution_empathic}
    \end{subfigure}
    \hfill
    \begin{subfigure}[b]{0.29\textwidth}
         \centering
         \includegraphics[width=\textwidth]{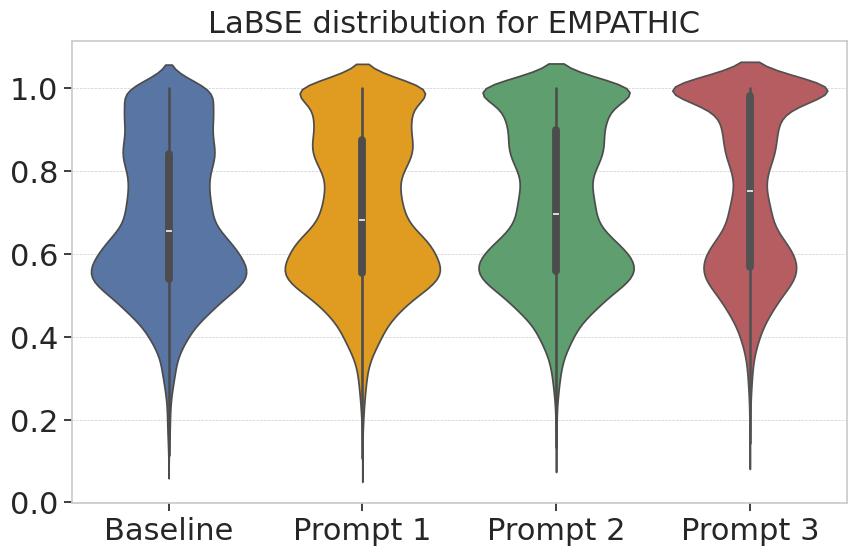}
         \caption{EMPATHIC (\textbf{LaBSE})}
         \label{fig:labse_distribution_empathic}
    \end{subfigure}
    }
    \caption{Distribution of the \textit{MR scores} of each reference-driven metric after the fine-tuning for each representation in each dataset.}
\label{fig:distributions_with_references}
\end{figure}

To obtain the results for each metric, first, a score (\textit{generation score}) is computed for each one of the five generations per input by comparing them separately with all the references. The scores for each MR (\textit{MR scores}) are the average of the five \textit{generation scores}. Finally, the \textit{average scores} for each metric is the average over all the \textit{MR scores}. These scores are calculated for each combination of metrics, representations, and datasets. Note that in this analysis, we focus on the \textit{average scores} after fine-tuning (Table~\ref{tab:score_results_with_references}) and the evolution of these scores before training, epoch 0, and after each epoch (Figure~\ref{fig:evolutions_with_reference}), as well as the distribution of the \textit{MR scores} for the outputs generated with the fine-tuned models (Figure~\ref{fig:distributions_with_references}).

\subsubsection*{\textbf{Average scores after fine-tuning}}

Tables~\ref{tab:bleu_results},~\ref{tab:bleurt_results}, and~\ref{tab:labse_results} show the scores obtained after fine-tuning the models with each representation-dataset pair for BLEU, BLEURT, and LaBSE, respectively. These tables provide an initial overview of the models' performance differences across metrics and allow us to identify which representation obtains the best results (denoted in bold) for each dataset.

\begin{table}[htb]
    \centering
    \caption{\textit{Average scores} after fine-tuning for each reference-driven metric, representation, and dataset. Base corresponds to the baseline representation, whereas Pi is the Prompt i representation with i=1,2,3. The datasets (Data) are E2E, ViGGO (ViG), MultiWOZ (MWOZ), and EMPATHIC (EMP). No \textit{Prompt 1} for E2E. The best score for each dataset is denoted in bold.}
    \label{tab:score_results_with_references}
    \begin{subtable}{0.30\linewidth}
        \centering
        \caption{\textbf{BLEU}}
        \label{tab:bleu_results}
        \resizebox{\textwidth}{!}{%
        \begin{tabular}{@{\extracolsep{\fill}}ccccc}
        \toprule
        \multicolumn{1}{c}{
        \textbf{Data}} & \textbf{Base} & \textbf{P1} & \textbf{P2} & \textbf{P3} \\ \midrule
        \multicolumn{1}{c}{\textbf{E2E}} & \textbf{0.15} & - & 0.12 & 0.13  \\ \addlinespace 
        \multicolumn{1}{c}{\textbf{ViG}} & \textbf{0.21} & 0.15 & 0.17 & 0.19 \\ \addlinespace 
        \multicolumn{1}{c}{\textbf{MWOZ}} & 0.16 & 0.14 & 0.14 & \textbf{0.17} \\ \addlinespace
        \multicolumn{1}{c}{\textbf{\begin{tabular}[c]{@{}c@{}}EMP\end{tabular}}} & 0.17 & 0.19 & 0.20 & \textbf{0.26} \\ \bottomrule
        \end{tabular}
        }
    \end{subtable}%
    \hfill
    \begin{subtable}{0.30\linewidth}
        \centering
        \caption{\textbf{BLEURT}}
        \label{tab:bleurt_results}
        \resizebox{\textwidth}{!}{%
        \begin{tabular}{@{\extracolsep{\fill}}ccccc}
        \toprule
        \multicolumn{1}{c}{
        \textbf{Data}} & \textbf{Base} & \textbf{P1} & \textbf{P2} & \textbf{P3} \\ \midrule
        \multicolumn{1}{c}{\textbf{E2E}} & \textbf{0.60} & - & 0.54 & 0.54  \\ \addlinespace 
        \multicolumn{1}{c}{\textbf{ViG}} & \textbf{0.65} & 0.58 & 0.61 & 0.63 \\ \addlinespace 
        \multicolumn{1}{c}{\textbf{MWOZ}} & \textbf{0.60} & 0.57 & 0.58 & 0.59 \\ \addlinespace
        \multicolumn{1}{c}{\textbf{\begin{tabular}[c]{@{}c@{}}EMP\end{tabular}}} & 0.55 & 0.56 & 0.57 & \textbf{0.60} \\ \bottomrule
        \end{tabular}
        }
    \end{subtable}%
    \hfill
    \begin{subtable}{0.30\linewidth}
        \centering
        \caption{\textbf{LaBSE}}
        \label{tab:labse_results}
        \resizebox{\textwidth}{!}{%
        \begin{tabular}{@{\extracolsep{\fill}}ccccc}
        \toprule
        \multicolumn{1}{c}{
        \textbf{Data}} & \textbf{Base} & \textbf{P1} & \textbf{P2} & \textbf{P3} \\ \midrule
        \multicolumn{1}{c}{\textbf{E2E}} & \textbf{0.76} & - & 0.69 & 0.70  \\ \addlinespace 
        \multicolumn{1}{c}{\textbf{ViG}} & \textbf{0.82} & 0.73 & 0.77 & 0.79 \\ \addlinespace 
        \multicolumn{1}{c}{\textbf{MWOZ}} & \textbf{0.68} & 0.66 & 0.67 & 0.67 \\ \addlinespace
        \multicolumn{1}{c}{\textbf{\begin{tabular}[c]{@{}c@{}}EMP\end{tabular}}} & 0.66 & 0.68 & 0.68 & \textbf{0.70} \\ \bottomrule
        \end{tabular}
        }
    \end{subtable}%
\end{table}

\paragraph{\textbf{Low BLEU results}} 
Table \ref{tab:score_results_with_references} shows very low results for BLEU. Its values (no higher than 0.3) are much lower than the results obtained with the other two metrics (no lower than 0.5). This discrepancy with the other two metrics suggests that lexical similarity based on n-gram overlap may not adequately quantify the output quality.

\paragraph{\textbf{Demonstrators effective in small datasets with complex tasks}} Table \ref{tab:score_results_with_references} shows that Prompt representations are effective in datasets with the special characteristics of EMPATHIC. It seems that the demonstrators take more relevance when the dataset is small, with a high variety in the input and the output, and a complex task. For E2E, ViGGO and MultiWOZ, the results with the enriched representations are satisfactory, but the models perform better with the Baseline MR. 

\paragraph{\textbf{Prompt 3: the best prompt}} Prompt representations obtain better results when they include more specific task demonstrators in their input, across all datasets. In other words, \textit{Prompt 3} consistently outperforms \textit{Prompt 2}, which in turn is better than \textit{Prompt 1}. These more specific representations require a higher number of demonstrators during training, as shown in Table~\ref{tab:prompt-size} of Section~\ref{sec:representations}. This may suggest that good performance depends on having a sufficient number of demonstrators. However, the results obtained with \textit{Prompt 1} for EMPATHIC show that enriched representations can still be effective even when the number of training demonstrators is low.

\subsubsection*{\textbf{Distribution of the scores}}

We introduce in this section a new type of analysis: the distribution of the scores assigned to each MR, i.e., the distribution of the \textit{MR scores}. This analysis enables us to determine whether the average values reported in the previous subsection stem from consistent behaviour across all examples or from a combination of highly variable outputs. In other words, it helps determine whether the observed averages reflect normally distributed scores around a central value or instead emerge from large disparities between high- and low-quality generations.

Figure~\ref{fig:distributions_with_references} presents these score distributions for each combination of metric, dataset, and representation. Each row in the figure corresponds to a specific dataset (E2E, ViGGO, MultiWOZ, and EMPATHIC, in that order), and each column corresponds to a specific metric (BLEU, BLEURT, and LaBSE, respectively). Within each subplot, the distribution of the \textit{MR scores} is plotted separately for each representation.

\paragraph{\textbf{Metrics, the main difference}} The distributions' shapes 
are mainly associated with the metric. While BLEU is characterised by having most of its values close to 0.0, BLEURT and LaBSE generally follow Gaussian distributions with a higher average for LaBSE.

\begin{figure}[htb]
    \makebox[\linewidth][c]{
     \centering
     \begin{subfigure}[b]{0.29\textwidth}
         \centering
         \includegraphics[width=\textwidth]{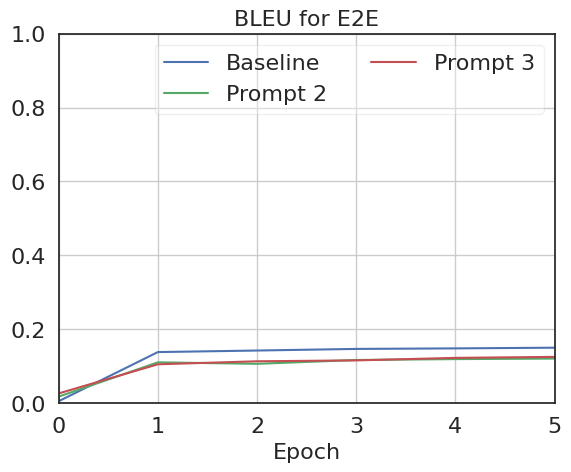}
         \caption{E2E (\textbf{BLEU})}
         \label{fig:bleu_evolution_e2e}
    \end{subfigure}
    \hfill
    \begin{subfigure}[b]{0.29\textwidth}
         \centering
        \includegraphics[width=\textwidth]{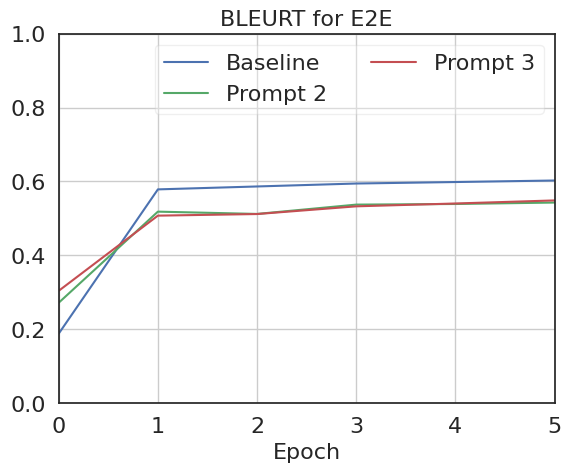}
        \caption{E2E (\textbf{BLEURT})}
        \label{fig:bleurt_evolution_e2e} 
    \end{subfigure}
    \hfill
    \begin{subfigure}[b]{0.29\textwidth}
         \centering
        \includegraphics[width=\textwidth]{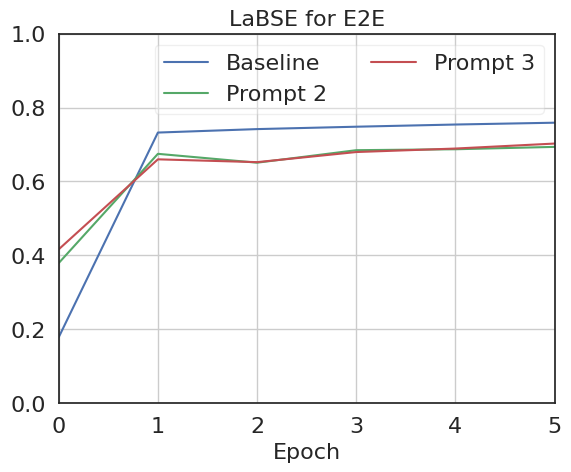}
         \caption{E2E (\textbf{LaBSE})}
         \label{fig:labse_evolution_e2e}
    \end{subfigure}
    }
    \makebox[\linewidth][c]{
     \centering
     \begin{subfigure}[b]{0.29\textwidth}
         \centering
         \includegraphics[width=\textwidth]{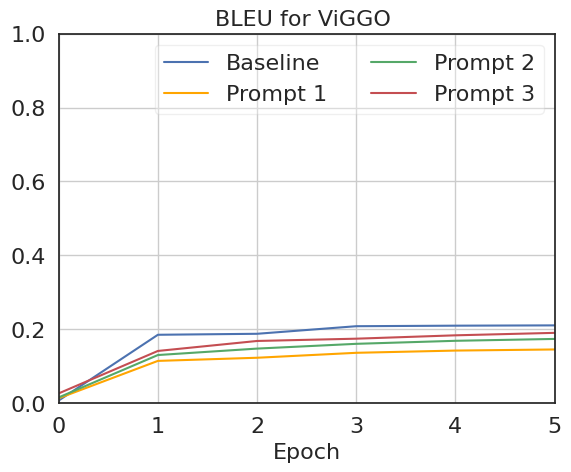}
         \caption{ViGGO (\textbf{BLEU})}
         \label{fig:bleu_evolution_viggo}
    \end{subfigure}
    \hfill
    \begin{subfigure}[b]{0.29\textwidth}
         \centering
         \includegraphics[width=\textwidth]{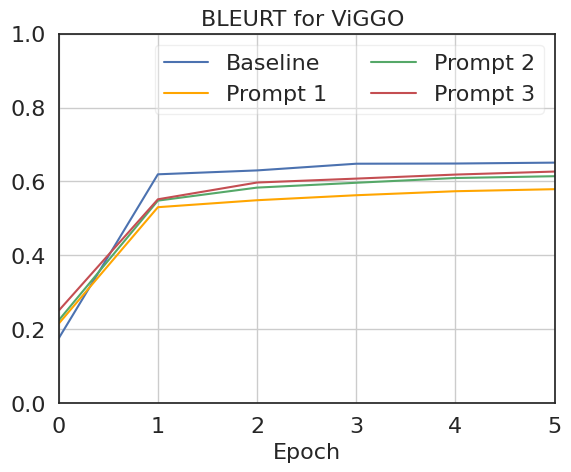}
         \caption{ViGGO (\textbf{BLEURT})}
         \label{fig:bleurt_evolution_viggo}
    \end{subfigure}
    \hfill
    \begin{subfigure}[b]{0.29\textwidth}
         \centering
         \includegraphics[width=\textwidth]{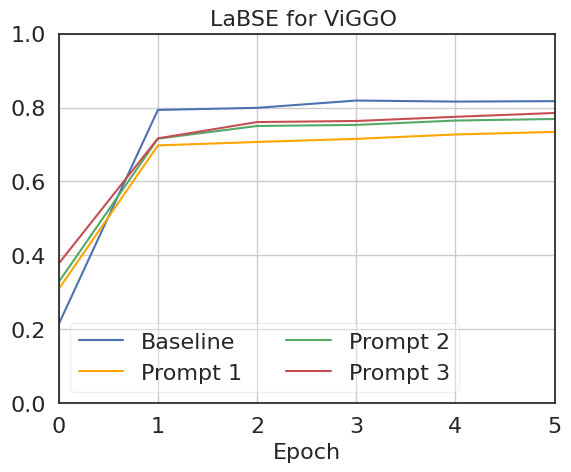}
         \caption{ViGGO (\textbf{LaBSE})}
         \label{fig:labse_evolution_viggo}
    \end{subfigure}
    }
    \makebox[\linewidth][c]{
     \centering
     \begin{subfigure}[b]{0.29\textwidth}
         \centering
         \includegraphics[width=\textwidth]{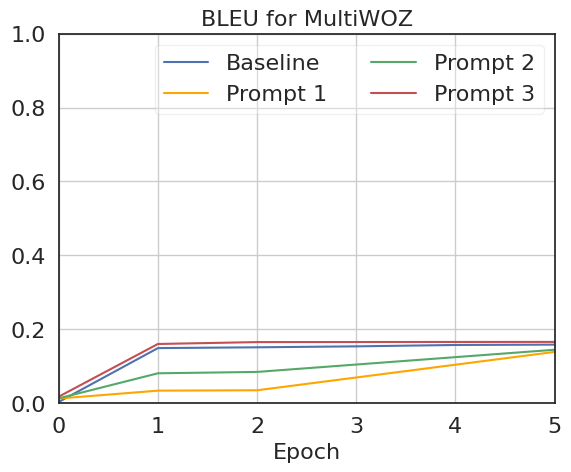}
         \caption{MultiWOZ (\textbf{BLEU})}
         \label{fig:bleu_evolution_multiwoz}
    \end{subfigure}
    \hfill
    \begin{subfigure}[b]{0.29\textwidth}
         \centering
         \includegraphics[width=\textwidth]{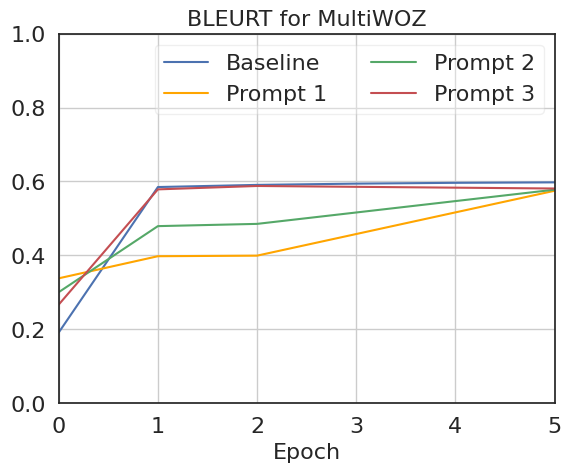}
         \caption{MultiWOZ (\textbf{BLEURT})}
         \label{fig:bleurt_evolution_multiwoz}
    \end{subfigure}
    \hfill
    \begin{subfigure}[b]{0.29\textwidth}
         \centering
         \includegraphics[width=\textwidth]{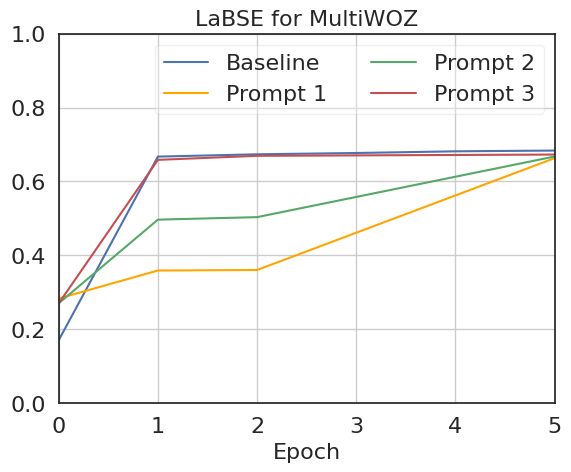}
         \caption{MultiWOZ (\textbf{LaBSE})}
         \label{fig:labse_evolution_multiwoz}
    \end{subfigure}
    }
    \makebox[\linewidth][c]{
    \begin{subfigure}[b]{0.29\textwidth}
         \centering
         \includegraphics[width=\textwidth]{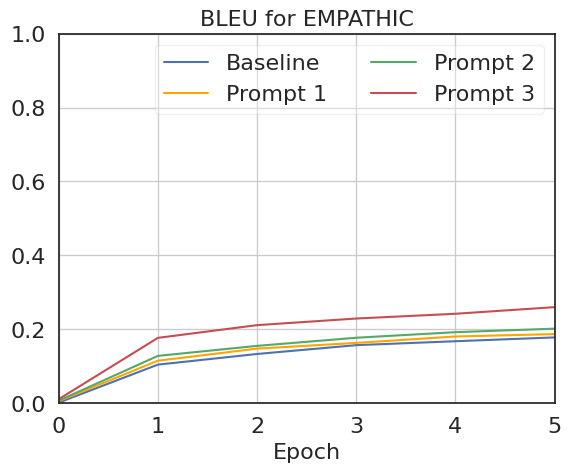}
         \caption{EMPATHIC (\textbf{BLEU})}
         \label{fig:bleu_evolution_empathic}
    \end{subfigure}
    \hfill
    \begin{subfigure}[b]{0.29\textwidth}
         \centering
          \includegraphics[width=\textwidth]{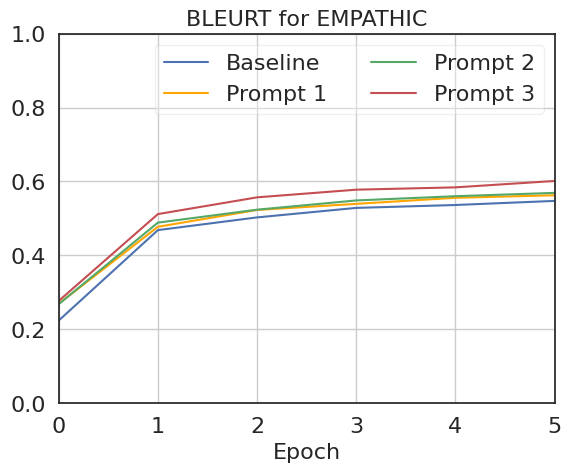}
         \caption{EMPATHIC (\textbf{BLEURT})}
         \label{fig:bleurt_evolution_empathic}
     \end{subfigure}
    \hfill
    \begin{subfigure}[b]{0.29\textwidth}
         \centering
          \includegraphics[width=\textwidth]{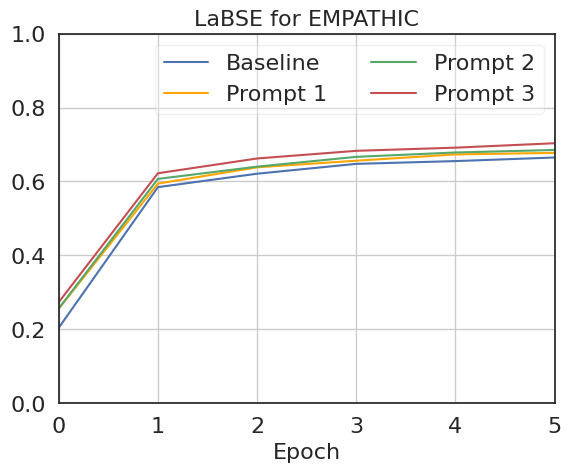}
         \caption{EMPATHIC (\textbf{LaBSE})}
         \label{fig:labse_evolution_empathic}
     \end{subfigure}
    }
    \caption{Evolution of the \textit{average score} of each reference-driven metric during the fine-tuning for each representation in each dataset.}
    \label{fig:evolutions_with_reference}
\end{figure}

\paragraph{\textbf{Singularities of EMPATHIC and MultiWOZ}} The distributions highlight the different behaviour of the model on EMPATHIC. Unlike the Gaussian-shaped curves observed for BLEURT and LaBSE in the other datasets, EMPATHIC displays bimodal-like distributions, particularly pronounced in LaBSE scores. Additionally, it is the only dataset with a considerable portion of BLEU scores spread across the full range, including high-end values. Meanwhile, MultiWOZ exhibits very similar distributions across representations for each metric. We hypothesise that representation influence diminishes when fine-tuning is performed on large-scale datasets, such as MultiWOZ.

\subsubsection*{\textbf{Evolution of the scores}}

To complement the analysis of \textit{average scores}, we examine how model performance evolves throughout the fine-tuning process. This dynamic perspective helps us understand not only the final outcomes but also how different representations support the model's progressive adaptation to the task. In particular, tracking the evolution of scores across epochs allows us to assess the zero-shot effectiveness of the representations and their impact on the model's learning.  

Figure~\ref{fig:evolutions_with_reference} presents learning curves for each combination of metric, dataset, and representation, showing the average score before training (zero-shot) and after each of the five epochs. The figure follows the same format as Figure~\ref{fig:distributions_with_references}, with datasets arranged by row and metrics by column. Within each plot, one learning curve is shown for each representation.

\paragraph{\textbf{Prompt-representations trigger model zero-shot capabilities}} All the learning curves present higher values for the Prompt representations than for the \textit{Baseline} before training (epoch 0), except for BLEU (the first column of plots), whose curves start very close to 0.0 across all datasets and representations. In other words, the zero-shot capabilities of the generative models are activated by the enriched representations. Even in datasets that do not benefit from these representations after fine-tuning, this improvement is observed in the zero-shot setting. Therefore, we show that basic enriched inputs with a single task demonstrator can improve generation quality before training, even in moderately sized models like GPT-2.  

\paragraph{\textbf{Good task adaptability}} The curves generally show most of their improvement during the first epoch, reaching reasonably high scores—except for BLEU. Thus, we confirm that a capable generative model, once fine-tuned, can be easily adapted to multiple tasks and domains.

\paragraph{\textbf{MultiWOZ: slow adaptability with overly general demonstrators}} The curves for \textit{Prompt 1} and \textit{Prompt 2} in MultiWOZ exhibit a more gradual learning process compared to the general behaviour. Although the final scores after five epochs are comparable, the curves do not show most of the improvement in the first epoch. This behaviour may be explained by two factors: a) the limited number of demonstrators for \textit{Prompt 1} and \textit{Prompt 2} (Table~\ref{tab:prompt-size}) compared to the size of MultiWOZ (Table~\ref{tab:corpus-characteristic}), and b) the high output variability of MultiWOZ. These two factors might result in generic task demonstrators that fail to cover the output variability in MultiWOZ. As a result, the demonstrators seem not to be informative enough for all MR, making the adaptation process more challenging for the generative model.

\subsection{Analysis of the results with referenceless metrics}
\label{sec:results_without_references}

Reference-driven metrics need scenarios with predefined references to compare, and they work better as long as these references are representative enough for the given inputs. In addition, their outcomes can be affected by the number of references associated with the MRs, as we will present in Section \ref{sec:distr_effect_results}. In this work, we decided to employ referenceless metrics to overcome this reference necessity and widen the perspective with two completely different metrics: Slot Accuracy and Dialogue Act Accuracy. We discuss the results of these two metrics separately because Slot Accuracy can be evaluated as a generation problem, while DAC is a classification problem.

\subsubsection*{\textbf{Slot Accuracy}}

\begin{table}[htb]
    \centering
    \caption{Slot accuracy \textit{average scores} after fine-tuning for each representation and dataset. Base corresponds to the \textit{Baseline} representation, whereas Pi is the \textit{Prompt i} representation with i=1,2,3. The datasets (Data) are E2E, ViGGO (ViG), MultiWOZ (MWOZ) and EMPATHIC (EMP). No \textit{Prompt 1} for E2E. The best score for each dataset is denoted in bold.}
    \label{tab:slot_acc_results}
    \resizebox{0.35\textwidth}{!}{%
    \begin{tabular}{@{\extracolsep{\fill}}ccccc}
    \toprule
    \multicolumn{1}{c}{
    \textbf{Data}} & \textbf{Base} & \textbf{P1} & \textbf{P2} & \textbf{P3} \\ \midrule
    \multicolumn{1}{c}{\textbf{E2E}} & \textbf{0.79} & - & 0.65 & 0.67  \\ \addlinespace 
    \multicolumn{1}{c}{\textbf{ViG}} & \textbf{0.80} & 0.69 & 0.72 & 0.72 \\ \addlinespace 
    \multicolumn{1}{c}{\textbf{MWOZ}} & \textbf{0.94} & 0.91 & 0.91 & 0.91 \\ \addlinespace
    \multicolumn{1}{c}{\textbf{\begin{tabular}[c]{@{}c@{}}EMP\end{tabular}}} & 0.88 & 0.86 & 0.87 & \textbf{0.90} \\ \bottomrule
    \end{tabular}
    }
\end{table}

\begin{figure}[tbh]
    \makebox[\linewidth][c]{
     \centering
     \begin{subfigure}[b]{0.29\textwidth}
         \centering \includegraphics[width=\textwidth]{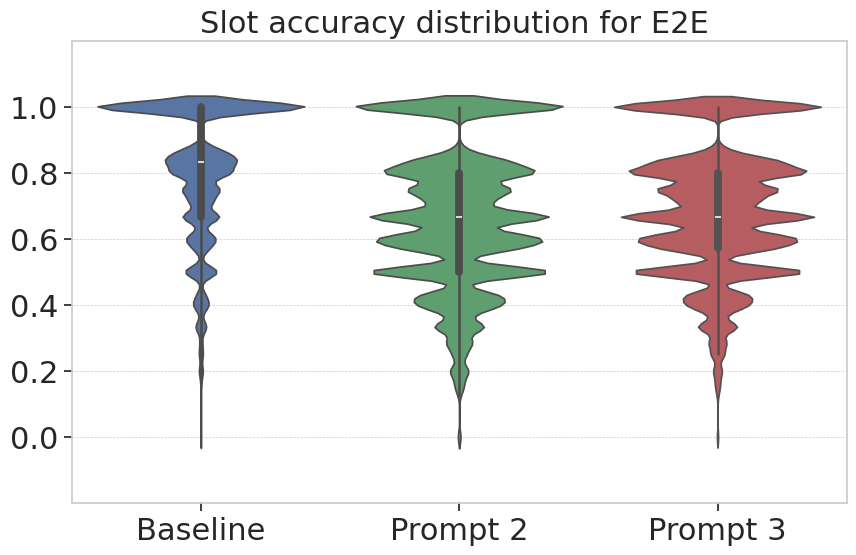}
         \caption{E2E (\textbf{Slot Accuracy})}
         \label{fig:slot_acc_distribution_e2e}
     \end{subfigure}
    \begin{subfigure}[b]{0.29\textwidth}
         \centering
         \includegraphics[width=\textwidth]{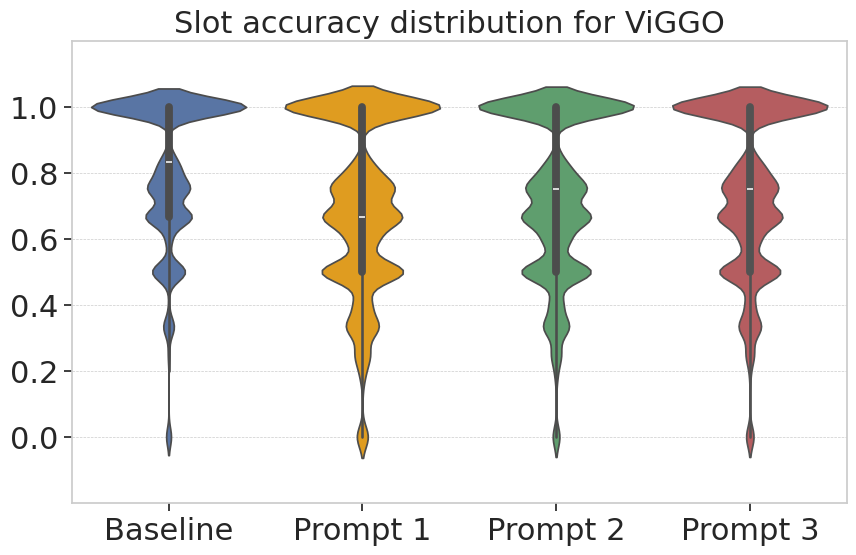}
         \caption{ViGGO (\textbf{Slot Accuracy})}
         \label{fig:slot_acc_distribution_viggo}
    \end{subfigure}
    }
    \makebox[\linewidth][c]{
    \begin{subfigure}[b]{0.29\textwidth}
         \centering
         \includegraphics[width=\textwidth]{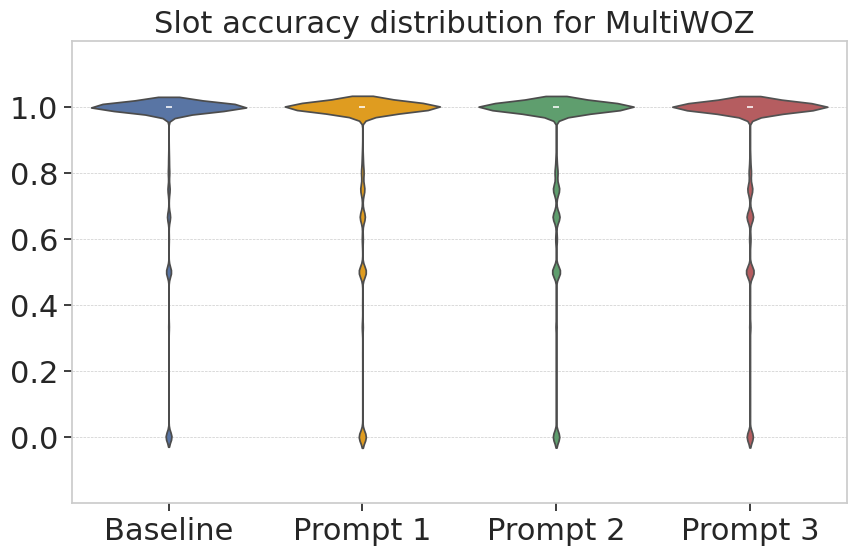}
         \caption{MultiWOZ (\textbf{Slot Accuracy})}
         \label{fig:slot_acc_distribution_multiwoz}
    \end{subfigure}
    \begin{subfigure}[b]{0.29\textwidth}
         \centering
         \includegraphics[width=\textwidth]{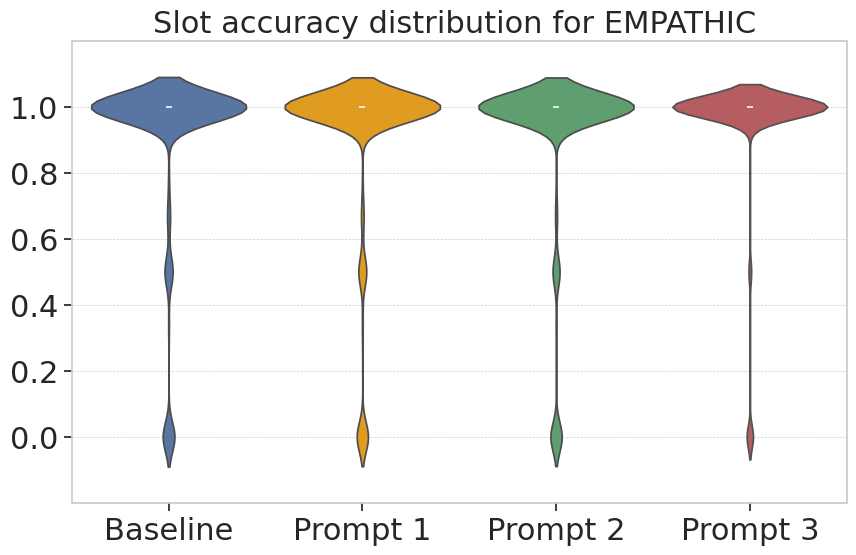}
         \caption{EMPATHIC (\textbf{Slot Accuracy})}
         \label{fig:slot_acc_distribution_empathic}
    \end{subfigure}
    }
    \caption{Distribution of the Slot Accuracy \textit{MR scores} after fine-tuning for each representation in each dataset.}
\label{fig:distributions_slot_accuracy}
\end{figure}

\begin{figure}[tbh]
    \makebox[\linewidth][c]{
     \centering
     \begin{subfigure}[b]{0.29\textwidth}
         \centering
        \includegraphics[width=\textwidth]{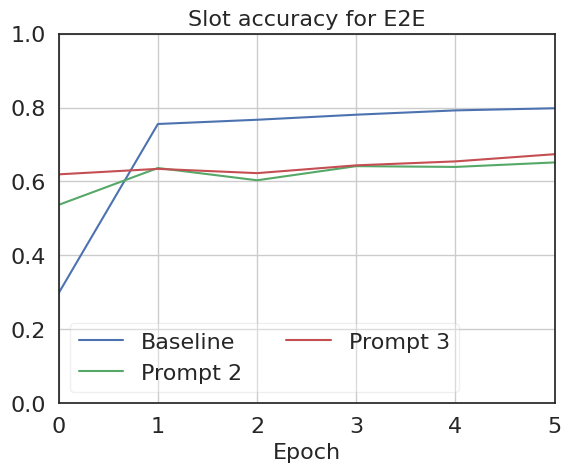}
        \caption{E2E \textbf{(Slot Accuracy)}}
        \label{fig:slot_acc_evolution_e2e} 
    \end{subfigure}
    \begin{subfigure}[b]{0.29\textwidth}
         \centering
         \includegraphics[width=\textwidth]{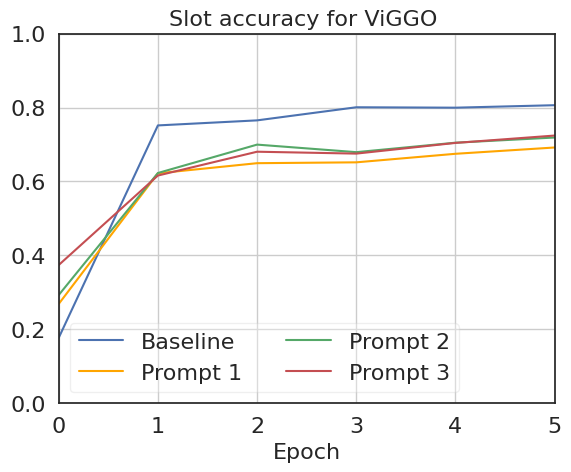}
         \caption{ViGGO \textbf{(Slot Accuracy)}}
         \label{fig:slot_acc_evolution_viggo}
    \end{subfigure} 
    }
    \makebox[\linewidth][c]{
    \begin{subfigure}[b]{0.29\textwidth}
         \centering
         \includegraphics[width=\textwidth]{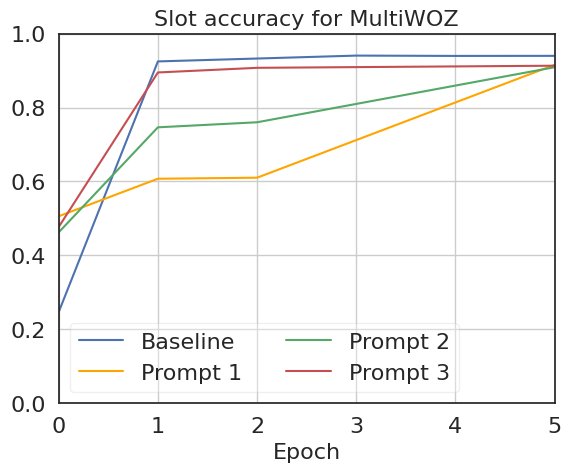}
         \caption{MultiWOZ \textbf{(Slot Accuracy)}}
         \label{fig:slot_acc_evolution_multiwoz}
    \end{subfigure}
    \begin{subfigure}[b]{0.29\textwidth}
         \centering
          \includegraphics[width=\textwidth]{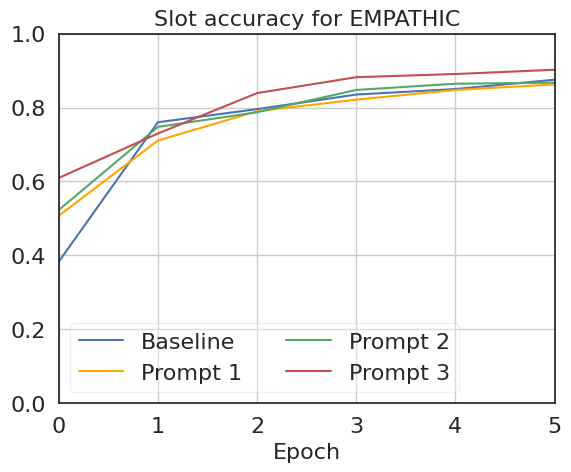}
         \caption{EMPATHIC \textbf{(Slot Accuracy)}}
         \label{fig:slot_acc_evolution_empathic}
    \end{subfigure}
    }
\caption{Evolution of the \textit{average scores} of Slot Accuracy during the fine-tuning for each representation in each dataset.}
\label{fig:evolutions_slot_accuracy}
\end{figure}

The Slot Accuracy measures how accurately the generated sentence expresses the slot values specified in the input. To analyse the scores for this metric, first Table~\ref{tab:slot_acc_results} presents the \textit{average scores} after fine-tuning, then Figure~\ref{fig:distributions_slot_accuracy} shows the distribution of the \textit{MR scores}, and finally Figure~\ref{fig:evolutions_slot_accuracy} depicts the evolution of the \textit{average scores} over training. Note that, unlike reference-driven metrics, a \textit{generation score} is computed by comparing such generation directly with the source MRs, rather than with multiple reference sentences. After that, the process to calculate the \textit{MR scores} and the \textit{average scores} is the same as for the reference-driven metrics.

\paragraph{\textbf{High semantic coherence}} Table \ref{tab:slot_acc_results} presents excellent results mainly for EMPATHIC and MultiWOZ. These two datasets obtain around 90\% of accuracy for all the MRs. E2E and ViGGO also obtain good results, mainly with the Baseline representation. These results mean that the fine-tuned models are semantically coherent with the input.

\paragraph{\textbf{Two distribution behaviours}} The distributions (Figure \ref{fig:distributions_slot_accuracy}) depict two kinds of behaviours, which are a direct consequence of the distribution of the number of attributes in each dataset, as we explain in Section \ref{sec:distr_effect_results}. MultiWOZ and EMPATHIC's distributions are concentrated around the top values due to the low number of attributes in their inputs. Conversely, the higher variance in the number of attributes for E2E and ViGGO results in multiple peaks in their distribution of scores.

\paragraph{\textbf{Slot Accuracy confirms the model behaviour}} Slot Accuracy confirms the model behaviour with the different representations and datasets. First, we observe the benefit of the Prompt representations for datasets like EMPATHIC (Table \ref{tab:slot_acc_results}). Next, the performance improvement in zero-shot settings across all the datasets with the enriched inputs is even more evident for this metric (Figure \ref{fig:evolutions_slot_accuracy}). Finally, this figure also depicts the fast adaptability of all the models to the new tasks, except for the representations with very generic demonstrators for MultiWOZ.

\subsubsection*{\textbf{Dialogue Act accuracy}}

\begin{figure}[tbh]
    \makebox[\linewidth][c]{
     \centering
    \hfill
    \begin{subfigure}[b]{0.29\textwidth}
         \centering
         \includegraphics[width=\textwidth]{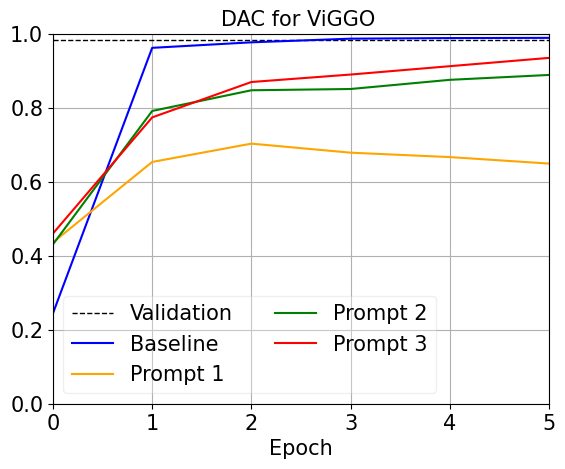}
         \caption{ViGGO \textbf{(DAC)}}
         \label{fig:da_class_evolution_weighted_viggo}
    \end{subfigure}
    \hfill
    \begin{subfigure}[b]{0.29\textwidth}
         \centering
         \includegraphics[width=\textwidth]{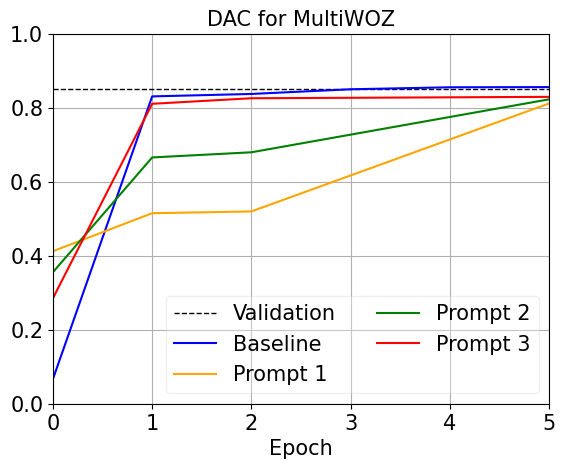}
         \caption{MultiWOZ \textbf{(DAC)}}
    \end{subfigure}
    \hfill
    \begin{subfigure}[b]{0.29\textwidth}
         \centering
          \includegraphics[width=\textwidth]{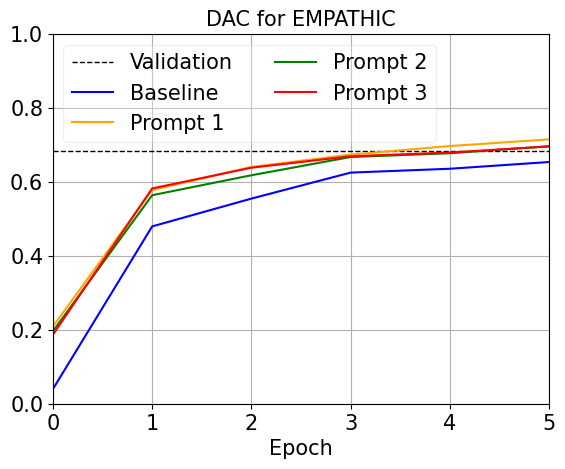}
         \caption{EMPATHIC \textbf{(DAC)}}
         \label{fig:da_class_evolution_weighted_empathic}
     \end{subfigure}
    }
    \caption{Evolution of the Dialogue Act Accuracy \textit{average scores} during the fine-tuning for each representation in each dataset. No E2E as it only contains the inform DA.}
    \label{fig:da_class_evolution}
\end{figure}

The DAC scores are the results of the classification of the generated sentences in terms of DAs. To discuss them, Table~\ref{tab:dac_results} presents the \textit{average scores} after fine-tuning, while Figure~\ref{fig:da_class_evolution} depicts the evolution of those scores during the fine-tuning process. Note that E2E is excluded from this analysis, as it contains only one DA, i.e., only one possible classification output. The analysis also does not present the distribution of \textit{MR scores} because 0 (incorrect prediction DA) and 1 (correct) are the only possible outcomes of the classifiers. In addition, the table and figure show the results obtained using the validation partition as a reference for a good accuracy value in each dataset. The sentences evaluated in this validation set are from the corpus, whereas the rest of the evaluation is over sentences generated by the model.

\begin{table}[tbh]
\centering
\caption{Dialogue Act Accuracy \textit{average scores} after fine-tuning for each representation in each dataset. Val and Base correspond to the Validation dataset and \textit{Baseline} representation, respectively, whereas Pi are the \textit{Prompt i} representations with i=1,2,3. The datasets (Data) are ViGGO (ViG), MultiWOZ (MWOZ) and EMPATHIC (EMP). The best score for each dataset is denoted in bold, the Validation result is excluded from this comparison.}
\resizebox{0.35\textwidth}{!}{%
\begin{tabular}{@{\extracolsep{\fill}}cccccc}
\toprule
\multicolumn{1}{c}{
\textbf{Data}} & \textbf{Val} & \textbf{Base} & \textbf{P1} & \textbf{P2} & \textbf{P3} \\ \midrule
\multicolumn{1}{c}{\textbf{ViG}} & 0.98 & \textbf{0.99} &  0.65 & 0.89 &  0.94 \\ \addlinespace 
\multicolumn{1}{c}{\textbf{MWOZ}} & 0.85 & \textbf{0.86} & 0.52 & 0.68 & 0.83 \\ \addlinespace
\multicolumn{1}{c}{\textbf{\begin{tabular}[c]{@{}c@{}}EMP\end{tabular}}} & 0.68 & 0.65 & \textbf{0.72} & 0.70 & 0.70 \\ \bottomrule
\end{tabular}%
}
\label{tab:dac_results}
\end{table}

\paragraph{\textbf{High coherence with communicative intentions}} 
The results in Table \ref{tab:dac_results} illustrate the excellent classification capabilities of the selected models and the coherence of the generated sentences with the communicative intentions given by the DAs. First, the Validation column proves the exceptional performance of the DistilBERT models as classifiers. 98\% with 9 DAs for ViGGO, 85\%  with 52 DAs in a multilabel classification problem for MultiWOZ, and 68\% with 78 DAs for EMPATHIC are excellent results, considering the number of DAs for each dataset. Secondly, the results obtained with the generations for the different representations (last four columns) prove that these models can generate coherent sentences regarding communicative intention. Some results have even exceeded those obtained with the validation set.

\paragraph{\textbf{Same model behaviour across all the metrics}} Figure \ref{fig:da_class_evolution} confirms some features observed for the model behaviour in the interaction with the datasets and representations, such as the zero-shot capabilities with the Prompt representations, the fast (or slow in some specific cases) model adaption to the task or the prompt influence in the generation for each dataset. This is very relevant because the metrics evaluate different aspects of the generated sentences.\\

\subsection{Impact of the corpus characteristic on the results}
\label{sec:distr_effect_results}

The datasets used in this work are highly diverse, as described in Section~\ref{sec:corpora}. In this section, we analyse whether this diversity in corpus characteristics significantly affects the results. Specifically, we focus on two features shaped by the design and acquisition protocols of the datasets: the number of references and the number of attributes per MR. The number of references determines how many reference sentences are available for comparison in reference-driven metrics, potentially influencing these scores. In turn, the number of attributes is considered a measure of the complexity of the input. We examine the impact of these two corpus features on both lexical (BLEU) and semantic (LaBSE) metrics computed over the fine-tuned models. The section first describes the distribution of these features across datasets, and then analyses their impact on the results separately.

\subsubsection*{\textbf{Relevant corpus characteristics for generation}}

\begin{figure}[h]
\makebox[\linewidth][c]{
    \centering
    \hfill
    \begin{subfigure}[b]{0.46\textwidth}
        \centering\includegraphics[width=\textwidth]{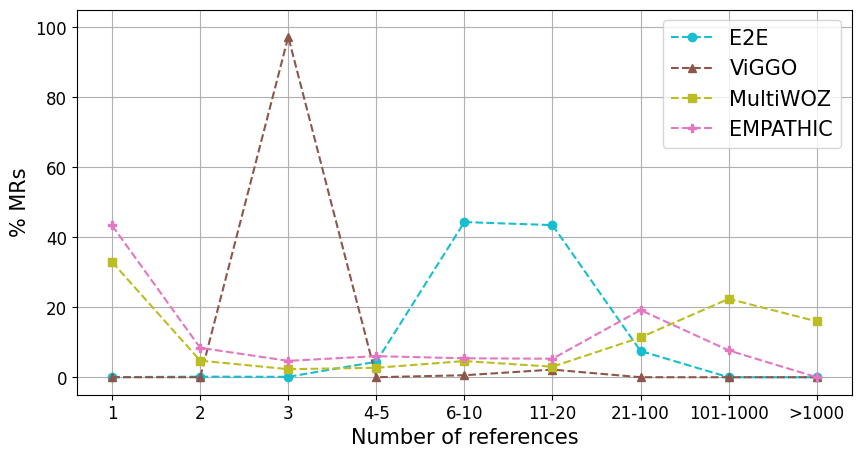}
        \caption{Number of references}
        \label{fig:distr_refs_per_mr}
    \end{subfigure}
    \hfill
    \begin{subfigure}[b]{0.54\textwidth}
        \centering    
        \includegraphics[width=\textwidth]{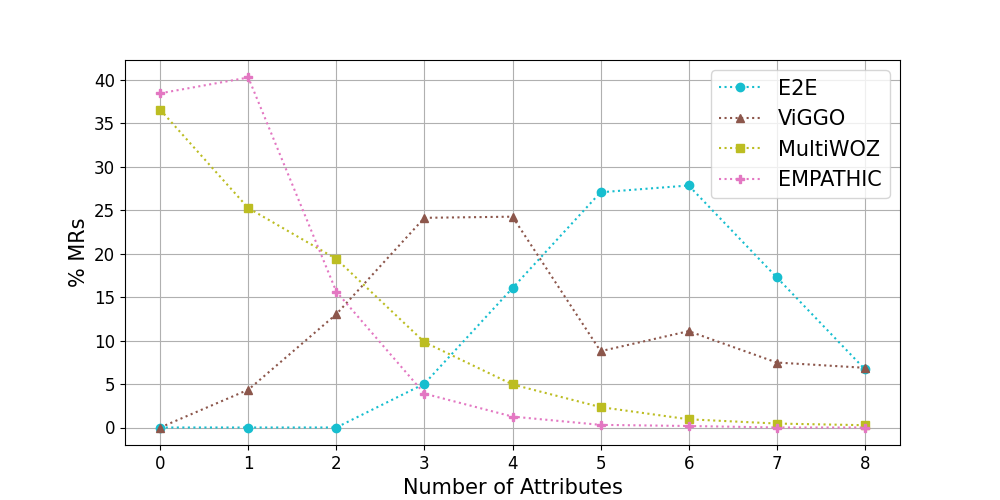}
        \caption{Number of attributes}
        \label{fig:distr_attr}
    \end{subfigure}
    }
    \caption{MR distributions in terms of the number of references/attributes. The number of references/attributes is given on the x-axis with specific numbers of references/attributes or intervals. The distributions are in percentages.} 
    \label{fig:distr_charac}
\end{figure}

For the study of the influence of these two features, we first examine the distribution of MRs in each dataset by number of attributes and references. (Figure \ref{fig:distr_charac}). The plots of this figure show the percentage of MRs that contain the number of references or attributes indicated on the x-axis of each plot. In the case of the number of attributes (Figure \ref{fig:distr_attr}), the numbers go from 0 to 8, whereas the x-axis of Figure \ref{fig:distr_refs_per_mr} presents both specific numbers of references and intervals, such as from 6 to 10 (6-10) or more than 1000 (>1000).

Figure \ref{fig:distr_refs_per_mr} shows that all the MRs of ViGGO have 3 references\footnote{There are some exceptions because some MRs were included several times to increase their training weight. But in practice, all the MRs are linked to 3 references.}, while E2E, acquired with the same protocol, has been designed to have MRs that principally contain between 6 and 20 references. By contrast, half of the MultiWOZ and EMPATHIC MRs have only one reference, but these datasets also have a significant number of MRs with more than 100 references or even more than 1000 in the case of MultiWOZ. These distributions are influenced by the acquisition process and the design of the datasets. Datasets like E2E and ViGGO, which write and select sentences for predefined MR, have more control over this feature than those that need to annotate previously created conversations, such as MultiWOZ and EMPATHIC.

The design of the datasets also influences the distribution of the number of attributes per MR (Figure \ref{fig:distr_attr}), with similarities between the same datasets. E2E and ViGGO were created to train NLG engines that can manage inputs with multiple attributes. Consequently, these two datasets always present at least one attribute and usually more than three, with a predominance of MRs between 3 and 4 attributes in ViGGO and between 5 and 6 in E2E. By contrast, the attributes for MultiWOZ and EMPATHIC are the entities found in the conversations during the annotation process. As a result, more than half of the MRs of these two datasets present no attributes or only one, whereas the proportion for MRs with more than four attributes is almost irrelevant.

\subsubsection*{\textbf{Influence of the number of references}}

\begin{figure}[h]
    \makebox[\linewidth][c]{
     \centering
    \hfill
    \begin{subfigure}[b]{0.5\textwidth}
         \centering
         \includegraphics[width=\textwidth]{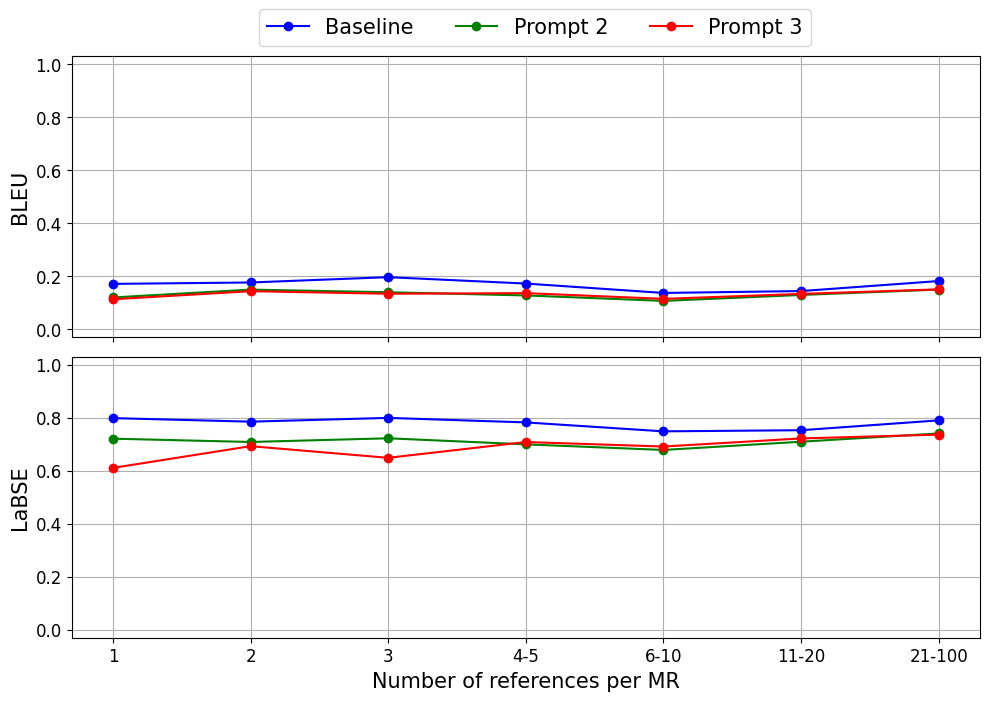}
         \caption{E2E}
        \label{fig:results_effect_MR_e2e}
    \end{subfigure}
    \hfill
    \begin{subfigure}[b]{0.5\textwidth}
         \centering
         \includegraphics[width=\textwidth]{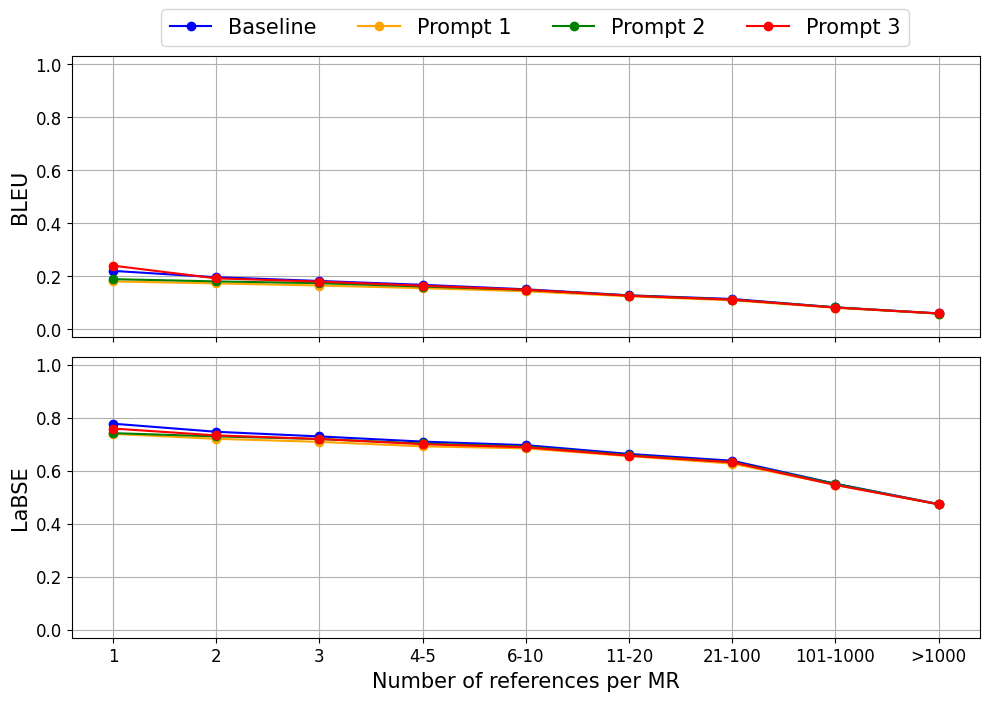}
         \caption{MultiWOZ}
          \label{fig:results_effect_MR_multiwoz}
    \end{subfigure}
    }
    \makebox[\linewidth][c]{
    \begin{subfigure}[b]{0.5\textwidth}
         \centering
         \includegraphics[width=\textwidth]{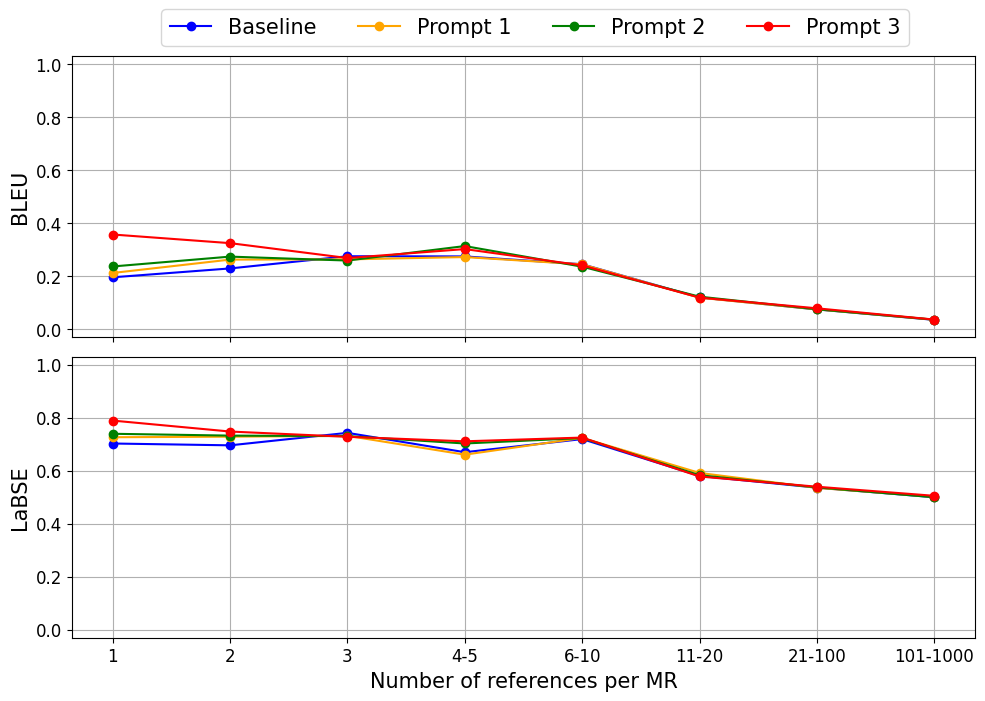}
         \caption{EMPATHIC}
         \label{fig:results_effect_MR_empathic}
    \end{subfigure}
    }
    \caption{Results for \textbf{BLEU} (top) and \textbf{LaBSE} (bottom) scores in terms of the \textbf{number of references} per MR with the fine-tuned models. The x-axis establishes specific numbers of references or intervals of numbers of references.} 
    \label{fig:results_effect_MR}
\end{figure}

Figure~\ref{fig:results_effect_MR} shows the distribution of scores in terms of the number of references across datasets. This figure includes three subplots—one per dataset—each showing the BLEU (top) and LaBSE (bottom) scores obtained with the fine-tuned models. Note that the ViGGO dataset is not included in this analysis because all its MRs have three references. Each point in the plots represents the average score over all MRs that share the same number (or interval) of references as indicated on the x-axis. For example, a point over the label 6--10 corresponds to the average score computed across all MRs that have between 6 and 10 references.

\paragraph{\textbf{No meaningful differences across metrics and representations}} The results for BLEU and LaBSE in each dataset only differ in the absolute values; the overall trends are the same. In the comparison between representations, the differences observed in the results mainly appear in the MRs with one reference. So, we can assume that the impact of the representations on the model is not affected by the number of references.

\paragraph{\textbf{Higher impact in datasets with lexically and semantically different references}} The comparison across datasets reveals a distinct effect of the number of references on the scores. The scores for E2E remain very similar regardless of the number of references. However, MultiWOZ and EMPATHIC exhibit a clear decrease in their scores as the number of references increases. We hypothesise that this is related to the output variability of the datasets: high for MultiWOZ and EMPATHIC, and low for E2E, as shown in Section~\ref{sec:corpora}. The high variability in MultiWOZ and EMPATHIC implies that the references often differ in both lexicon and semantics. For example, the MR \texttt{SYSTEM\_general\_bye (  )} from MultiWOZ presents 5394 references that contain very lexically and semantically different sentences, such as \textit{"I hope you enjoy your stay. Contact us anytime. Goodbye."}, \textit{"Thanks for using our system today!"} and \textit{"You have a good day too."}.  This reduces the likelihood that a generated sentence will obtain a high \textit{generation score} against all the references, producing a lower average. In contrast, in E2E, even the MRs with the highest number of references contain very similar sentences. In fact, the MR with the highest number of references, 46, for this dataset is \texttt{inform ( name = Loch Fyne ; eatType = restaurant ; food = English ; familyFriendly = yes )} with references like \textit {"A child friendly restaurant that has English food is Loch Fyne."}, \textit{"A good kid-friendly, English restaurant is Loch Fyne."} and \textit{"Loch Fyne serves English food. It's a family friendly restaurant."}. Consequently, the number of reference have little impact on this dataset.

\subsubsection*{\textbf{Influence of the number of attributes}}

\begin{figure}[tbh]
    \makebox[\linewidth][c]{
     \centering
    \hfill
    \begin{subfigure}[b]{0.5\textwidth}
         \centering
         \includegraphics[width=\textwidth]{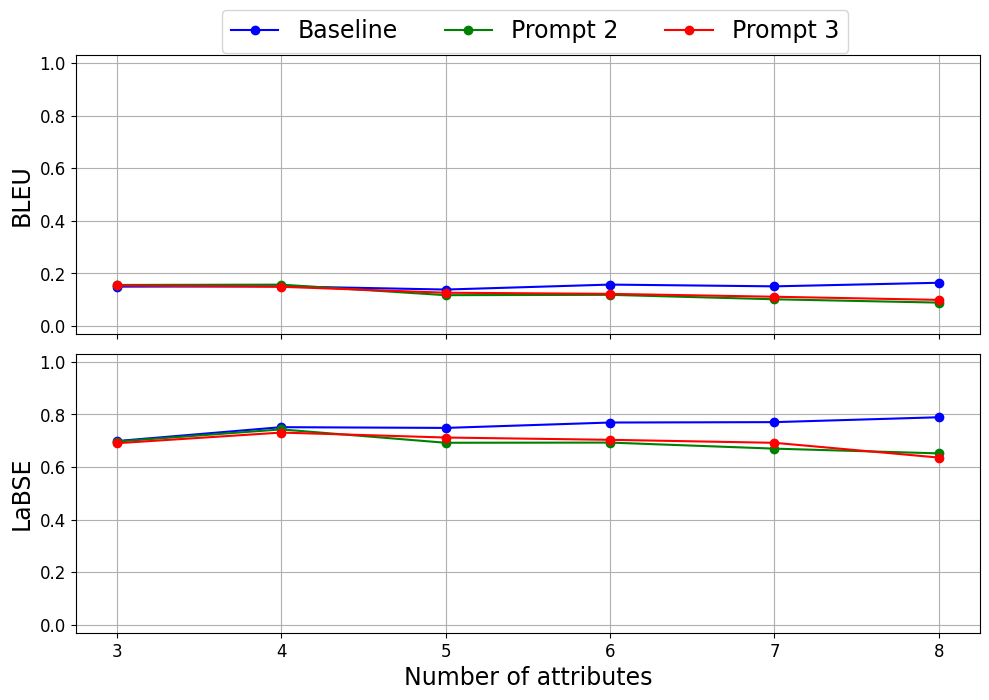}
         \caption{E2E}
         \label{fig:results_effect_attr_e2e}
    \end{subfigure}
    \hfill
    \begin{subfigure}[b]{0.5\textwidth}
         \centering
         \includegraphics[width=\textwidth]{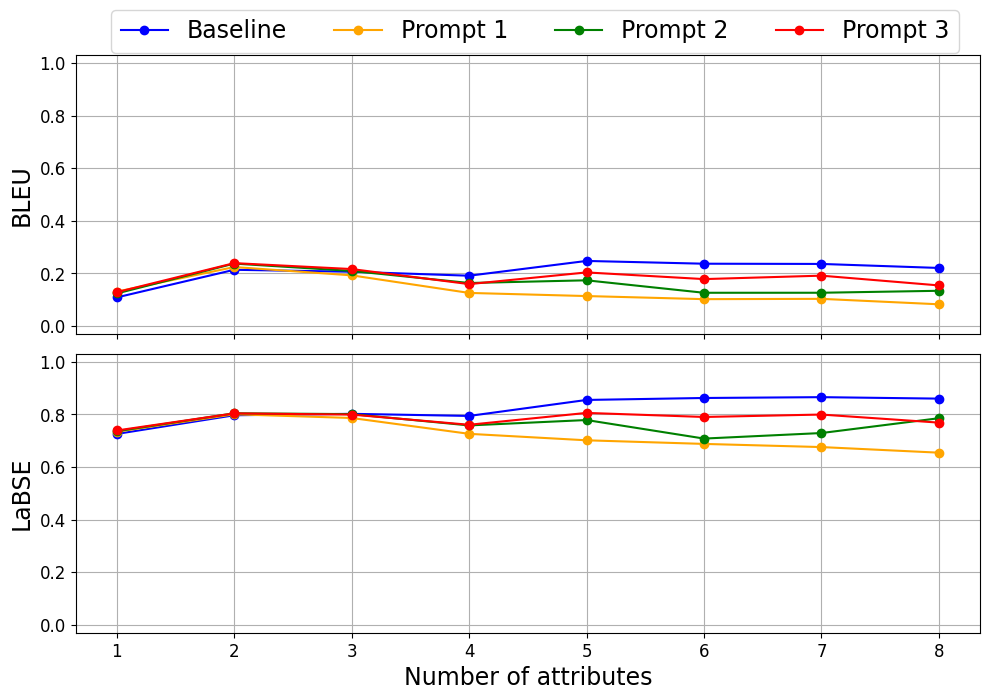}
         \caption{ViGGO}
         \label{fig:results_effect_attr_viggo}
    \end{subfigure}
    }
    \makebox[\linewidth][c]{
     \centering
    \hfill
    \begin{subfigure}[b]{0.5\textwidth}
         \centering
         \includegraphics[width=\textwidth]{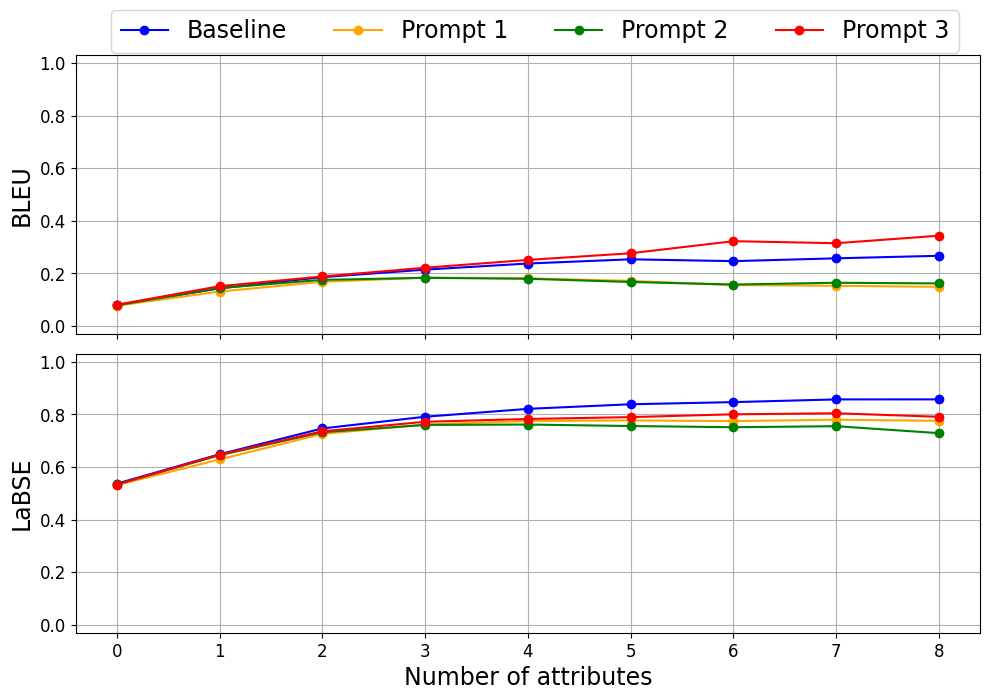}
         \caption{MultiWOZ}
         \label{fig:results_effect_attr_multiwoz}
    \end{subfigure}
    \hfill
    \begin{subfigure}[b]{0.5\textwidth}
         \centering
         \includegraphics[width=\textwidth]{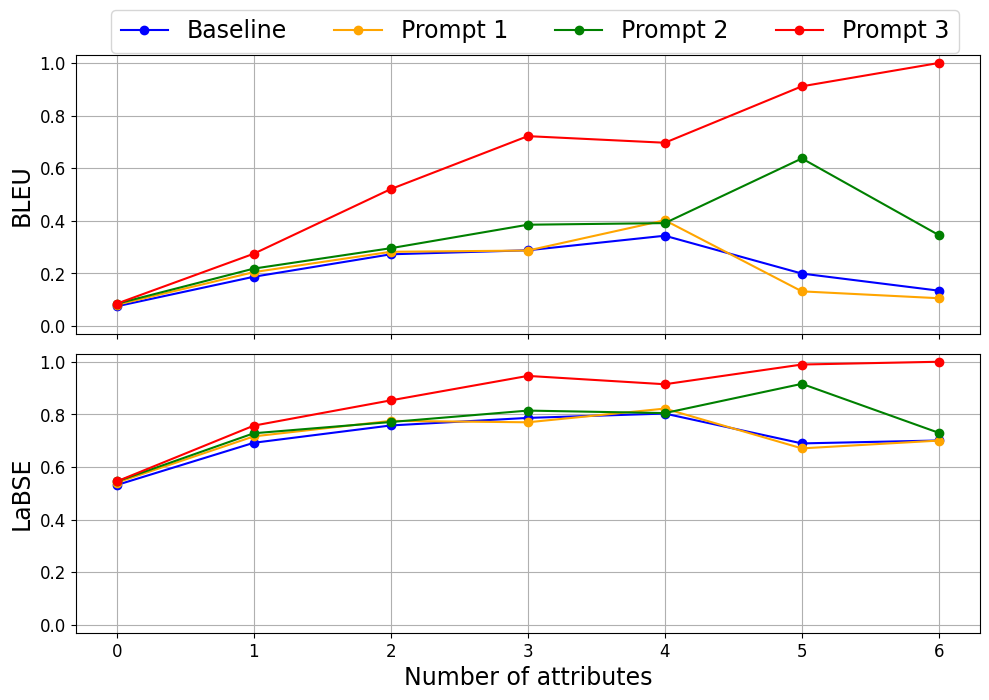}
         \caption{EMPATHIC}
         \label{fig:results_effect_attr_empathic}
    \end{subfigure}
    }
    \caption{Results for \textbf{BLEU} (top) and \textbf{LaBSE} (bottom) scores in terms of the \textbf{number of attributes} per MR with the fine-tuned models.} 
    \label{fig:results_effect_attr}
\end{figure}

Figure \ref{fig:results_effect_attr} shows the effect of the distribution of the number of attributes over the four datasets. The figure is the equivalent of Figure \ref{fig:results_effect_MR} but for the number of attributes. In this case, the ViGGO dataset is included in the study as its MRs do present variability in the number of attributes (Figure \ref{fig:distr_attr}).

\paragraph{\textbf{Increasing impact of the meaning representation with the number of attributes}} Figure~\ref{fig:results_effect_attr} shows consistent trends for BLEU and LaBSE: the higher the number of attributes, the bigger the differences among the representations. Therefore, the structure of the MRs becomes more relevant when the input includes a high number of attributes.

\paragraph{\textbf{Issue with artificial perfect scores}} Figure~\ref{fig:results_effect_attr_empathic} shows scores of 1.0 for EMPATHIC MRs with 6 attributes in both metrics. Although these values may look highly positive, they actually reflect an undesired behaviour. Due to the small size of EMPATHIC and the very limited number of MRs with 6 attributes (Figure~\ref{fig:distr_attr}), we found that these perfect scores come from MRs with a single reference that also appears in the task demonstrator of the \textit{Prompt 3} representation. In these cases, the model simply copies the demonstrator sentence, which matches the only reference used for evaluation, leading to artificially perfect scores. For example, the MR \texttt{RQ\_curr\_sit\_freq (action = think ; action = eat ; number = five ; food = fruits ; food = vegetables ; freq = per day )} is the only MR with this DA and these attributes. Therefore, its reference and the \textit{Prompt 3} demonstrator are identical: \textit{"And do you think you eat five fruits and vegetables per day?"}. Moreover, the 5 generated samples are exactly the same sentence. Note that the effect on the \textit{average scores} is negligible, since this happens in less than 1\% of the samples. Still, these observations suggest that datasets that include complex MRs with few references may benefit from preprocessing steps such as paraphrasing or other data augmentation techniques to reduce this behaviour.

\paragraph{\textbf{Direct relation with Slot Accuracy results}}  The number of attributes has a direct impact on the Slot Accuracy results. If we go back to the distributions of results for the Slot Accuracy (Figure~\ref{fig:distributions_slot_accuracy}) and the distributions of the number of attributes per MR Figure~\ref{fig:distr_attr}, we can observe one behaviour for MultiWOZ and EMPATHIC and another for E2E and ViGGO. The first two datasets tend to have fewer than three attributes in their MRs, so the percentages of Slot Accuracy are usually 0, 50, or 100\%. Consequently, the distributions of results for these two datasets are mostly observed around these values. Conversely, E2E and ViGGO usually present inputs with between three and eight attributes. So, the possible values for the Slot Accuracy percentage are varied, and the distribution of Slot Accuracy scores presents a significant proportion of MRs across the entire range. In addition, the size of the evaluation set for Slot Accuracy is drastically reduced for MultiWOZ and EMPATHIC, since inputs with no attributes are not included in their evaluation, which accounts for a substantial portion of these two datasets (around 40\%).

\subsection{Qualitative analysis of the generations}
\label{sec:generation_analysis}

To complement the quantitative evaluation with a more in-depth understanding of the model outputs, we provide a qualitative analysis of the generated sentences. For this human evaluation, a manual selection of examples from ViGGO and EMPATHIC was carried out. We have selected these two datasets because they present notable differences in both design and performance. In this assessment, we first analyse why BLEU scores are much lower than those obtained with the other two reference-driven metrics: BLEURT and LaBSE. Then, we carry out a similar and novel comparison between these two semantic reference-based metrics. Finally, we study the opposing effect of the \textit{Prompt 3} representation in ViGGO and EMPATHIC, which was selected as the best among the enriched representations.

For all the comparisons, a set of samples was selected from the outputs of the fine-tuned models. The metric comparisons include generated sentences that show the largest discrepancies between the metrics analysed. Meanwhile, the comparison between the Baseline and Prompt 3 focuses on those MRs that, on average, show the largest differences in reference-based metric scores between the two representations, i.e., the MRs with the highest average absolute difference in MR scores across the three reference-based metrics.

From this qualitative inspection, we confirm that metrics focusing on lexical aspects do not reliably reflect generation quality. In contrast, the human-trained BLEURT metric appears to capture omissions and semantic nuances that go undetected by LaBSE. Finally, the comparison of representations sheds light on the different impact of the enriched representations on ViGGO and EMPATHIC generations. Specifically, adding a demonstrator to some long ViGGO inputs can induce a forgetting effect in the models, while in EMPATHIC, the demonstrator help fine-tuned models to learn sentence structures principally.

\subsubsection*{\textbf{Analysis of metric scores}}

Tables \ref{tab:low_bleu_high_sem_exs} and \ref{tab:bleurt_vs_labse_exs} present representative generation examples that show the different behaviour of the metrics. Each table contains ViGGO and EMPATHIC examples. These examples consist of the MR, the references (with the number of them) linked to this MR, the generated sentence, the dataset and the generation scores for all the metrics, except DAC, whose individual scores are irrelevant. In the tables, the headers of the input elements are orange, while those of the output are purple. For the compared metrics, we highlighted the lowest values in red and the highest ones in green. 

\begin{table}[h]
\centering
\caption{Examples to compare \textbf{BLEU} with \textbf{BLEURT} and \textbf{LaBSE}. Slot Accuracy (Slot acc) is also included. Each example contains, from top to bottom: MR, references, a generation, as well as the dataset and scores for all the metrics in the last row. Relevant elements are highlighted in bold and different colours as follows: orange for the headers related to the inputs, purple for those related to the outputs, blue for the datasets, red for BLEU scores, and green for BLEURT and LaBSE scores.} 
\label{tab:low_bleu_high_sem_exs}
\resizebox{\textwidth}{!}{%
\begin{tabular}{llllllllll}
\toprule
\textcolor{Orange}{\textbf{MR}} & \multicolumn{9}{l}{inform ( name = Crysis ; release\_year = 2007 ; esrb = M (for Mature) ; rating = good ; genres = action-adventure, shooter ; has\_multiplayer = yes )} \\ \addlinespace
\textcolor{Orange}{\textbf{References}} & \multicolumn{9}{l}{\begin{tabular}[t]{@{}l@{}}\textbf{3 references:} - A solid M-rated action-adventure shooter, Crysis, came out in 2007. The game offers multiplayer.\\ - An action-adventure shooter game for mature audiences is the 2007 game, Crysis. It was a fun game offering also a multiplayer mode.\\ - Crysis is an action-adventure shooter that was released in 2007, rated M (for Mature). It is considered good quality, and has multiplayer mode.\end{tabular}} \\ \addlinespace
\textcolor{Purple}{\textbf{Generation}} & \multicolumn{9}{l}{Crysis is an action-adventure shooter released in 2007. It has received good reviews.} \\ \addlinespace
\textbf{Dataset} & \textcolor{Blue}{\textbf{ViGOO}} & \textcolor{Purple}{\textbf{BLEU}} & \textcolor{Red}{\textbf{0.06}} & \textcolor{Purple}{\textbf{BLEURT}} & \textcolor{Green}{\textbf{0.66}} & \textcolor{Purple}{\textbf{LaBSE}} & \textcolor{Green}{\textbf{0.81}} & \textcolor{Purple}{\textbf{Slot acc}} & 0.80 \\ \midrule
\textcolor{Orange}{\textbf{MR}} & \multicolumn{9}{l}{Gen\_Open\_question ( action = cook ; food = fish )} \\ \addlinespace
\textcolor{Orange}{\textbf{References}} & \multicolumn{9}{l}{\begin{tabular}[t]{@{}l@{}} \textbf{1 reference:} What's your favourite fish to cook? \end{tabular}} \\ \addlinespace
\textcolor{Purple}{\textbf{Generation}} & \multicolumn{9}{l}{Do you like to cook fish?} \\ \addlinespace
\textbf{Dataset} & \textcolor{Blue}{\textbf{EMPATHIC}} & \textcolor{Purple}{\textbf{BLEU}} & \textcolor{Red}{\textbf{0.07}} & \textcolor{Purple}{\textbf{BLEURT}} & \textcolor{Green}{\textbf{0.71}} & \textcolor{Purple}{\textbf{LaBSE}} & \textcolor{Green}{\textbf{0.83}} & \textcolor{Purple}{\textbf{Slot acc}} & 1.00 \\ \midrule
\textcolor{Orange}{\textbf{MR}} & \multicolumn{9}{l}{Gen\_Hello ( )} \\ \addlinespace
\textcolor{Orange}{\textbf{References}} & \multicolumn{9}{l}{\begin{tabular}[t]{@{}l@{}} \textbf{25 references:} \qquad - Good morning \qquad - Hello, how are you?	\qquad - Hey, what's up? good morning. \qquad - Hello. \qquad - (21 references more) \end{tabular}} \\ \addlinespace
\textcolor{Purple}{\textbf{Generation}} & \multicolumn{9}{l}{Hello!} \\ \addlinespace
\textbf{Dataset} & \textcolor{Blue}{\textbf{EMPATHIC}} & \textcolor{Purple}{\textbf{BLEU}} & \textcolor{Red}{\textbf{0.05}} & \textcolor{Purple}{\textbf{BLEURT}} & \textcolor{Green}{\textbf{0.64}}  & \textcolor{Purple}{\textbf{LaBSE}} & \textcolor{Green}{\textbf{0.59}} & \textcolor{Purple}{\textbf{Slot acc}} & - \\ \bottomrule
\end{tabular}%
}
\end{table}

\begin{table}[h]
\centering
\caption{Examples to compare \textbf{BLEURT} with \textbf{LaBSE}. BLEU and Slot Accuracy (Slot acc) are also included. Each example contains, from top to bottom: MR, references, a generation, as well as the dataset and scores for all the metrics in the last row. Relevant elements are highlighted in bold and different colours as follows: orange for the headers related to the inputs, purple for those related to the outputs, blue for the datasets, red for BLEURT scores, and green for LaBSE scores.}
\label{tab:bleurt_vs_labse_exs}
\resizebox{\textwidth}{!}{%
\begin{tabular}{llllllllll}
\toprule
\textcolor{Orange}{\textbf{MR}} & \multicolumn{9}{l}{inform ( name = Tony Hawk's Pro Skater 3 ; release\_year = 2001 ; esrb = T (for Teen) ; genres = sport ; player\_perspective = third person )} \\ \addlinespace
\textcolor{Orange}{\textbf{References}} & \multicolumn{9}{l}{\begin{tabular}[t]{@{}l@{}} \textbf{3 references:} - Tony Hawk's Pro Skater 3 was released in 2001. It is a third person sports game that is rated T (for Teen).\\ - Tony Hawk's Pro Skater 3 was a third person sport game released in 2001. It was rated T (for Teen).\\ - Another third person perspective sport game is Tony Hawk's Pro Skater 3. It came out in 2001 and is rated T (for Teen).\end{tabular}} \\ \addlinespace
\textcolor{Purple}{\textbf{Generation}} & \multicolumn{9}{l}{The 2001 sport game, Tony Hawk's Pro Skater 3, was a third person game. 
It has a t (for teen) rating and is rated t (for physical).} \\ \addlinespace
\textbf{Dataset} & \textcolor{Blue}{\textbf{ViGGO}} & \textcolor{Purple}{\textbf{BLEU}} & 0.34 & \textcolor{Purple}{\textbf{BLEURT}} & \textcolor{Red}{\textbf{0.61}} & \textcolor{Purple}{\textbf{LaBSE}} & \textcolor{Green}{\textbf{0.92}} & \textcolor{Purple}{\textbf{Slot acc}} & 1.00 \\ \midrule
\textcolor{Orange}{\textbf{MR}} & \multicolumn{9}{l}{Gen\_Agreement (  )} \\ \addlinespace
\textcolor{Orange}{\textbf{References}} & \multicolumn{9}{l}{\begin{tabular}[t]{@{}l@{}} \textbf{91 references:} \qquad - Absolutely \qquad - Of course. \qquad - Yeah \qquad - Right  \qquad - I totally understand. \qquad - (86 references more) \end{tabular}} \\ \addlinespace
\textcolor{Purple}{\textbf{Generation}} & \multicolumn{9}{l}{Sure.} \\ \addlinespace
\textbf{Dataset} & \textcolor{Blue}{\textbf{EMPATHIC}} & \textcolor{Purple}{\textbf{BLEU}} & 0.02 & \textcolor{Purple}{\textbf{BLEURT}} & \textcolor{Red}{\textbf{0.09}} & \textcolor{Purple}{\textbf{LaBSE}} & \textcolor{Green}{\textbf{0.56}} & \textcolor{Purple}{\textbf{Slot acc}} & - \\ \bottomrule
\end{tabular}%
}
\end{table}

\paragraph{\textbf{BLEU is not adequate for evaluation of output quality}} Table~\ref{tab:low_bleu_high_sem_exs} shows three examples (one for ViGGO and two for EMPATHIC) of good generations with very low BLEU (in red in the table) and reasonably high BLEURT and LaBSE scores (in green). All three generations accurately reflect their corresponding MR. The ViGGO example's generation, \textit{"Crysis is an action-adventure shooter released in 2007. It has received good reviews"}, contains two omissions (\texttt{esrb} and \texttt{has\_multiplayer}), but it is a good \texttt{inform} sentence that conveys nearly all the intended semantic content. Regarding the EMPATHIC generations, \textit{"Do you like to cook fish?"} and \textit{"Hello!"} are perfectly acceptable outputs for an open question about cooking fish and a greeting, respectively. Thus, the poor BLEU scores in these examples confirm that it may not be suitable for assessing output quality. In addition, in the first EMPATHIC example, the generated sentence contains the n-gram \textit{"to cook fish?"}, which uses the same words as the reference n-gram \textit{"fish to cook?"}. However, this does not lead to a high score, presumably due to the different word order. Therefore, BLEU appears to be highly sensitive to lexical order.

\paragraph{\textbf{BLEURT detects omissions and semantic nuances}} Table~\ref{tab:bleurt_vs_labse_exs} shows representative examples of the comparison performed between BLEURT and LaBSE. The two examples (one for each dataset) present higher scores for LaBSE. However, the human-trained BLEURT seems to capture more subtle aspects than LaBSE. For instance, the ViGGO example presents a hallucination in its generation, where the slot \texttt{esrb = T (for Teen)} is generated twice, and one of them contains incorrect information: \textit{"It has a t (for teen) rating and is rated t (for physical)"}. This error goes unnoticed by LaBSE, which gives an almost perfect score of 0.92, whereas BLEURT produces a more adequate score of 0.61. In the EMPATHIC example, the generation \textit{"Sure."} can be considered good for expressing agreement and obtains 0.09 from BLEURT and 0.56 from LaBSE. We attribute this difference to the fact that while all references are appropriate to convey agreement, not all express it with the same degree of emphasis. These semantic nuances in the meanings of \textit{"Yeah"}, \textit{"I totally understand"}, or the generated \textit{"Sure"} appear to influence BLEURT scores, which may provide evidence of BLEURT's sensitivity to such distinctions.

\subsubsection*{\textbf{Analysis of the influence of Prompt representation for different datasets}}

\begin{table}[h]
\centering
\caption{Examples of the generations obtained with the \textbf{Baseline} and \textbf{Prompt 3} representations. Each example contains from top to bottom: MR, references, the task demonstrator linked to the MR in the Prompt 3 representation, generations obtained with the Baseline representation, generations obtained with the Prompt 3 representation, and dataset and scores for all the metrics for both representations (Baseline/Prompt 3) in the last row. The highest score for each metric is highlighted in bold.}
\label{tab:base_vs_prompt_exs}
\resizebox{\textwidth}{!}{%
\begin{tabular}{llllllllll}
\toprule
\textcolor{Orange}{\textbf{MR}} & \multicolumn{9}{l}{\begin{tabular}[t]{@{}l@{}} inform ( name = HSS ; genres = adventure, hack-and-slash, indie ; platforms = PlayStation, Xbox, PC ; available\_on\_steam = yes ; \\has\_linux\_release = no ; has\_mac\_release = no ) \end{tabular}} \\ \addlinespace
\textcolor{Orange}{\textbf{References}} & \multicolumn{9}{l}{\begin{tabular}[t]{@{}l@{}} \textbf{3 references:} - HSS is an indie adventure, hack-and-slash game for PlayStation, Xbox, and PC. 
It is available for Steam, however, \\ it is not out for Linux or even Mac. \\ - HSS is an indie hack-and-slash adventure game available on Steam. You can play it on PlayStation, Xbox, and PC, but not on \\ Linux or Mac. \\ - HSS is an indie, adventure, hack-and-slash game that's available on PlayStation, Xbox and PC. You can also get it from Steam,\\ but it does not have Linux or Mac releases.\end{tabular}} \\ \addlinespace
\begin{tabular}[t]{@{}l@{}} \textcolor{Orange}{\textbf{Prompt 3}} \\ \textcolor{Orange}{\textbf{task demonstrator}} \end{tabular} & \multicolumn{9}{l}{\begin{tabular}[t]{@{}l@{}} inform ( name = The Sims ; genres = simulation, strategy ; platforms = PlayStation, Xbox, Nintendo, PC ; available\_on\_steam = no ; \\has\_linux\_release = no ; has\_mac\_release = yes ) + The Sims is a simulation-strategy game that is available for PlayStation, Xbox,\\Nintendo, and PC. It can operate on Mac, but it is not released for Linux. It is not available on Steam. \end{tabular}} \\ \addlinespace
\begin{tabular}[t]{@{}l@{}} \textcolor{Purple}{\textbf{Baseline}} \\ \textcolor{Purple}{\textbf{Generations}} \end{tabular} & \multicolumn{9}{l}{\begin{tabular}[t]{@{}l@{}} \textbf{15 generations:} - HSS is an indie adventure hack-and-slash game. It's available on the following platforms: PlayStation, Xbox, \\ and PC. Steam is also the home of the game, though it isn't available for Linux or Mac. \\  - If you're into indie adventure and hack-and-slash games, try HSS. It's an indie game released for PlayStation, Xbox, PC,\\ and Mac. You can get it on Steam now. \\ - If you like indie hack-and-slash adventures on the PC, try HSS. It's available on Steam, and you can play it on all platforms,\\ such as the PlayStation, Xbox, and PC. There was no Linux or Mac release. \qquad - (12 generations more)
\end{tabular}} \\ \addlinespace
\begin{tabular}[t]{@{}l@{}} \textcolor{Purple}{\textbf{Prompt 3}} \\ \textcolor{Purple}{\textbf{Generations}} \end{tabular} & \multicolumn{9}{l}{\begin{tabular}[t]{@{}l@{}} \textbf{15 generations:} - HSS is an indie adventure-shooter with puzzle solving elements. \\  - HSS is an indie adventure-hack and slash with a Mac twist. \\ - HSS is an indie adventure-hack-and-slash on the PC that has been released on Steam. \qquad - (12 generations more)
\end{tabular}} \\ \addlinespace
\textbf{Dataset} & \textcolor{Blue}{\textbf{ViGGO}} & \textcolor{Purple}{\textbf{BLEU}} & \textbf{0.31} / 0.15 & \textcolor{Purple}{\textbf{BLEURT}} & \textbf{0.70} / 0.56 & \textcolor{Purple}{\textbf{LaBSE}} & \textbf{0.89} / 0.68 & \textcolor{Purple}{\textbf{Slot acc}} & \textbf{0.82} / 0.33 \\ \midrule
\textcolor{Orange}{\textbf{MR}} & \multicolumn{9}{l}{Gen\_Open\_topic ( hobbies = travel )} \\ \addlinespace
\textcolor{Orange}{\textbf{References}} & \multicolumn{9}{l}{\begin{tabular}[t]{@{}l@{}} \textbf{1 reference:} Let's talk about travel. \end{tabular}} \\ \addlinespace
\begin{tabular}[t]{@{}l@{}} \textcolor{Orange}{\textbf{Prompt 3}} \\ \textcolor{Orange}{\textbf{task demonstrator}} \end{tabular} & \multicolumn{9}{l}{Gen\_Open\_topic ( hobbies = cinema ) + Let's talk about cinema.} \\ \addlinespace
\begin{tabular}[t]{@{}l@{}} \textcolor{Purple}{\textbf{Baseline}} \\ \textcolor{Purple}{\textbf{Generations}} \end{tabular} & \multicolumn{9}{l}{\begin{tabular}[t]{@{}l@{}} \textbf{5 generations:} - We are going to travel, aren't we? \qquad - We will travel together. \qquad  - We're going to travel together! \\ - Which of these trips did you like the most? \qquad - Then we'll travel together.
\end{tabular}} \\ \addlinespace
\begin{tabular}[t]{@{}l@{}} \textcolor{Purple}{\textbf{Prompt 3}} \\ \textcolor{Purple}{\textbf{Generations}} \end{tabular} & \multicolumn{9}{l}{\begin{tabular}[t]{@{}l@{}} \textbf{3 generations:} - Let's talk about travel. \qquad - We can have a trip to the islands. \qquad - Let's talk about travelling
\end{tabular}} \\ \addlinespace
\textbf{Dataset} & \textcolor{Blue}{\textbf{EMPATHIC}} & \textcolor{Purple}{\textbf{BLEU}} & 0.03 / \textbf{0.45} & \textcolor{Purple}{\textbf{BLEURT}} & 0.36 / \textbf{0.69} & \textcolor{Purple}{\textbf{LaBSE}} & 0.54 / \textbf{0.78} & \textcolor{Purple}{\textbf{Slot acc}} & \textbf{0.80} / 0.67 \\ \bottomrule
\end{tabular}%
}
\end{table}

Table~\ref{tab:base_vs_prompt_exs} shows two representative examples of generations for Baseline and Prompt 3 inputs (one for ViGGO and one for EMPATHIC). In these examples, the input elements (with headers in orange in the table) are the selected Input MR (labelled MR), the references (including their number), and the task demonstrator used in the \textit{Prompt 3} representation. For the output (headers in purple), the table shows both the \textit{Baseline} and \textit{Prompt 3} generations (including their number) and the \textit{MR scores} separately (Baseline/Prompt 3). The highest MR scores for each metric are highlighted in bold. Finally, the dataset is specified (in blue). In this comparison, although metric scores were used to select the examples, we focus on comparing the generated sentences to understand why enriched representations have an opposite effect across the two datasets.

\paragraph{\textbf{Prompt representations produce omissions in ViGGO}} ViGGO generations with \textit{Prompt 3} tend to be shorter and contain omissions when the inputs contain high semantic context, i.e, a high number of attributes. The ViGGO example of Table~\ref{tab:base_vs_prompt_exs} illustrates it perfectly: all the \textit{Prompt 3} generations included in the table contain multiple omissions and are much shorter than the references, whereas only the second \textit{Baseline} generation omits \texttt{has\_linux\_release} and hallucinates (incorrect generation of \texttt{has\_mac\_release} ).  As a consequence, we may assume that concatenating the MR and the task demonstrator in the \textit{Prompt 3} representation results in an excessively long input, which makes the model omit some information.

\paragraph{\textbf{Model learns from the demonstrator in EMPATHIC}} The inspection of the EMPATHIC MRs shows \textit{Prompt 3} generations where the model seems to learn from the task demonstrator, especially when it fails to correctly generate the MR with the \textit{Baseline} representation. In the EMPATHIC example of Table~\ref{tab:base_vs_prompt_exs}, the generations \textit{"Let's talk about travel."} and \textit{"Let's talk about travelling"} clearly adopt the structure of \textit{Let's talk about cinema} from the task demonstrator. Although the other \textit{Prompt 3} generation \textit{"We can have a trip to the islands."} is incorrect for \texttt{Gen\_Open\_topic ( hobbies = travel )}, the \textit{Prompt 3} generations are more accurate overall than the \textit{Baseline} generations, all of which are incorrect. Therefore, we can assume that the proposed representations provide useful guidance in cases where fine-tuning alone is insufficient with the \textit{Baseline} MRs.

In conclusion, the enriched representation proves effective in smaller datasets where fine-tuning on the original MRs is insufficient to capture the generation patterns. However, in datasets with inputs containing a high density of semantic content, such as ViGGO, the resulting enriched MRs tend to become excessively long, leading the model to omit relevant information.

\section{Conclusions}
\label{sec:conclusion}

In this work, we have carried out a comprehensive analysis of the impact of different meaning representations on the performance of fine-tuned NLG models across multiple dialogue datasets. In particular, we propose prompt representations that integrate a task demonstrator in the meaning representation structure. To validate our approach, we assess them across four different datasets using five distinct metrics, offering a detailed analysis of the interactions between representation, dataset and metric features.

Our study shows that the proposed representations of meaning are especially effective for small datasets for complex tasks and high-output variability. In this context, the demonstrator enriches the input for MRs where the reduced fine-tuning is insufficient for an accurate generation. Conversely, in other datasets, the models perform better when they are fine-tuned with the original MRs. These datasets usually present long inputs with high semantic content, and the extra information of the task demonstrator might produce a forgetting effect in the models that leads to short generations with omission. Another positive effect of the enriched representations appears in zero-shot settings, i.e., before training the models. We have observed that even in those datasets where these enriched inputs have not been effective after fine-tuning, the models without training perform better with prompt representations than the baseline one. Finally, the comparison between prompt representations proves that more specific demonstrators in the input are more beneficial for the models, while the more generic ones may act as distractors, making the learning process for the models more difficult.

The thorough analysis of the metrics has also displayed relevant findings. First, metrics that evaluate lexical aspects based on n-gram overlapping do not seem adequate to assess output quality. In this regard, semantic metrics are more accurate. In a novel comparison of semantic metrics, we have observed that those metrics trained with human ratings detect omissions and semantic nuances not appreciable by others based on sentence embeddings. Finally, the reference-less metrics have confirmed the effect of the representations and datasets on the model's behaviour and revealed the model coherence with the input in terms of communicative intention and semantics.  

This study also highlights the suitability of GPT-2 models for generating language in the context of dialogue. Particularly, they present good adaptation capabilities. The excellent performance observed after just one training epoch, regardless of the representation or dataset, confirms this feature.

Future work should extend this study to the new era of LLMs to explore how prompt-enhanced MRs interact with the few-shot and zero-shot learning capabilities of these new models. Another promising direction is to explore whether prompt-based MRs enhance robustness and adaptability across different languages, contributing to the development of more inclusive and generalized dialogue systems. 

\section*{Acknowledgments}

This work has been partially supported by the CRYSTAL HORIZON-MSCA-SE grant 101182965, Spanish MCIU by the BRAINS project grant PID2024-155948OB-C51 and by the Basque Government under grant PRE 2020 1 0274. 
\begin{figure}[htbp]
    \centering
    \begin{subfigure}[t]{0.29\textwidth}
         \centering
         \includegraphics[width=\textwidth]{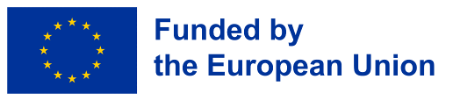}
         \label{fig:da_class_evolution_weighted_viggo}
    \end{subfigure}
    \begin{subfigure}[t]{0.29\textwidth}
         \centering
         \includegraphics[width=\textwidth]{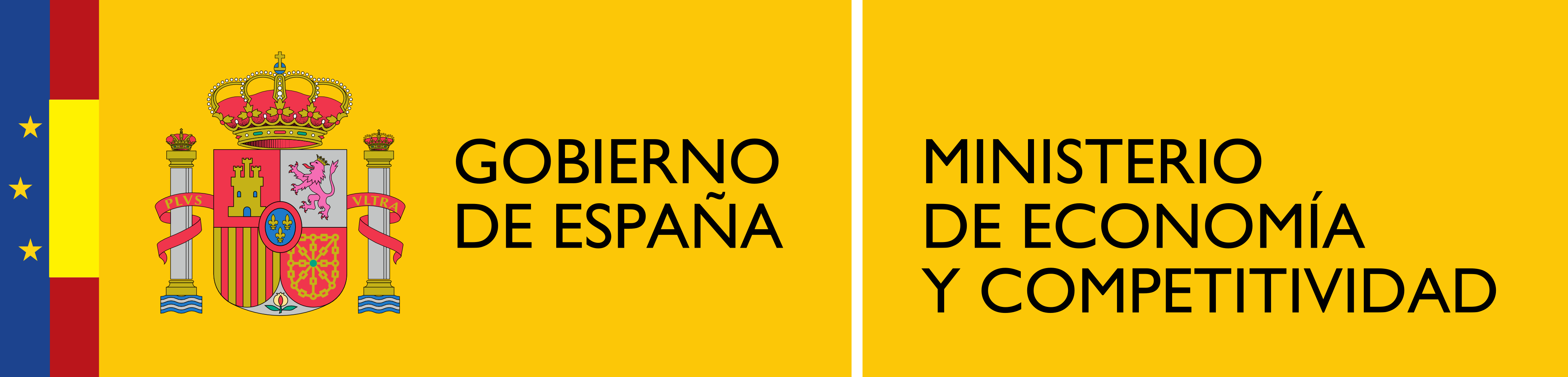}
    \end{subfigure}
\end{figure}

\bibliographystyle{unsrtnat}
\bibliography{Sample}  

@article{vazquez2023dialogue,
  title={Dialogue management and language generation for a robust conversational virtual coach: Validation and user study},
  author={V{\'a}zquez, Alain and L{\'o}pez Zorrilla, Asier and Olaso, Javier Mikel and Torres, Mar{\'\i}a In{\'e}s},
  journal={Sensors},
  volume={23},
  number={3},
  pages={1423},
  year={2023},
  publisher={MDPI}
}

@inproceedings{olaso2021empathic,
  title={The {EMPATHIC} virtual coach: A demo},
  author={Olaso, Javier M and V{\'a}zquez, Alain and Ben Letaifa, Leila and De Velasco, Mikel and Mtibaa, Aymen and Hmani, Mohamed Amine and Petrovska-Delacr{\'e}taz, Dijana and Chollet, G{\'e}rard and Montenegro, C{\'e}sar and L{\'o}pez-Zorrilla, Asier and others},
  booktitle={Proceedings of the 2021 International Conference on Multimodal Interaction},
  pages={848--851},
  year={2021}
}

@inproceedings{torres2019empathic,
  title={The {EMPATHIC} project: mid-term achievements},
  author={Torres, Mar{\'\i}a In{\'e}s and Olaso, Javier Mikel and Montenegro, C{\'e}sar and Santana, Roberto and V{\'a}zquez, Alain and Justo, Raquel and Lozano, Jos{\'e} Antonio and Schl{\"o}gl, Stephan and Chollet, G{\'e}rard and Dugan, Nazim and others},
  booktitle={Proceedings of the 12th ACM International Conference on Pervasive Technologies Related to Assistive Environments},
  pages={629--638},
  year={2019}
}

@incollection{leach2020behavioural,
  title={Behavioural coaching: The {GROW} model},
  author={Leach, Sarah},
  booktitle={The Coaches' Handbook},
  pages={176--186},
  year={2020},
  publisher={Routledge}
}

@inproceedings{juraska2019viggo,
  title={{ViGGO}: A Video Game Corpus for Data-To-Text Generation in Open-Domain Conversation},
  author={Juraska, Juraj and Bowden, Kevin and Walker, Marilyn},
  booktitle={Proceedings of the 12th International Conference on Natural Language Generation},
  pages={164--172},
  year={2019}
}

@inproceedings{budzianowski2018multiwoz,
  title={{MultiWOZ}-A Large-Scale Multi-Domain Wizard-of-Oz Dataset for Task-Oriented Dialogue Modelling},
  author={Budzianowski, Pawe{\l} and Wen, Tsung-Hsien and Tseng, Bo-Hsiang and Casanueva, I{\~n}igo and Ultes, Stefan and Ramadan, Osman and Gasic, Milica},
  booktitle={Proceedings of the 2018 Conference on Empirical Methods in Natural Language Processing},
  pages={5016--5026},
  year={2018}
}

@inproceedings{vazquez2023how,
  title={How Should We Represent Dialog Acts to
Leverage Pretrained Natural Language
Generators?},
  author={\vphantom{2}V{\'a}zquez, Alain and L{\'o}pez Zorrilla, Asier and Torres, M. In{\'e}s},
  year={2023},
  booktitle={Presented at 13th International Workshop on Spoken Dialogue Systems Technology},
}

@inproceedings{chen2014systematic,
  title={A systematic comparison of smoothing techniques for sentence-level {BLEU}},
  author={Chen, Boxing and Cherry, Colin},
  booktitle={Proceedings of the ninth workshop on statistical machine translation},
  pages={362--367},
  year={2014}
}

@inproceedings{ramirez2023controllable,
  title={Controllable Generation of Dialogue Acts for Dialogue Systems via Few-Shot Response Generation and Ranking},
  author={Ramirez, Angela and Agarwal, Kartik and Juraska, Juraj and Garg, Utkarsh and Walker, Marilyn},
  booktitle={Proceedings of the 24th Annual Meeting of the Special Interest Group on Discourse and Dialogue},
  pages={355--369},
  year={2023}
}

@article{sanh2019distilbert,
  title={{DistilBERT}, a distilled version of {BERT}: smaller, faster, cheaper and lighter},
  author={Sanh, V},
  journal={arXiv preprint arXiv:1910.01108},
  year={2019}
}

@inproceedings{reed2022jurassic,
  title={Jurassic is (almost) all you need: Few-shot meaning-to-text generation for open-domain dialogue},
  author={Reed, Lena and Li, Cecilia and Ramirez, Angela and Wu, Liren and Walker, Marilyn},
  booktitle={Conversational AI for Natural Human-Centric Interaction: 12th International Workshop on Spoken Dialogue System Technology, IWSDS 2021, Singapore},
  pages={99--119},
  year={2022},
  organization={Springer}
}

@inproceedings{novikova2017e2e,
  title={The {E2E} Dataset: New Challenges For End-to-End Generation},
  author={Novikova, Jekaterina and Du{\v{s}}ek, Ond{\v{r}}ej and Rieser, Verena},
  booktitle={Proceedings of the 18th Annual SIGdial Meeting on Discourse and Dialogue},
  pages={201--206},
  year={2017}
}

@article{radford2019language,
  title={Language models are unsupervised multitask learners},
  author={Radford, Alec and Wu, Jeffrey and Child, Rewon and Luan, David and Amodei, Dario and Sutskever, Ilya},
  journal={OpenAI blog},
  volume={1},
  number={8},
  pages={9},
  year={2019}
}

@inproceedings{papineni2002bleu,
  title={{BLEU}: a method for automatic evaluation of machine translation},
  author={Papineni, Kishore and Roukos, Salim and Ward, Todd and Zhu, Wei-Jing},
  booktitle={Proceedings of the 40th annual meeting of the Association for Computational Linguistics},
  pages={311--318},
  year={2002}
}

@inproceedings{sellam2020bleurt,
  title={{BLEURT}: Learning Robust Metrics for Text Generation},
  author={Sellam, Thibault and Das, Dipanjan and Parikh, Ankur},
  booktitle={Proceedings of the 58th Annual Meeting of the Association for Computational Linguistics},
  pages={7881--7892},
  year={2020}
}

@inproceedings{feng2022language,
  title={Language-agnostic {BERT} Sentence Embedding},
  author={Feng, Fangxiaoyu and Yang, Yinfei and Cer, Daniel and Arivazhagan, Naveen and Wang, Wei},
  booktitle={Proceedings of the 60th Annual Meeting of the Association for Computational Linguistics (Volume 1: Long Papers)},
  pages={878--891},
  year={2022}
}

@article{celikyilmaz2020evaluation,
  title={Evaluation of text generation: A survey},
  author={Celikyilmaz, Asli and Clark, Elizabeth and Gao, Jianfeng},
  journal={arXiv preprint arXiv:2006.14799},
  year={2020}
}

@inproceedings{gehrmann2021gem,
  title={The {GEM} Benchmark: Natural Language Generation, its Evaluation and Metrics},
  author={Gehrmann, Sebastian and Adewumi, Tosin and Aggarwal, Karmanya and Ammanamanchi, Pawan Sasanka and Anuoluwapo, Aremu and Bosselut, Antoine and Chandu, Khyathi Raghavi and Clinciu, Miruna and Das, Dipanjan and Dhole, Kaustubh D and others},
  booktitle={1st Workshop on Natural Language Generation, Evaluation, and Metrics 2021},
  pages={96--120},
  year={2021},
  organization={Association for Computational Linguistics}
}

@inproceedings{sennrich2016neural,
  title={Neural Machine Translation of Rare Words with Subword Units},
  author={Sennrich, Rico and Haddow, Barry and Birch, Alexandra},
  booktitle={Proceedings of the 54th Annual Meeting of the Association for Computational Linguistics (Volume 1: Long Papers)},
  pages={1715--1725},
  year={2016}
}

@article{kingma2014adam,
  title={Adam: A method for stochastic optimization},
  author={Kingma, Diederik P and Ba, Jimmy},
  journal={arXiv preprint arXiv:1412.6980},
  year={2014}
}

@inproceedings{li-etal-2020-slot,
    title = "Slot-consistent {NLG} for Task-oriented Dialogue Systems with Iterative Rectification Network",
    author = "Li, Yangming  and
      Yao, Kaisheng  and
      Qin, Libo  and
      Che, Wanxiang  and
      Li, Xiaolong  and
      Liu, Ting",
    booktitle = "Proceedings of the 58th Annual Meeting of the Association for Computational Linguistics",
    month = jul,
    year = "2020",
    address = "Online",
    publisher = "Association for Computational Linguistics",
    url = "https://aclanthology.org/2020.acl-main.10",
    doi = "10.18653/v1/2020.acl-main.10",
    pages = "97--106",
}

@article{brown2020language,
  title={Language models are few-shot learners},
  author={Brown, Tom and Mann, Benjamin and Ryder, Nick and Subbiah, Melanie and Kaplan, Jared D and Dhariwal, Prafulla and Neelakantan, Arvind and Shyam, Pranav and Sastry, Girish and Askell, Amanda and others},
  journal={Advances in neural information processing systems},
  volume={33},
  pages={1877--1901},
  year={2020}
}

@article{mousavi2024llms,
  title={Are {LLMs} Robust for Spoken Dialogues?},
  author={Mousavi, Seyed Mahed and Roccabruna, Gabriel and Alghisi, Simone and Rizzoli, Massimo and Ravanelli, Mirco and Riccardi, Giuseppe},
  journal={arXiv preprint arXiv:2401.02297},
  year={2024}
}

@inproceedings{zhang2020dialogpt,
  title={{DialoGPT}: Large-Scale Generative Pre-training for Conversational Response Generation},
  author={Zhang, Yizhe and Sun, Siqi and Galley, Michel and Chen, Yen-Chun and Brockett, Chris and Gao, Xiang and Gao, Jianfeng and Liu, Jingjing and Dolan, William B},
  booktitle={Proceedings of the 58th Annual Meeting of the Association for Computational Linguistics: System Demonstrations},
  pages={270--278},
  year={2020}
}

@article{vaswani2017attention,
  title={Attention is all you need},
  author={Vaswani, A},
  journal={Advances in Neural Information Processing Systems},
  year={2017}
}

@article{raffel2020exploring,
  title={Exploring the limits of transfer learning with a unified text-to-text transformer},
  author={Raffel, Colin and Shazeer, Noam and Roberts, Adam and Lee, Katherine and Narang, Sharan and Matena, Michael and Zhou, Yanqi and Li, Wei and Liu, Peter J},
  journal={Journal of machine learning research},
  volume={21},
  number={140},
  pages={1--67},
  year={2020}
}

@inproceedings{devlin2019bert,
  title={{BERT}: Pre-training of deep bidirectional transformers for language understanding},
  author={Devlin, Jacob and Chang, Ming-Wei and Lee, Kenton and Toutanova, Kristina},
  booktitle={Proceedings of the 2019 conference of the North American chapter of the association for computational linguistics: human language technologies, volume 1 (long and short papers)},
  pages={4171--4186},
  year={2019}
}

@inproceedings{peng2020few,
    title = "Few-shot Natural Language Generation for Task-Oriented Dialog",
    author = "Peng, Baolin  and
      Zhu, Chenguang  and
      Li, Chunyuan  and
      Li, Xiujun  and
      Li, Jinchao  and
      Zeng, Michael  and
      Gao, Jianfeng",
    editor = "Cohn, Trevor  and
      He, Yulan  and
      Liu, Yang",
    booktitle = "Findings of the Association for Computational Linguistics: EMNLP 2020",
    month = nov,
    year = "2020",
    address = "Online",
    publisher = "Association for Computational Linguistics",
    url = "https://aclanthology.org/2020.findings-emnlp.17",
    doi = "10.18653/v1/2020.findings-emnlp.17",
    pages = "172--182",
}

@inproceedings{vazquez2024knowledge,
  title={Knowledge-Grounded Dialogue Act Transfer using Prompt-Based Learning for Controllable Open-Domain {NLG}},
  author={Vazquez, Alain and Ramirez, Angela Maria and Pullabhotla, Neha and Qiang, Nan and Zhang, Ranran Haoran and Walker, Marilyn and Torres, Maria In{\'e}s},
  booktitle={Proceedings of the 25th Annual Meeting of the Special Interest Group on Discourse and Dialogue},
  pages={78--91},
  year={2024}
}

@inproceedings{liu2023large,
  title={Large Language Models guided Generative Prompt for Dialogue Generation},
  author={Liu, Sijie and Fang, Yiquan and Cheng, Hua and Pan, Yiming and Liu, Yufei and Gao, Caiting},
  booktitle={2023 International Conference on Cyber-Enabled Distributed Computing and Knowledge Discovery (CyberC)},
  pages={10--17},
  year={2023},
  organization={IEEE}
}

@inproceedings{wen2015semantically,
    title = "Semantically Conditioned {LSTM}-based Natural Language Generation for Spoken Dialogue Systems",
    author = "Wen, Tsung-Hsien  and
      Ga{\v{s}}i{\'c}, Milica  and
      Mrk{\v{s}}i{\'c}, Nikola  and
      Su, Pei-Hao  and
      Vandyke, David  and
      Young, Steve",
    editor = "M{\`a}rquez, Llu{\'\i}s  and
      Callison-Burch, Chris  and
      Su, Jian",
    booktitle = "Proceedings of the 2015 Conference on Empirical Methods in Natural Language Processing",
    month = sep,
    year = "2015",
    address = "Lisbon, Portugal",
    publisher = "Association for Computational Linguistics",
    url = "https://aclanthology.org/D15-1199",
    doi = "10.18653/v1/D15-1199",
    pages = "1711--1721",
}

@inproceedings{rastogi2020towards,
  title={Towards scalable multi-domain conversational agents: The schema-guided dialogue dataset},
  author={Rastogi, Abhinav and Zang, Xiaoxue and Sunkara, Srinivas and Gupta, Raghav and Khaitan, Pranav},
  booktitle={Proceedings of the AAAI conference on artificial intelligence},
  volume={34},
  number={05},
  pages={8689--8696},
  year={2020}
}

@inproceedings{moon2019opendialkg,
  title={{OpenDialKG}: Explainable Conversational Reasoning with Attention-based Walks over Knowledge Graphs},
  author={Moon, Seungwhan and Shah, Pararth and Kumar, Anuj and Subba, Rajen},
  booktitle={Proceedings of the 57th annual meeting of the association for computational linguistics},
  pages={845--854},
  year={2019}
}

@inproceedings{mi2020continual,
  title={Continual Learning for Natural Language Generation in Task-oriented Dialog Systems},
  author={Mi, Fei and Chen, Liangwei and Zhao, Mengjie and Huang, Minlie and Faltings, Boi},
  booktitle={Findings of the Association for Computational Linguistics: EMNLP 2020},
  pages={3461--3474},
  year={2020}
}

@inproceedings{wang2021modelling,
  title={Modelling hierarchical structure between dialogue policy and natural language generator with option framework for task-oriented dialogue system},
  author={Wang, Jianhong and Zhang, Y and Kim, Tae-Kyun and Gu, Yunjie},
  booktitle={Ninth International Conference on Learning Representation (ICLR)},
  year={2021},
  organization={The International Conference on Learning Representations (ICLR)}
}

@inproceedings{gao2021making,
  title={Making Pre-trained Language Models Better Few-shot Learners},
  author={Gao, Tianyu and Fisch, Adam and Chen, Danqi},
  booktitle={Proceedings of the 59th Annual Meeting of the Association for Computational Linguistics and the 11th International Joint Conference on Natural Language Processing (Volume 1: Long Papers)},
  year={2021},
  organization={Association for Computational Linguistics}
}

@inproceedings{jin2024analyzing,
  title={Analyzing the Role of Semantic Representations in the Era of Large Language Models},
  author={Jin, Zhijing and Chen, Yuen and Adauto, Fernando Gonzalez and Liu, Jiarui and Zhang, Jiayi and Michael, Julian and Sch{\"o}lkopf, Bernhard and Diab, Mona},
  booktitle={Proceedings of the 2024 Conference of the North American Chapter of the Association for Computational Linguistics: Human Language Technologies (Volume 1: Long Papers)},
  pages={3781--3798},
  year={2024}
}

@inproceedings{baptista2024lexicalized,
  title={Lexicalized Meaning Representation {(LMR)}},
  author={Baptista, Jorge and Reis, S{\'o}nia and Dias, Jo{\~a}o and Santos, Pedro A},
  booktitle={Workshop on Designing Meaning Representation},
  pages={101--111},
  year={2024},
  organization={ELRA; ICCL}
}

@inproceedings{banarescu2013abstract,
  title={Abstract meaning representation for sembanking},
  author={Banarescu, Laura and Bonial, Claire and Cai, Shu and Georgescu, Madalina and Griffitt, Kira and Hermjakob, Ulf and Knight, Kevin and Koehn, Philipp and Palmer, Martha and Schneider, Nathan},
  booktitle={Proceedings of the 7th linguistic annotation workshop and interoperability with discourse},
  pages={178--186},
  year={2013}
}

@inproceedings{du2020schema,
  title={Schema-Guided Natural Language Generation},
  author={Du, Yuheng and Oraby, Shereen and Perera, Vittorio and Shen, Minmin and Narayan-Chen, Anjali and Chung, Tagyoung and Venkatesh, Anu and Hakkani-Tur, Dilek},
  booktitle={13th International Conference on Natural Language Generation, INLG 2020},
  pages={283--295},
  year={2020},
  organization={Association for Computational Linguistics (ACL)}
}

@inproceedings{moryossef2019step,
  title={Step-by-Step: Separating Planning from Realization in Neural Data-to-Text Generation},
  author={Moryossef, Amit and Goldberg, Yoav and Dagan, Ido},
  booktitle={Proceedings of the 2019 Conference of the North American Chapter of the Association for Computational Linguistics: Human Language Technologies, Volume 1 (Long and Short Papers)},
  pages={2267--2277},
  year={2019}
}

@inproceedings{colin2016webnlg,
  title={The {WebNLG} Challenge: Generating Text from {DBPedia} Data},
  author={Colin, Emilie and Gardent, Claire and M’rabet, Yassine and Narayan, Shashi and Perez, Laura},
  booktitle={9th International Natural Language Generation conference},
  pages={163--167},
  year={2016},
  organization={Association for Computational Linguistics}
}

@inproceedings{parikh2020totto,
  title={{ToTTo}: A Controlled Table-To-Text Generation Dataset},
  author={Parikh, Ankur and Wang, Xuezhi and Gehrmann, Sebastian and Faruqui, Manaal and Dhingra, Bhuwan and Yang, Diyi and Das, Dipanjan},
  booktitle={Proceedings of the 2020 Conference on Empirical Methods in Natural Language Processing (EMNLP)},
  pages={1173--1186},
  year={2020}
}

@inproceedings{wang2022task,
  title={Task-oriented dialogue system as natural language generation},
  author={Wang, Weizhi and Zhang, Zhirui and Guo, Junliang and Dai, Yinpei and Chen, Boxing and Luo, Weihua},
  booktitle={Proceedings of the 45th international ACM SIGIR conference on research and development in information retrieval},
  pages={2698--2703},
  year={2022}
}

@inproceedings{wu2023diacttod,
  title={{DiactTOD}: Learning Generalizable Latent Dialogue Acts for Controllable Task-Oriented Dialogue Systems},
  author={Wu, Qingyang and Gung, James and Shu, Raphael and Zhang, Yi},
  booktitle={Proceedings of the 24th Annual Meeting of the Special Interest Group on Discourse and Dialogue},
  pages={255--267},
  year={2023}
}

@inproceedings{gu2022ppt,
  title={{PPT}: Pre-trained Prompt Tuning for Few-shot Learning},
  author={Gu, Yuxian and Han, Xu and Liu, Zhiyuan and Huang, Minlie},
  booktitle={Proceedings of the 60th Annual Meeting of the Association for Computational Linguistics (Volume 1: Long Papers)},
  pages={8410--8423},
  year={2022}
}

@article{tohidi2022short,
  title={A Short Review of Abstract Meaning Representation Applications},
  author={Tohidi, Nasim and Dadkhah, Chitra},
  journal={Modeling and Simulation in Electrical and Electronics Engineering},
  volume={2},
  number={3},
  pages={1--9},
  year={2022},
  publisher={Semnan University}
}

@inproceedings{yang2024improving,
  title={Improving medical dialogue generation with abstract meaning representations},
  author={Yang, Bohao and Tang, Chen and Lin, Chenghua},
  booktitle={ICASSP 2024-2024 IEEE International Conference on Acoustics, Speech and Signal Processing (ICASSP)},
  pages={11826--11830},
  year={2024},
  organization={IEEE}
}

@inproceedings{brutti2022abstract,
  title={Abstract meaning representation for gesture},
  author={Brutti, Richard and Donatelli, Lucia and Lai, Kenneth and Pustejovsky, James},
  booktitle={Proceedings of the thirteenth language resources and evaluation conference},
  year={2022}
}

@inproceedings{bonial2023abstract,
  title={Abstract meaning representation for grounded human-robot communication},
  author={Bonial, Claire and Foresta, Julie and Fung, Nicholas C and Hayes, Cory and Osteen, Philip and Arkin, Jacob and Hedegaard, Benned and Howard, Thomas},
  booktitle={Proceedings of the Fourth International Workshop on Designing Meaning Representations},
  pages={34--44},
  year={2023}
}

@article{sai2022survey,
  title={A survey of evaluation metrics used for {NLG} systems},
  author={Sai, Ananya B and Mohankumar, Akash Kumar and Khapra, Mitesh M},
  journal={ACM Computing Surveys (CSUR)},
  volume={55},
  number={2},
  pages={1--39},
  year={2022},
  publisher={ACM New York, NY}
}

@inproceedings{schmidtova2024automatic,
  title={Automatic Metrics in Natural Language Generation: A survey of Current Evaluation Practices},
  author={Schmidtov{\'a}, Patr{\'\i}cia and Mahamood, Saad and Balloccu, Simone and Du{\v{s}}ek, Ond{\v{r}}ej and Gatt, Albert and Gkatzia, Dimitra and Howcroft, David M and Pl{\'a}tek, Ond{\v{r}}ej and Sivaprasad, Adarsa},
  booktitle={Proceedings of the 17th International Natural Language Generation Conference},
  pages={557--583},
  year={2024}
}

@inproceedings{banerjee2005meteor,
  title={{METEOR}: An automatic metric for {MT} evaluation with improved correlation with human judgments},
  author={Banerjee, Satanjeev and Lavie, Alon},
  booktitle={Proceedings of the acl workshop on intrinsic and extrinsic evaluation measures for machine translation and/or summarization},
  pages={65--72},
  year={2005}
}

@inproceedings{doddington2002automatic,
  title={Automatic evaluation of machine translation quality using n-gram co-occurrence statistics},
  author={Doddington, George},
  booktitle={Proceedings of the second international conference on Human Language Technology Research},
  pages={138--145},
  year={2002}
}

@inproceedings{lin2004rouge,
  title={{ROUGE}: A Package for Automatic Evaluation of Summaries},
  author={Lin, Chin-Yew},
  booktitle={Text summarization branches out},
  pages={74--81},
  year={2004}
}

@inproceedings{wang2023dspm,
  title={{DSPM-NLG}: A Dual Supervised Pre-trained Model for Few-shot Natural Language Generation in Task-oriented Dialogue System},
  author={Wang, Yufan and Zou, Bowei and Fan, Rui and Aw, Aiti and He, Tingting},
  booktitle={Findings of the Association for Computational Linguistics: ACL 2023},
  pages={12389--12402},
  year={2023}
}

@article{algherairy2024review,
  title={A review of dialogue systems: current trends and future directions},
  author={Algherairy, Atheer and Ahmed, Moataz},
  journal={Neural Computing and Applications},
  volume={36},
  number={12},
  pages={6325--6351},
  year={2024},
  publisher={Springer}
}

@inproceedings{wang2024towards,
  title={Towards Robustness and Diversity: Continual Learning in Dialog Generation with Text-Mixup and Batch Nuclear-Norm Maximization},
  author={Wang, Zihan and Xiao, Jiayu and Li, Mengxiang and He, Zhongjiang and Li, Yongxiang and Wang, Chao and Song, Shuangyong},
  booktitle={2024 International Joint Conference on Neural Networks (IJCNN)},
  pages={1--8},
  year={2024},
  organization={IEEE}
}

@inproceedings{galland2024generating,
  title={Generating Unexpected yet Relevant User Dialog Acts},
  author={Galland, Lucie and Pelachaud, Catherine and Pecune, Florian},
  booktitle={Proceedings of the 25th Annual Meeting of the Special Interest Group on Discourse and Dialogue},
  pages={192--203},
  year={2024}
}

@article{griol2024combining,
  title={Combining statistical dialog management and intent recognition for enhanced response selection},
  author={Griol, David and Callejas, Zoraida},
  journal={Logic Journal of the IGPL},
  pages={jzae045},
  year={2024},
  publisher={Oxford University Press}
}

@article{ji2024wavchat,
  title={{WavChat}: A Survey of Spoken Dialogue Models},
  author={Ji, Shengpeng and Chen, Yifu and Fang, Minghui and Zuo, Jialong and Lu, Jingyu and Wang, Hanting and Jiang, Ziyue and Zhou, Long and Liu, Shujie and Cheng, Xize and others},
  journal={arXiv preprint arXiv:2411.13577},
  year={2024}
}

@article{lappin2024assessing,
  title={Assessing the strengths and weaknesses of Large Language Models},
  author={Lappin, Shalom},
  journal={Journal of Logic, Language and Information},
  volume={33},
  number={1},
  pages={9--20},
  year={2024},
  publisher={Springer}
}

@inproceedings{hu2022dialogue,
  title={Dialogue Meaning Representation for Task-Oriented Dialogue Systems},
  author={Hu, Xiangkun and Dai, Junqi and Yan, Hang and Zhang, Yi and Guo, Qipeng and Qiu, Xipeng and Zhang, Zheng},
  booktitle={Findings of the Association for Computational Linguistics: EMNLP 2022},
  pages={223--237},
  year={2022}
}

@article{montenegro2019dialogue,
  title={A dialogue-act taxonomy for a virtual coach designed to improve the life of elderly},
  author={Montenegro, C{\'e}sar and L{\'o}pez Zorrilla, Asier and Mikel Olaso, Javier and Santana, Roberto and Justo, Raquel and Lozano, Jose A and Torres, Mar{\'\i}a In{\'e}s},
  journal={Multimodal Technologies and Interaction},
  volume={3},
  number={3},
  pages={52},
  year={2019},
  publisher={MDPI}
}

@article{zorrilla2022multilingual,
author = {L\'{o}pez Zorrilla, Asier and Torres, M. In\'{e}s},
title = {A Multilingual Neural Coaching Model with Enhanced Long-term Dialogue Structure},
year = {2022},
issue_date = {June 2022},
publisher = {Association for Computing Machinery},
address = {New York, NY, USA}, volume = {12},
number = {2},
issn = {2160-6455},
url = {https://doi.org/10.1145/3487066},
doi = {10.1145/3487066},
journal = {ACM Trans. Interact. Intell. Syst.},
month = jul,
articleno = {16},
numpages = {47} }

@article{audio2023zorrilla,
  author={López Zorrilla, Asier and Torres, María Inés and Cuayáhuitl, Heriberto},
  journal={IEEE/ACM Transactions on Audio, Speech, and Language Processing}, 
  title={Audio Embedding-Aware Dialogue Policy Learning}, 
  year={2023},
  volume={31},
  number={},
  pages={525-538},
  doi={10.1109/TASLP.2022.3225658}}

@article{question2023ruiz,
title = {Question answering models for human–machine interaction in the manufacturing industry},
journal = {Computers in Industry},
volume = {151},
pages = {103988},
year = {2023},
issn = {0166-3615},
doi = {https://doi.org/10.1016/j.compind.2023.103988},
url = {https://www.sciencedirect.com/science/article/pii/S0166361523001380},
author = {Eneko Ruiz and María Inés Torres and Arantza {del Pozo}}
}

@inproceedings{vazquez2025prompt,
    title = "Prompt-based Language Generation for Complex Conversational Coaching Tasks across Languages",
    author = "V{\'a}zquez, Alain  and
    Torres, Maria Ines",
    booktitle = "Proceedings of the 26th Annual Meeting of the Special Interest Group on Discourse and Dialogue",
    month = aug,
    year = "2025",
    address = "Avignon, France",
    publisher = "Association for Computational Linguistics",
    url = "https://aclanthology.org/2025.sigdial-1.48/",
    pages = "601--608",
}

@article{achiam2023gpt,
  title={{GPT-4} technical report},
  author={Achiam, Josh and Adler, Steven and Agarwal, Sandhini and Ahmad, Lama and Akkaya, Ilge and Aleman, Florencia Leoni and Almeida, Diogo and Altenschmidt, Janko and Altman, Sam and Anadkat, Shyamal and others},
  journal={arXiv preprint arXiv:2303.08774},
  year={2023}
}

@article{team2023gemini,
  title={Gemini: A family of highly capable multimodal models},
  author={Team, Gemini and Anil, Rohan and Borgeaud, Sebastian and Alayrac, Jean-Baptiste and Yu, Jiahui and Soricut, Radu and Schalkwyk, Johan and Dai, Andrew M and Hauth, Anja and Millican, Katie and others},
  journal={arXiv preprint arXiv:2312.11805},
  year={2023}
}

@article{dubey2024llama,
  title={The {Llama} 3 Herd of Models},
  author={Dubey, Abhimanyu and Jauhri, Abhinav and Pandey, Abhinav and Kadian, Abhishek and Al-Dahle, Ahmad and Letman, Aiesha and Mathur, Akhil and Schelten, Alan and Yang, Amy and Fan, Angela and others},
  journal={arXiv e-prints},
  pages={arXiv--2407},
  year={2024}
}

@phdthesis{hryhoryeva2025data,
    author ={Hryhoryeva, Darya} ,
    title = {Data-to-text Generation with Detailed Inputs} ,
    school = {Univerzita Karlova, Matematicko-fyzik{\'a}ln{\'\i} fakulta},
    year = {2025}
}

@phdthesis{vazquez2026role,
    author = "V{\'a}zquez, Alain",
    title = {The Role of Meaning Representations in Natural Language Generation for Dialogue},
    school = {University of the Basque Country},
    year = {2026}
}

\end{document}